\documentclass[lettersize,journal]{IEEEtran}
\usepackage{amsmath,amsfonts,amssymb}
\usepackage{algorithmic}
\usepackage{algorithm}
\usepackage{array}
\usepackage{cite}
\usepackage[caption=false,font=normalsize,labelfont=sf,textfont=sf]{subfig}
\usepackage{enumitem}
\usepackage{textcomp}
\usepackage{stfloats}
\usepackage{url}
\usepackage{verbatim}
\usepackage{graphicx}
\usepackage{gensymb}
\usepackage{booktabs}
\usepackage{tabularx}
\usepackage{titlesec}
\usepackage{makecell}
\usepackage{multirow}
\usepackage{hhline}
\usepackage{indentfirst}
\usepackage{svg}
\usepackage[font=small]{caption}
\usepackage[caption=false]{subfig}
\usepackage[utf8]{inputenc}
\usepackage{pifont}
\usepackage[pagebackref=true,breaklinks=true,letterpaper=true,colorlinks,bookmarks=false]{hyperref}

\definecolor{bbox}{RGB}{48, 188, 227}
\definecolor{rbbox}{RGB}{121, 224, 167}
\definecolor{bfov}{RGB}{224, 161, 101}
\definecolor{rbfov}{RGB}{224, 107, 96}
\definecolor{metmsk}{RGB}{3,130,255}
\definecolor{metgrd}{RGB}{255,130,3}
\definecolor{sec}{RGB}{39, 87, 146}
\definecolor{fir}{RGB}{199, 85, 93}
\newcommand{\st}[1]{\textcolor{fir}{\bf #1}} 
\newcommand{\nd}[1]{\textcolor{sec}{\bf #1}}

\usepackage{colortbl}

\definecolor{firstcolor}{RGB}{241, 213, 187}
\definecolor{secondcolor}{RGB}{248, 234, 221}
\definecolor{thirdcolor}{RGB}{252, 245, 238}

\makeatletter
\renewcommand{\thebibliography}[1]{%
  \section*{\refname}%
  \scriptsize  
  \list{\@biblabel{\@arabic\c@enumiv}}%
       {\settowidth\labelwidth{\@biblabel{#1}}%
        \leftmargin\labelwidth
        \advance\leftmargin\labelsep
        \itemsep 0pt
        \parsep 0pt
        \@openbib@code
        \usecounter{enumiv}%
        \let\p@enumiv\@empty
        \renewcommand\theenumiv{\@arabic\c@enumiv}}%
  \sloppy
  \clubpenalty4000
  \@clubpenalty \clubpenalty
  \widowpenalty4000%
  \sfcode`\.\@m}
\makeatother

\newcommand{\edit}[1]{\textcolor{magenta}{#1}}

\newcolumntype{H}{>{\setbox0=\hbox\bgroup}c<{\egroup}@{}}
\makeatletter
\newcommand{\multiline}[1]{%
  \begin{tabularx}{\dimexpr\linewidth-\ALG@thistlm}[t]{@{}X@{}}
    #1
  \end{tabularx}
}
\makeatother

\begin{document}

\title{360VOTS: Visual Object Tracking and Segmentation in Omnidirectional Videos}

\author{Yinzhe Xu, Huajian Huang*, Yingshu Chen and Sai-Kit Yeung \\ \quad \\
\vspace{-0.1cm}
Homepage: \url{https://360vots.hkustvgd.com/}
\thanks{
Yinzhe Xu is with the Division of Integrative Systems and Design. Huajian Huang and Yingshu Chen are with the Department of Computer Science and Engineering (e-mail: hhuangbg, ychengw, yxuck@connect.ust.hk).

Sai-Kit Yeung is with the Division of Integrative Systems and Design, the Department of Computer Science and Engineering, and the Department of Ocean Science (e-mail: saikit@ust.hk).

The Hong Kong University of Science and Technology, Clear Water Bay, Kowloon, Hong Kong 

* corresponding author
}
}


\maketitle

\begin{abstract}
Visual object tracking and segmentation in omnidirectional videos are challenging due to the wide field-of-view and large spherical distortion brought by 360$\degree$ images. 
To alleviate these problems, we introduce a novel representation, extended bounding field-of-view (eBFoV), for target localization and use it as the foundation of a general 360 tracking framework which is applicable for both omnidirectional visual object tracking and segmentation tasks. 
Building upon our previous work on omnidirectional visual object tracking (360VOT), we propose a comprehensive dataset and benchmark that incorporates a new component called omnidirectional video object segmentation (360VOS). The 360VOS dataset includes 290 sequences accompanied by dense pixel-wise masks and covers a broader range of target categories. To support both the development and evaluation of algorithms in this domain, we divide the dataset into a training subset with 170 sequences and a testing subset with 120 sequences.
Furthermore, we tailor evaluation metrics for both omnidirectional tracking and segmentation to ensure rigorous assessment. Through extensive experiments, we benchmark state-of-the-art approaches and demonstrate the effectiveness of our proposed 360 tracking framework and training dataset.
\end{abstract}



\section{Introduction}
\label{sec:intro}
\IEEEPARstart{V}{isual} object tracking (VOT) and video object segmentation (VOS) are two essential tasks in computer vision for spatially and temporally localizing target objects in video sequences. A robust and precise approach is demanded in various applications such as video analysis, human-machine interaction, and intelligent robots. 
While both VOT and VOS are integral to understanding object dynamics in videos, their fundamental differences are in the level of localization granularity. VOT approaches typically utilize the bounding box to represent the target position and rough size. 
In contrast, VOS approaches seek to provide pixel-wise localization of the object's silhouette frame-by-frame.
In the last decade, a remarkable development can be seen in these tasks with the flourish of excellent trackers like \cite{kcf, mdnet, eco, siamrpn++, dimp} in VOT and \cite{stm, unicorn, cfbi, deaot, xmem} in VOS, as well as various benchmarks including \cite{otb100, uav123, VOT, lasot, got10k} for VOT and \cite{fbms, lvos, davis2017, ytvos2018} for VOS. Whereas most existing research focuses on perspective scenarios, there is little attention paid to object tracking and segmentation in the omnidirectional view.

\begin{figure*}
    \centering
    \captionsetup{type=figure}
    \def\imgw{0.246}
    \def\imgh{0.116}
    
    \subfloat{\includegraphics[width=\imgw\linewidth, height=\imgh\linewidth]{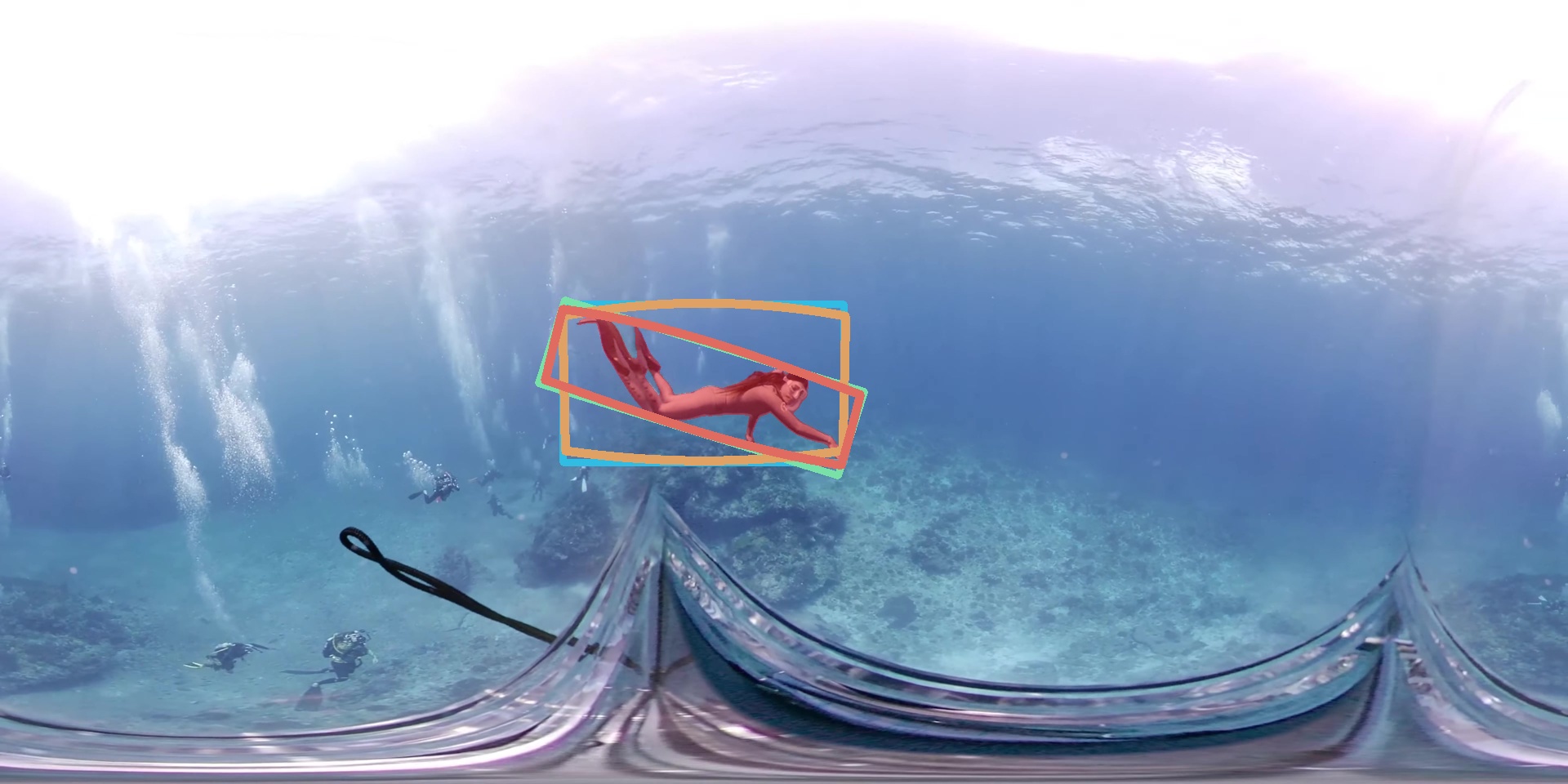}}\!
    \subfloat{\includegraphics[width=\imgw\linewidth, height=\imgh\linewidth]{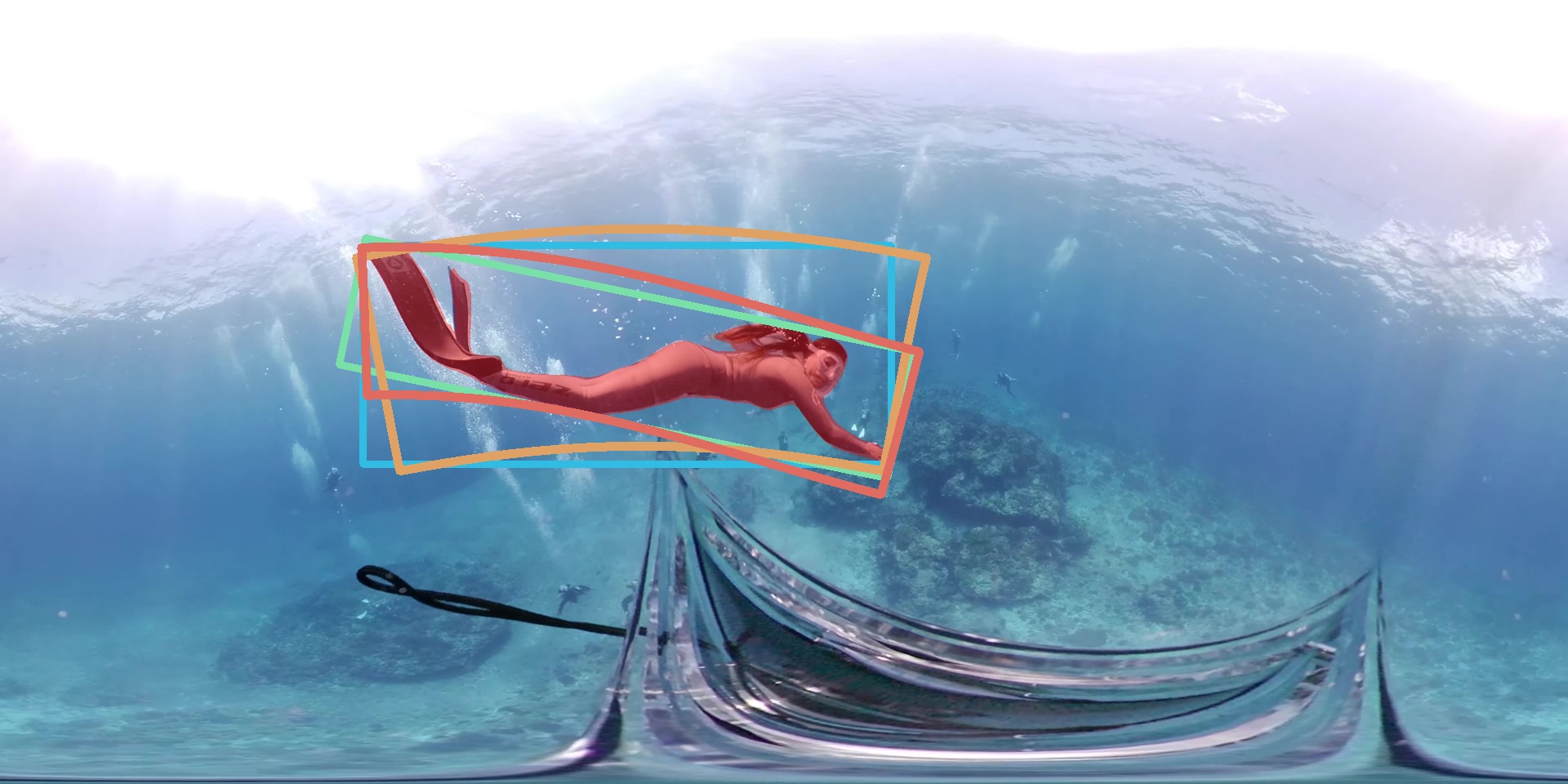}}\!
    \subfloat{\includegraphics[width=\imgw\linewidth, height=\imgh\linewidth]{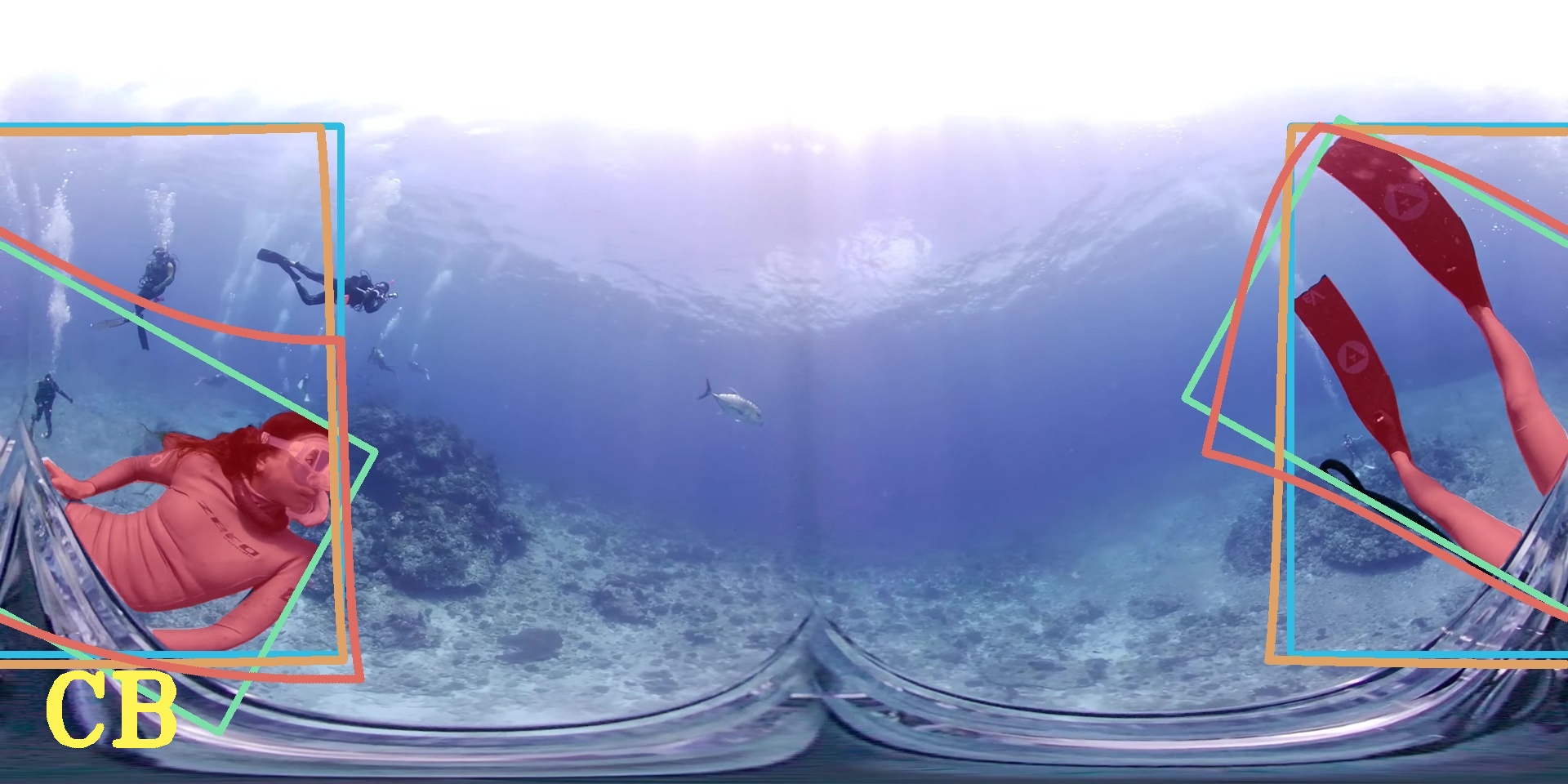}}\!
    \subfloat{\includegraphics[width=\imgw\linewidth, height=\imgh\linewidth]{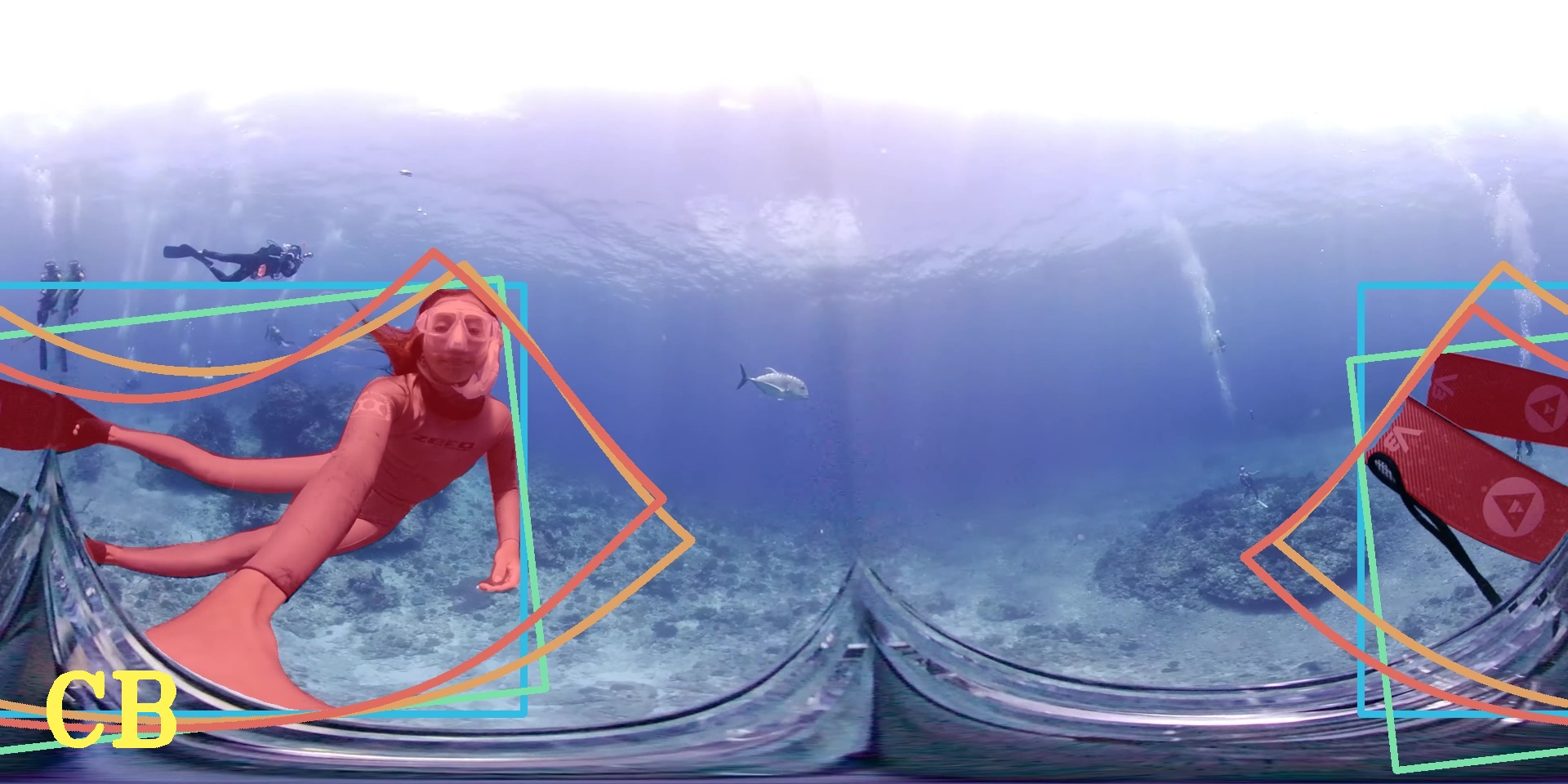}}
    \vspace{-0.8em}    \\
    \subfloat{\includegraphics[width=\imgw\linewidth, height=\imgh\linewidth]{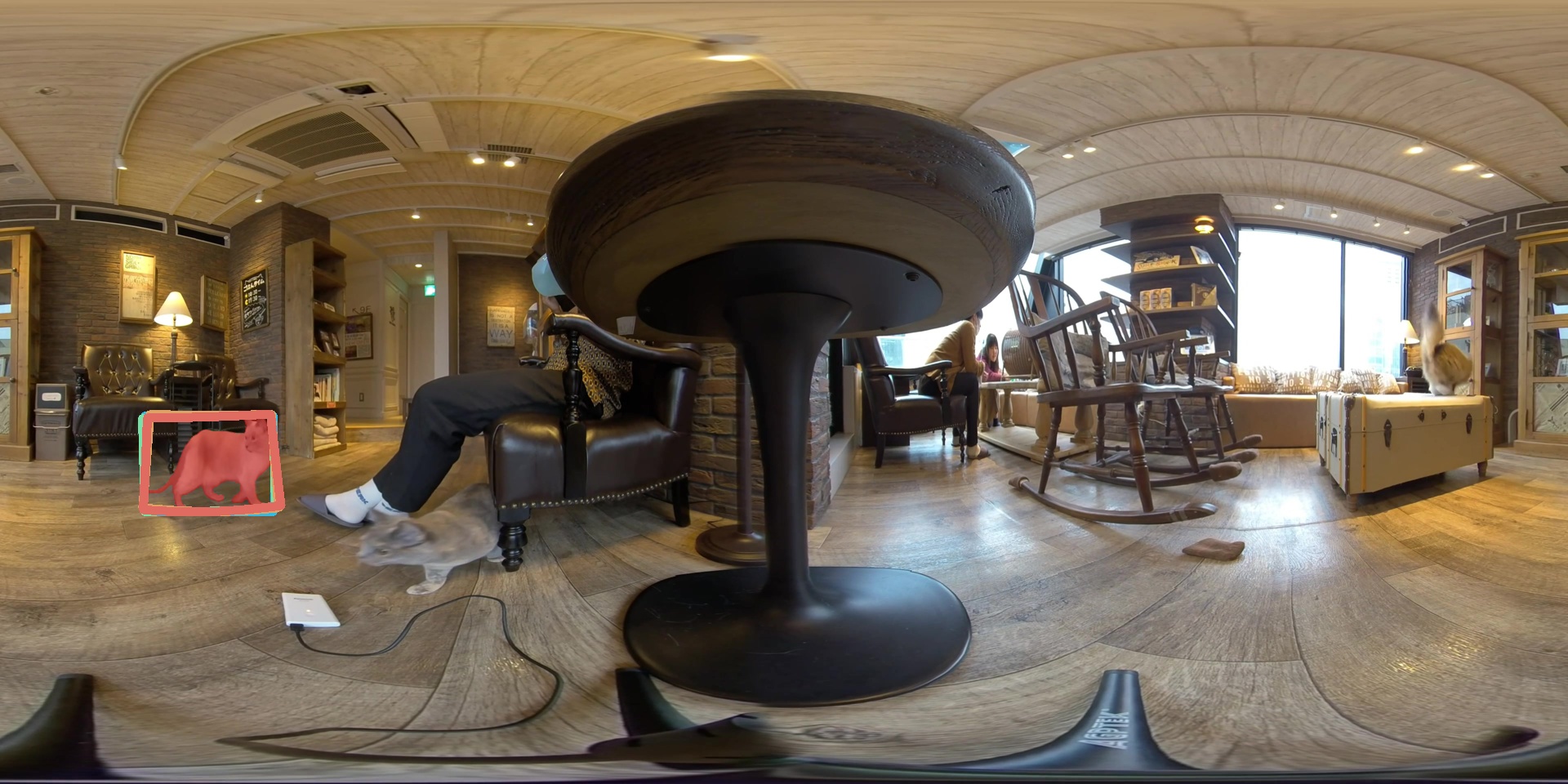}}\!
    \subfloat{\includegraphics[width=\imgw\linewidth, height=\imgh\linewidth]{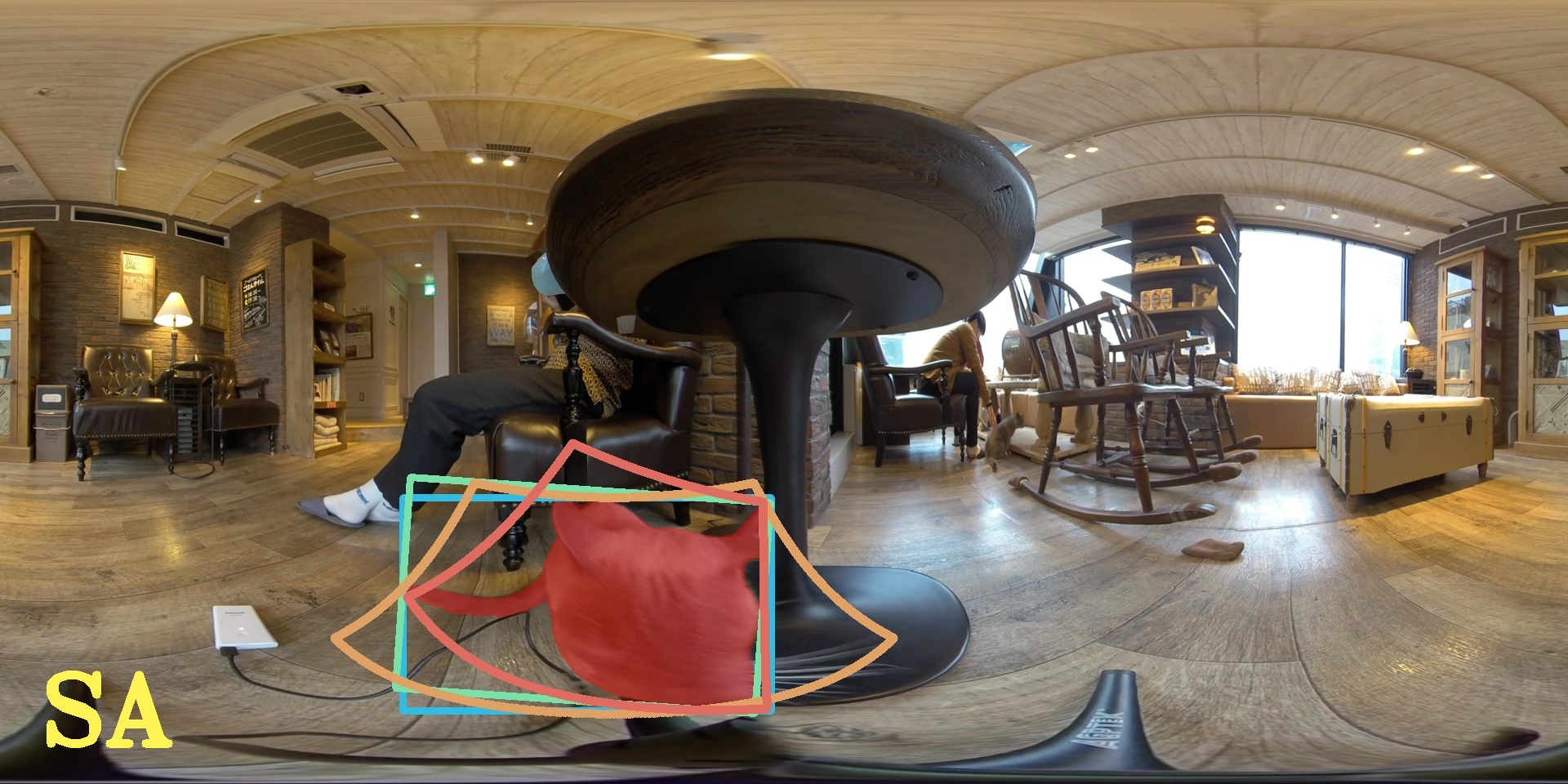}}\!
    \subfloat{\includegraphics[width=\imgw\linewidth, height=\imgh\linewidth]{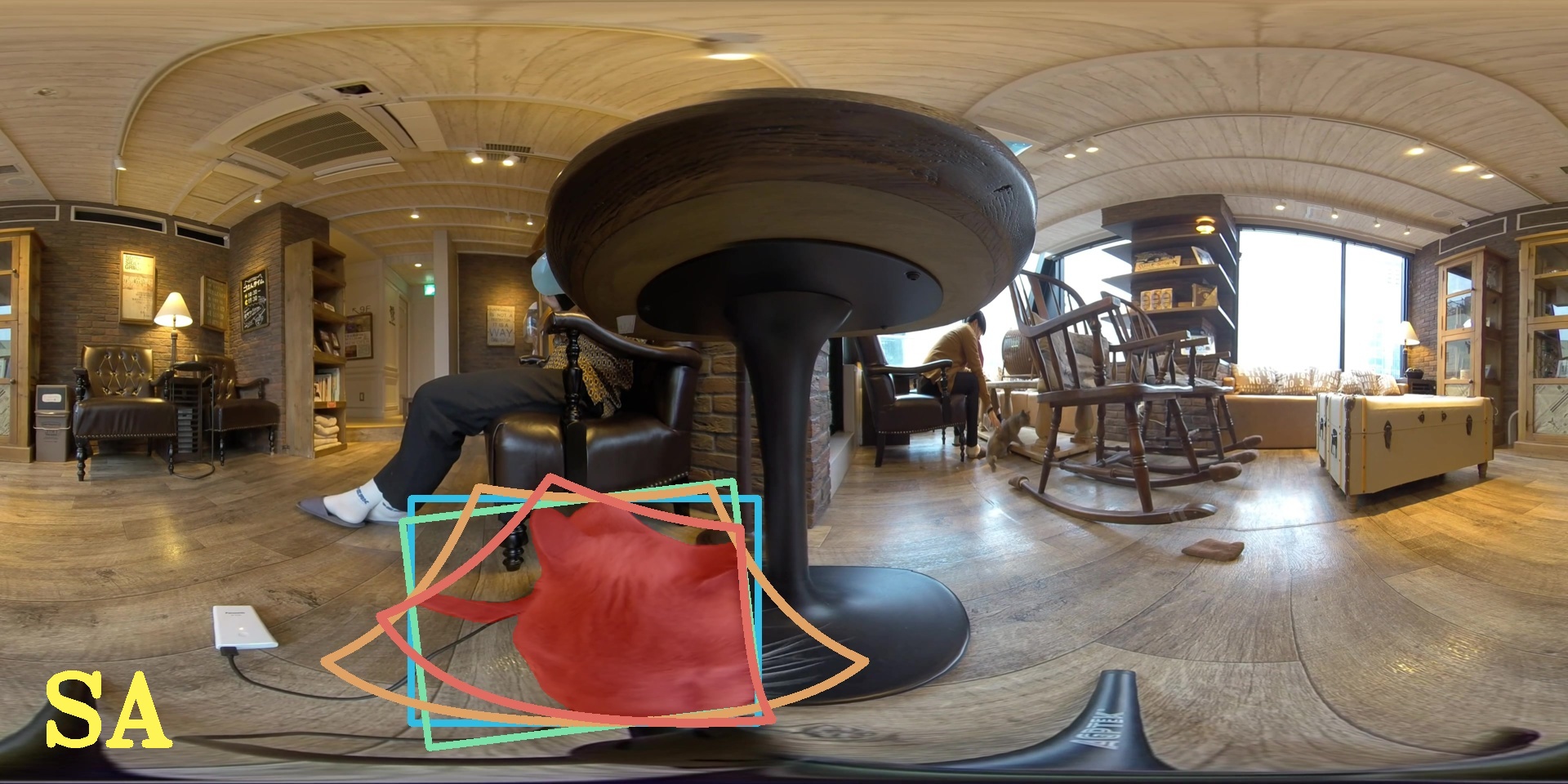}}\!
    \subfloat{\includegraphics[width=\imgw\linewidth, height=\imgh\linewidth]{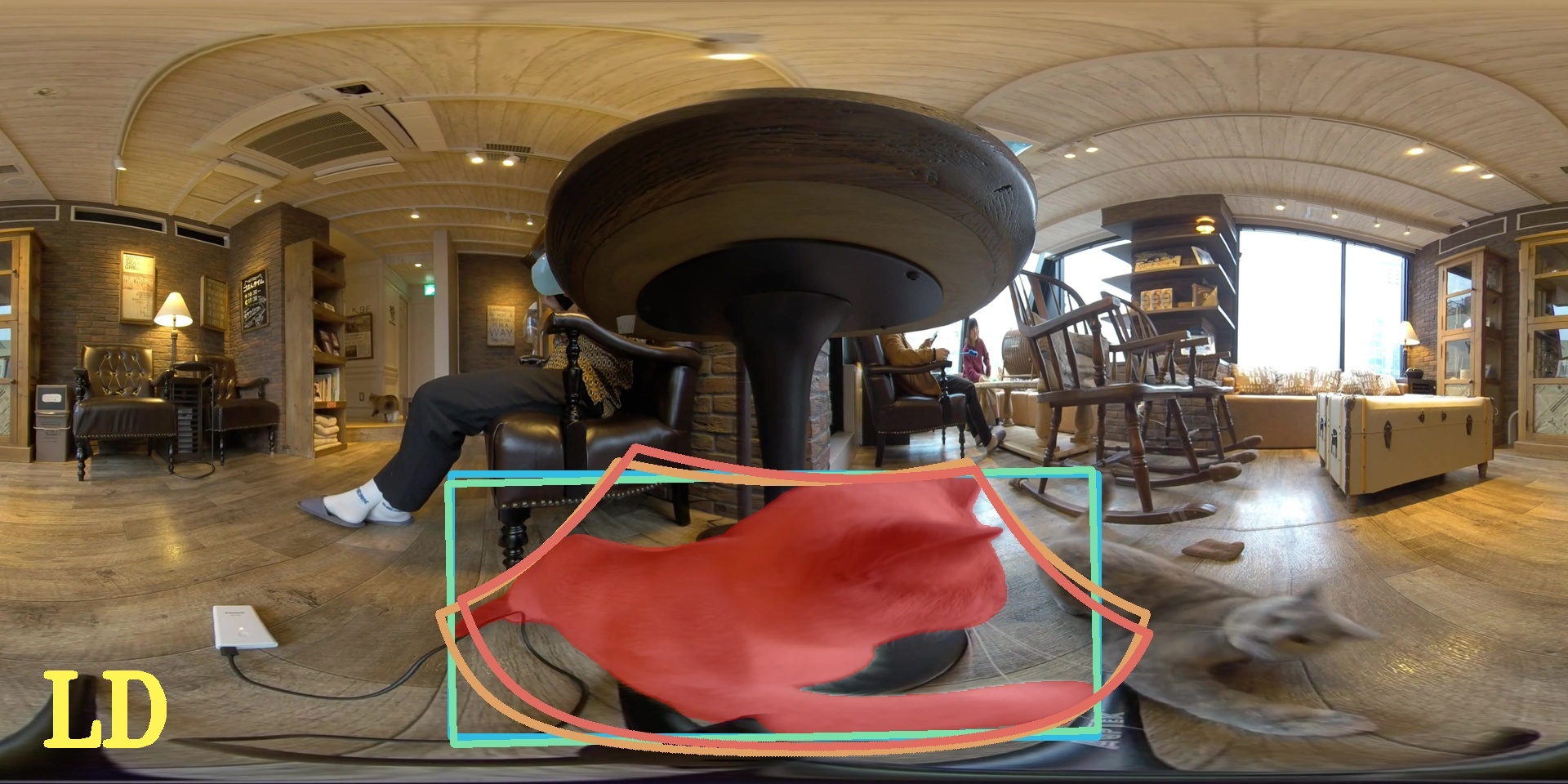}}
    \vspace{-0.8em}\\
    \subfloat{\includegraphics[width=\imgw\linewidth, height=\imgh\linewidth]{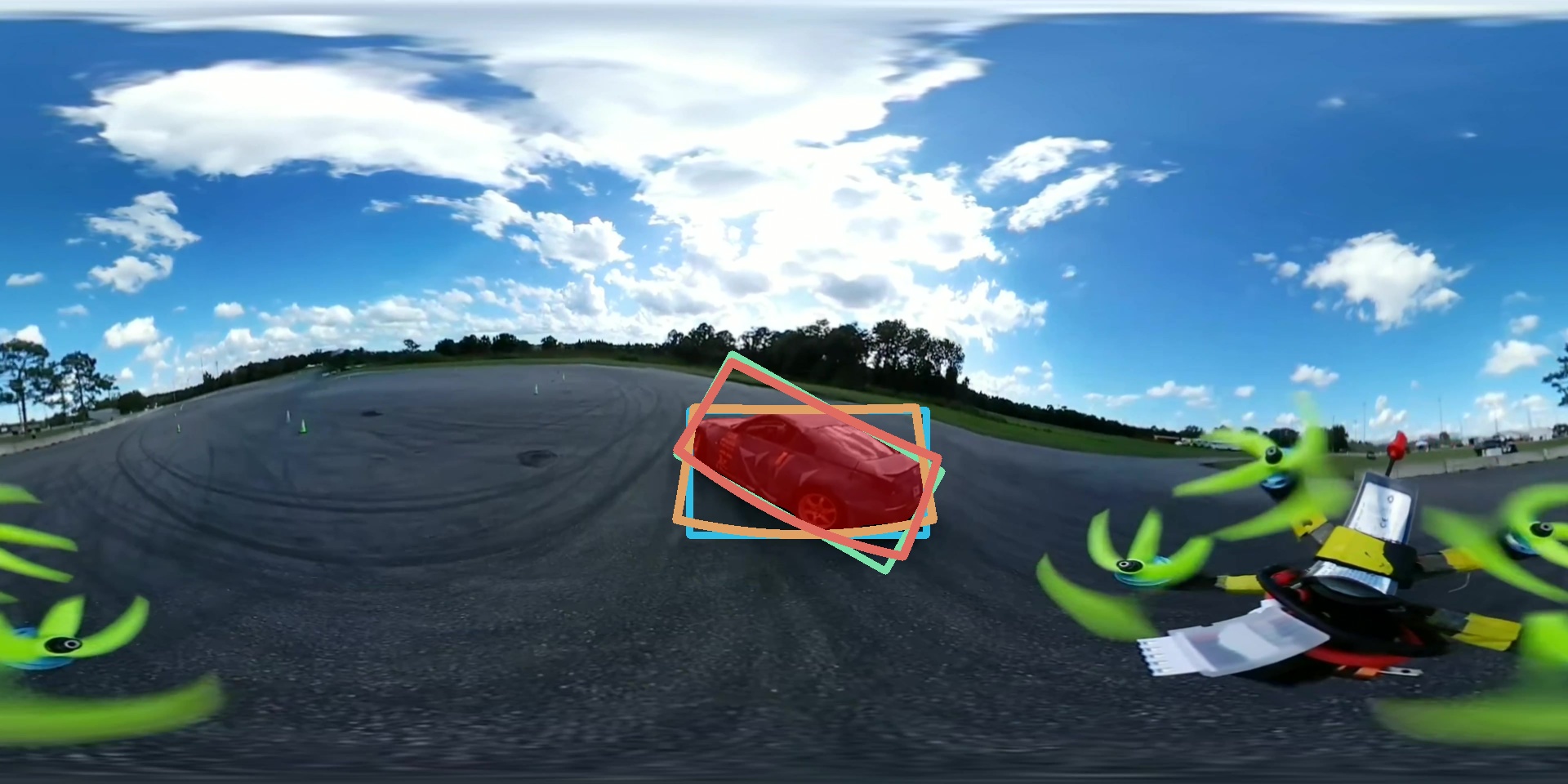}}\!
    \subfloat{\includegraphics[width=\imgw\linewidth, height=\imgh\linewidth]{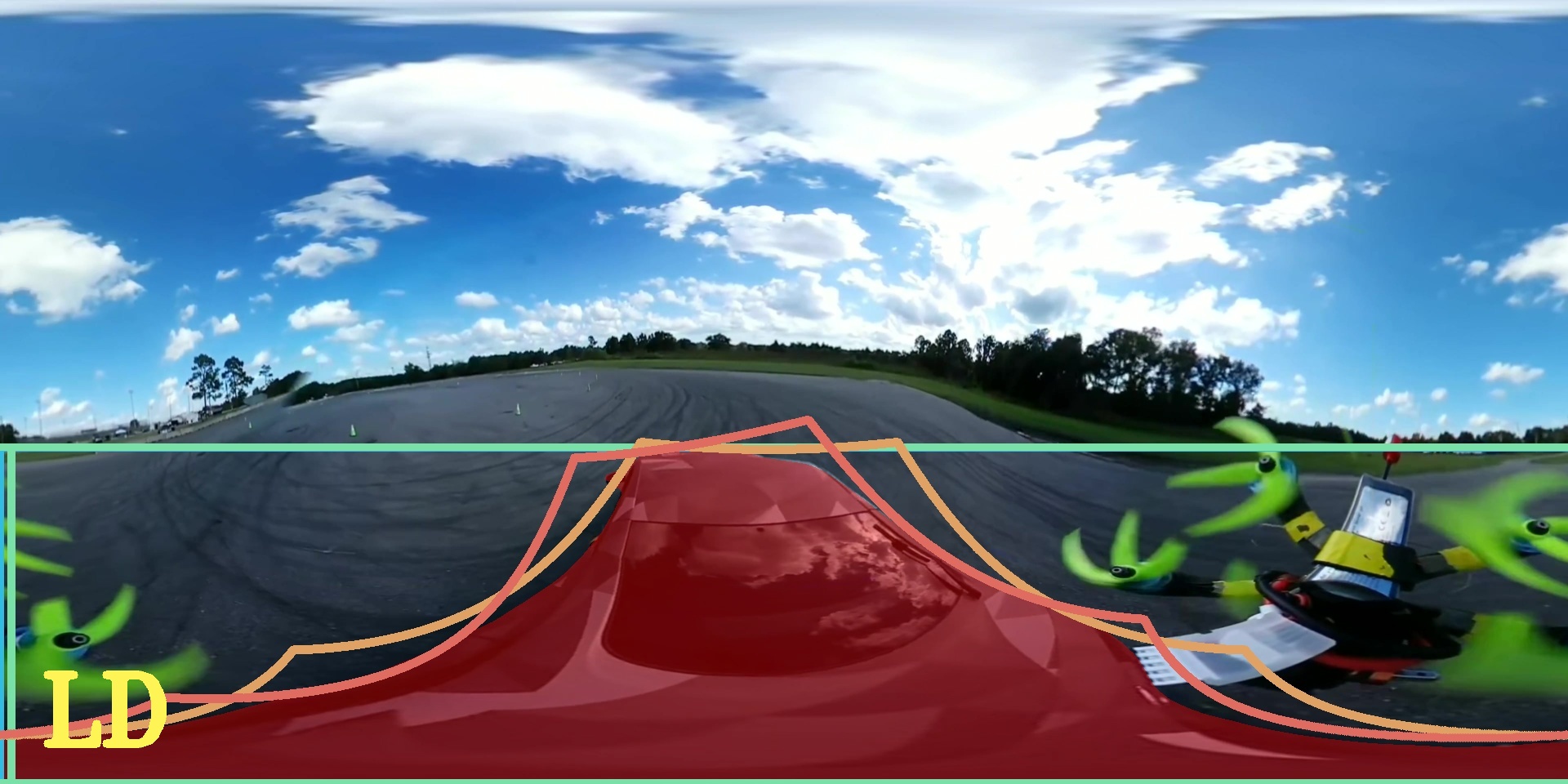}}\!
    \subfloat{\includegraphics[width=\imgw\linewidth, height=\imgh\linewidth]{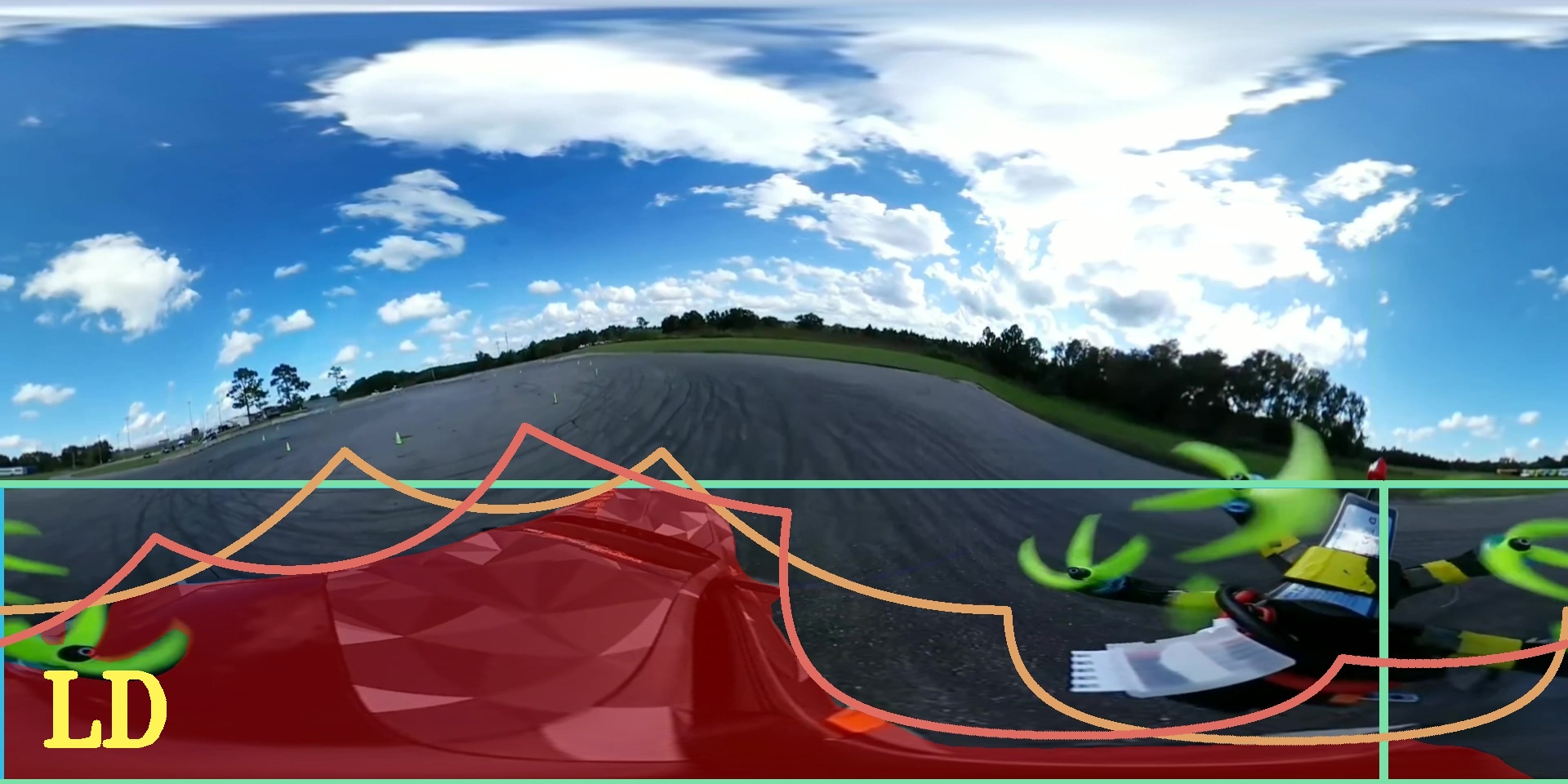}}\!
    \subfloat{\includegraphics[width=\imgw\linewidth, height=\imgh\linewidth]{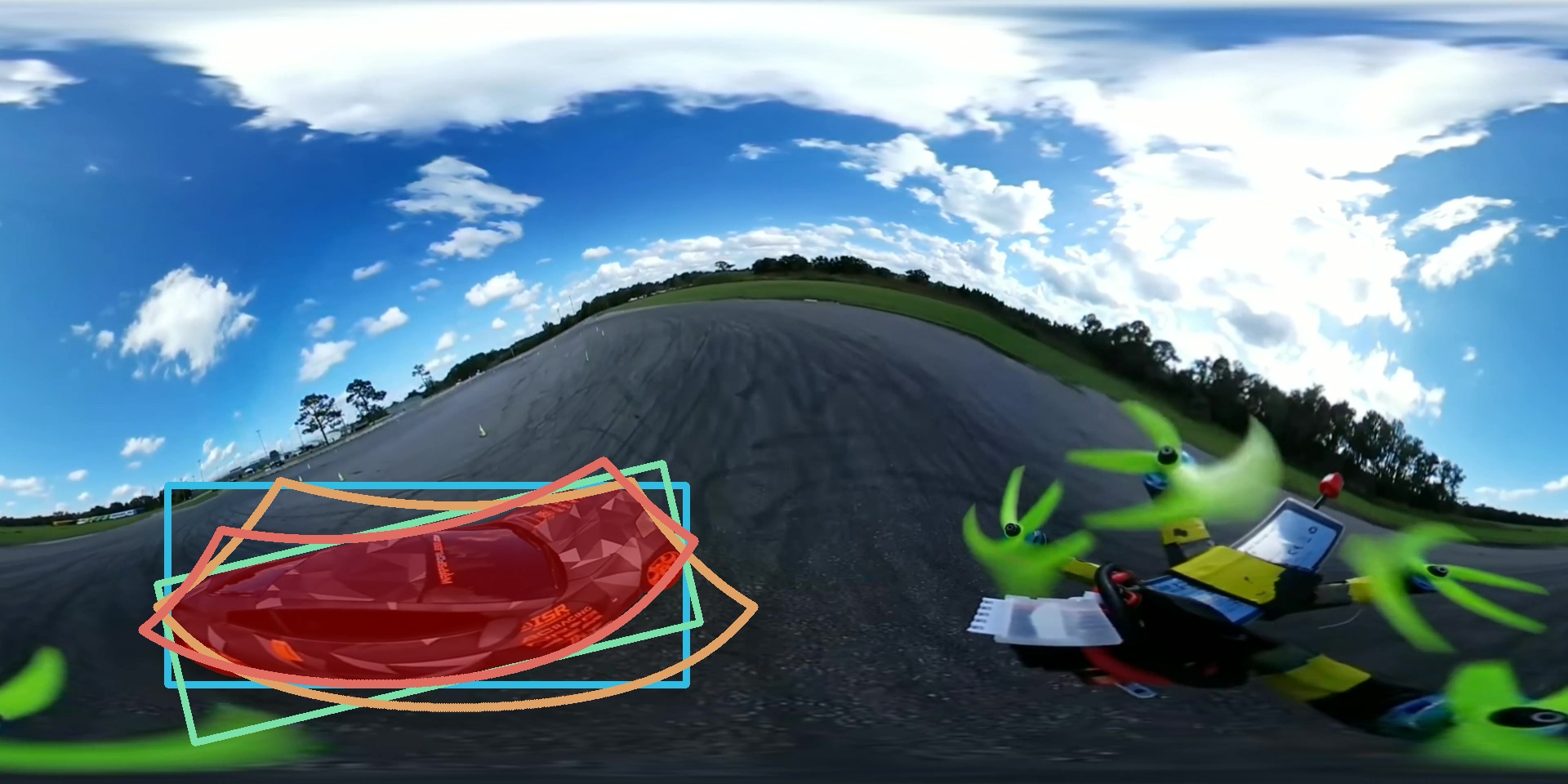}}\\
    
    \caption{Example sequences and annotations of 360VOTS benchmark dataset. The target objects in each 360$\degree$ frame are annotated with \textcolor{red}{mask}s which can be converted into four different representations as ground truth, including bounding box (\textcolor{bbox}{\textbf{BBox}}), rotated bounding box (\textcolor{rbbox}{\textbf{rBBox}}), bounding field-of-view (\textcolor{bfov}{\textbf{BFoV}}), and rotated bounding field-of-view (\textcolor{rbfov}{\textbf{rBFoV}}).
    360VOTS brings distinct challenges for tracking and segmentation, i.e., crossing border (CB), large distortion (LD), and stitching artifact (SA).  
    }\label{fig:teaser}
    \vspace{-0.2cm}
\end{figure*}

Omnidirectional object tracking and segmentation employs a 360$\degree$ camera to track the target object. With its omnidirectional field-of-view (FoV), a 360$\degree$ camera offers continuous observation of the target over a longer period, minimizing the out-of-view issue. This advantage is crucial for intelligent agents to achieve stable, long-term tracking and perception. In general, an ideal spherical camera model is used to describe the projection relationship of a 360$\degree$ camera. The 360$\degree$ image is widely represented by equirectangular projection (ERP)~\cite{erp}, which has two main features: 1) crossing the image border and 2) extreme distortion as the latitude increases. Moreover, due to inherent limitations or manufacturing defects of the camera, the 360$\degree$ image may suffer from stitching artifacts that would blur, break, or duplicate the shape of objects. Meanwhile, omnidirectional FoV means it is inevitable to capture the photographers. They would distract and occlude the targets. These phenomena are illustrated in Figure~\ref{fig:teaser}. Eventually, they bring new challenges to perform object tracking and segmentation in 360$\degree$ videos.

To tackle these challenges, we first explore a new representation in omnidirectional tracking that utilizes extended bounding field-of-view (eBFoV) for precise target localization. Compared to the commonly used bounding box (BBox), BFoV~\cite{uiou, PANDORA} represents object localization on the unit sphere in an angular fashion. This representation takes the spherical nature of 360$\degree$ images into account, providing improved constraints for accurately localizing objects and eliminating dependence on image resolution. We therefore can utilize BFoV to extract less-distorted images for target searching.
To enable proper handling of objects even when they occupy significant portions of the omnidirectional image, we further extend the BFoV definition and propose eBFoV to accommodate the extraction of search regions. 
By leveraging the benefits of eBFoV, we propose a general 360 tracking framework that effectively enhances the performance of the conventional local trackers devised for perspective scenes in omnidirectional scenes. 

Furthermore, we present 360VOTS, a challenging benchmark dataset to facilitate research and evaluation on omnidirectional tracking and segmentation, including 360VOT~\cite{vot360} and 360VOS datasets.
360VOT provides dense (rotated) BBox and BFoV annotations of 120 test sequences.
{Inheriting parts of sequences from the 360VOT dataset, 360VOS incorporates additional 191 sequences and 30 tracking object categories. It contains a total of 290 sequences in 62 categories, while 170 of them are assigned as training set. Importantly, 360VOS provides dense pixel-wise annotations as ground truth that are further utilized to obtain unbiased (rotated) BBoxes and BFoVs for VOT tasks.}
To accurately evaluate the performance of state-of-the-art (SOTA) approaches on omnidirectional tracking and segmentation, we develop new metrics tailored for 360$\degree$ images.
These metrics encompass dual success, dual (normalized) precision, and angle precision to assess tracking performance, as well as spherical region similarity and spherical contour accuracy to evaluate segmentation performance. 
Moreover, we exploit the 360VOS training set to retrain the tracker and apply our 360 tracking framework to achieve new SOTA results.
With the introduction of new representations, datasets, metrics, and baselines, we believe 360VOTS serves as a vital resource to foster development in omnidirectional visual object tracking and segmentation.

In summary, the contribution of this work includes:
\begin{itemize}[topsep=0pt, itemsep=0pt, leftmargin=*]
    \item We explore the new representations for object location in 360$\degree$ images, as well as a general 360 tracking framework.
    \item Our proposed 360VOTS is the first benchmark dataset for both omnidirectional visual object tracking and segmentation. We provide densely annotated masks for segmentation and subsequently convert masks into four types of unbiased ground truth for tracking. 
    \item We propose a set of new metrics specifically designed to enable rigorous performance evaluation of omnidirectional tracking and segmentation algorithms.
    \item We benchmark 24 tracking algorithms and 16 segmentation algorithms on 360VOT and 360VOS respectively, and develop new baselines for future comparisons.    
\end{itemize}

{
\normalsize
\renewcommand{\arraystretch}{1.0}
\begin{table*}[t]
    \captionsetup{labelsep=newline,justification=centering}
    \caption{
    Comparison of popular visual object tracking (VOT) and video object segmentation (VOS) benchmarks.
    }
    \centering
    \centering
    \tabcolsep=2.8pt
\resizebox{\linewidth}{!}{ 
    \begin{tabular}{r||ccccccccccc} 
    \hline 
	Benchmark & \makecell{Videos\\test / train} & \makecell{Total\\frames} &\makecell{Min\\frames} & \makecell{Mean\\frames} & \makecell{Median\\frames} & \makecell{Max\\frames} & \makecell{Object\\classes} &Attr. &Annotation &Feature &Year \\
    \hline \hline
    {ALOV300}\cite{alov300}&314&152K&19&483&276&5,975&64&14&{sparse BBox}&diverse scenes&2013\\
    {OTB100}\cite{otb100}&100&81K&71&590&393&3,872&16&11&{dense BBox}&short-term&2015\\
    {NUS-PRO}\cite{nus-pro}&365&135K&146&371&300&5,040&8&12&{dense BBox}&occlusion-level&2015\\
    {TC128}\cite{tc128}&129&55K&71&429&365&3,872&27&11&{dense BBox}&{color enhanced}&2015\\
    {UAV123}\cite{uav123}&123&113K&109&915&882&3,085&9&12&{dense BBox}&UAV&2016\\
    {DTB70}\cite{dtb70}&70&16K&68&225&202&699&29&11&{dense BBox}&UAV&2016\\
    {NfS}\cite{nfs}&100&383K&169&3,830&2,448&20,665&17&9&{dense BBox}&high FPS&2017\\
    {UAVDT}\cite{uavdt}&100&78K&82&778&602&2,969&27&14&{sparse BBox}&UAV&2017\\
    
    {OxUvA}\cite{OxUvA}&337&1.55M&900&4,260&2,628&37,440&22&6&{sparse BBox}&long-term&2018\\    
    {TrackingNet}\cite{trackingnet}&511/30K&226K/14M&96/87&441/471&390/480&2,368/690&27&15&{sparse BBox}&large scale&2018\\
    
    {LaSOT}\cite{lasot}&280/1,120&685K/2,816K&1,000/898&2,448/2,514&2,102/2,023&9,999/11K&85&14&{dense BBox}&{category balance}&2018\\
    {GOT-10k}\cite{got10k}&420/9,335&56K/1,403K&29/29&127/150&100/101&920/1,481&84&6&{dense BBox}&generic&2019\\
    {TOTB}\cite{totb}&225&86K&126&381&389&500&15&12&{dense BBox} &transparent &2021\\
    {TREK-150}\cite{trek150}&150&97K&161&649&484&4,640&34&17&{dense BBox}&FPV&2021\\
    {VOT}\cite{VOT}&62&20K&41&321&242&1,500&37&9&{dense BBox}&annual&2022\\
    \hline 
    {\textbf{360VOT}}\cite{vot360}&120&113K&251&940&775&2,400&32&20&\makecell{dense (r)BBox\\\& (r)BFoV}&360$\degree$ images&2023\\    
    \hline \hline
    {FBMS-59}\cite{fbms}&30/29&7,306/6,554&19/19&244/226&210/193&800/720&-&-&{sparse Mask}&moving objects&2014\\
    {DAVIS}\cite{davis2017}&30/90&2,860/6,208&31/25&70/69&70/70&127/104&376&15&{dense Mask}&dense annotated&2017\\
    {YouTube-VOS}\cite{ytvos2018}&508/3471&70K/470K&36/20&139/135&150/150&180/180&94&-&{sparse Mask}&comprehensive&2018\\
    {LVOS}\cite{lvos}&50/120&30K/65K&257/200&598/544&489/414&2,000/2,280&27&13&{dense Mask}&long-term&2023\\
    \hline
    \textbf{360VOS}&120/170&98K/144K&111/62&821/845&684/714&2,400/2,402&62&20&\makecell{dense Mask}&360$\degree$ images &2024\\
    \hline
    \end{tabular}
}
    \label{tab:benchmark}
\end{table*}
}

\section{Related Work}\label{sec2}
\subsection{Benchmarks for Visual Object Tracking}
With the remarkable development of the visual object tracking community, previous works have proposed numerous benchmarks in various scenarios. 

{ALOV300~\cite{alov300} is a sparse benchmark introducing 152K frames and 16K annotations.
TrackingNet~\cite{trackingnet} is a large-scale sparse dataset collecting more than 14M frames based on the YT-BB dataset~\cite{YT-BB}.
OxUvA~\cite{OxUvA} targets long-term tracking by constructing 337 video sequences.}

One of the first dense BBox benchmarks is OTB100~\cite{otb100} which is extended from OTB50~\cite{otb50} and has 100 sequences. NUS-PRO~\cite{nus-pro} takes the feature of occlusion-level annotation and provides 365 sequences, while TC128~\cite{tc128} researches the chromatic information in visual tracking.
{UAV123~\cite{uav123}, DTB70~\cite{dtb70}, and UAVDT~\cite{uavdt} offer hundreds of aerial videos containing rigid objects and humans in various scenes.} NfS~\cite{nfs} consists of more than 380K frames captured at 240 FPS studying higher frame rate tracking, while LaSOT~\cite{lasot} is a large-scale and category-balanced benchmark of premium quality. GOT-10k~\cite{got10k} provides about 1.5M annotations and 84 classes of objects, aiming at generic object tracking. The annual tracking challenge VOT~\cite{VOT} offered 62 sequences and 20K frames in 2022. 
{Under specific conditions, TOTB~\cite{totb} focuses on transparent object tracking, while TREK-150~\cite{trek150} introduces First Person Vision (FPV) videos that interact with the target object.}

Differently, our 360VOT focuses on object tracking and explores new representations on omnidirectional videos.  
A summarized comparison with existing VOT benchmarks is reported in the upper part of Table~\ref{tab:benchmark}.

\subsection{Benchmarks for Video Object Segmentation}
There are several benchmarks for video object segmentation proposed in the last decade as described in Table~\ref{tab:benchmark}. 
FBMS-59\cite{fbms} acknowledges the importance of motion information and concentrates on motion segmentation, providing a total of 13,860 frames across 59 sequences. 
DAVIS\cite{davis2017} is the first dataset designed for video object segmentation (VOS), annotating 376 classes of objects on each frame of 120 short videos. 
{Leveraging the diverse scenarios and photography conditions on YouTube videos, YouTube-VOS\cite{ytvos2018} introduces a large-scale VOS dataset with more than 540K frames and 3471 training sequences.}
One of the most recent datasets is LVOS\cite{lvos} which is the first long-term VOS benchmark dataset with an average length of 560 densely labeled frames.  In contrast, our 360VOS provides pixel-wise masks of omnidirectional images for both training and testing.

\subsection{Other Omnidirectional Vision Benchmarks}\label{sec2c}
{
In addition to the video-based benchmarks, other related benchmarks such as 360-Indoor \cite{indoor360} and PANDORA \cite{PANDORA} are proposed for detection tasks on static 360$\degree$ images. Meaningful progress has been made by introducing innovative network architecture \cite{FlyingCars} to address the challenge of large distortions, along with tailored solutions for specific real-world applications \cite{omnipede}.
}
{
Like omnidirectional data, panoramic data covers a wide FoV, usually up to 360$\degree$ horizontally. 2D-3D-S \cite{stanford2d3d} and SynPASS \cite{trans4PASSp} are synthetic panoramic benchmarks containing over 70K and 9K RGB images with semantic annotations. Existing methods \cite{trans4PASS, trans4PASSp, DPPASS} take advantage of distortion awareness and achieve large advancements in the panoramic semantic segmentation task. Our 360VOS dataset includes numerous high-resolution frames and annotated object-based masks, demonstrating its potential for related tasks through further processing and extension.
}

\subsection{Adapting Object Tracking Schemes in Segmentation}
To guarantee high tracking speed, the trackers for single object tracking generally crop the image and search for the target in small local regions. 
{The tracking scheme is vital in selecting searching regions and interpreting network predictions over sequences, which enhances tracking performance in the inference phase.}
For example, DaSiamRPN~\cite{desiamrpn} explored a local-to-global searching strategy for long-term tracking. SiamX~\cite{siamx2022} proposed an adaptive inference scheme to prevent tracking loss and realized fast target re-localization. 
Similarly, localizing the target objects is also vital for VOS tasks. Both RTS\cite{rts} and SiamMask\cite{siammask} demonstrated that pixel-wise masks can provide a more accurate estimation of bounding boxes for VOT tasks. Additionally, their tracking schemes improve the robustness of VOS methods and contribute to target localization in long-term videos. 

{In 360VOTS, we utilize eBFoV to localize the targets on 360$\degree$ videos and obtain less-distorted search regions. It allows us to adopt arbitrary local visual trackers trained by perspective images on 360$\degree$ object tracking and segmentation.}

\begin{figure}
    \centering
    \includegraphics[width=0.99\linewidth]{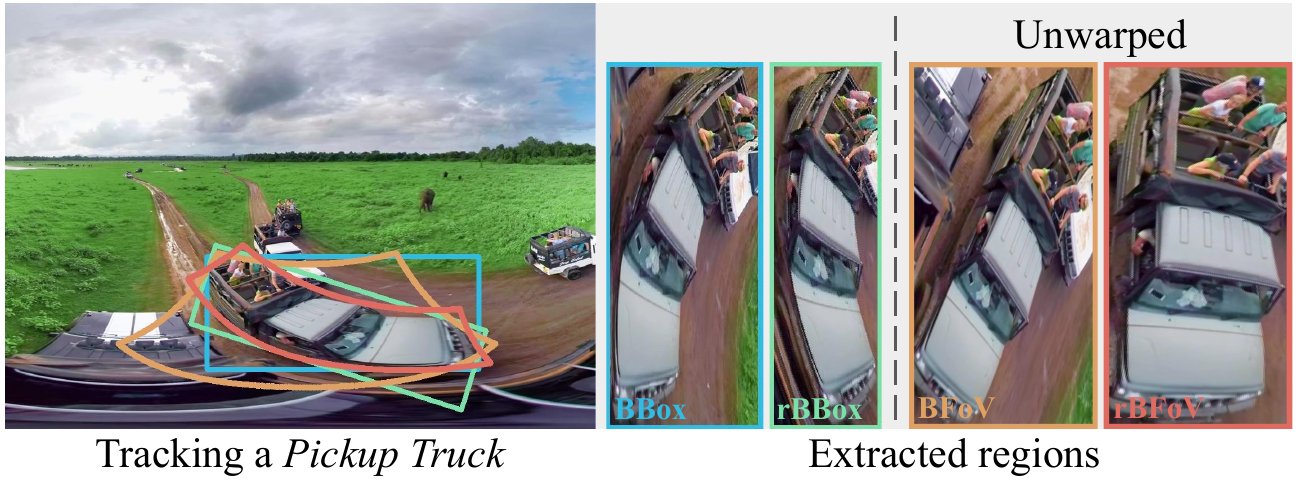}
    \caption{\textit{Pickup Truck}. The comparison of the bounding regions of different representations on a 360$\degree$ image. The unwarped images based on BFoV and rBFoV are less distorted. 
    }
    \label{fig:representation}
\end{figure}

\section{Tracking on 360$\degree$ Video}
The 360$\degree$ video is composed of frames using the most common ERP. 
Each frame can capture 360$\degree$ horizontal and 180$\degree$ vertical field of view. Although omnidirectional FoV avoids out-of-view issues, the target may cross the left and right borders of a 2D image. 
Additionally, nonlinear projection distortion makes the target largely distorted when they are near the top or bottom of the image, as illustrated in Figure~\ref{fig:teaser}.
Therefore, a new representation and framework that fits the 360$\degree$ video for omnidirectional visual tracking is necessary.

 \begin{figure*}
    \includegraphics[width=0.99\linewidth]{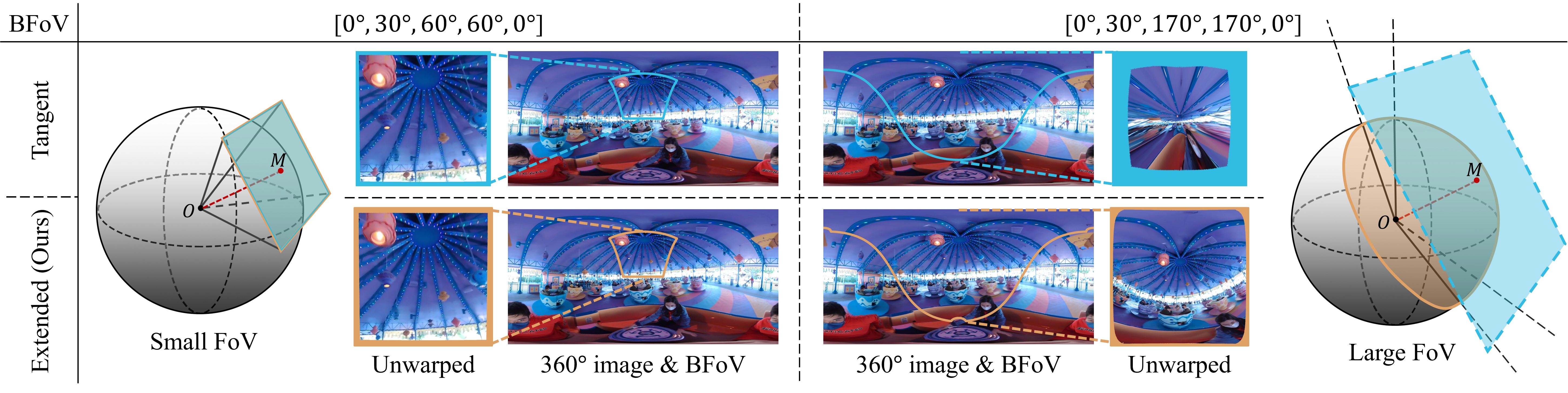}
    \caption{The boundaries on the 360$\degree$ images and the corresponding unwarped images of different BFoV definitions. The tangent BFoV is displayed in \textcolor{cyan}{blue} and the extended BFoV is in \textcolor{orange}{orange}. $M$ on the sphere surface denotes the object center and tangent point. \textcolor{cyan}{Blue} plane with dotted borders represents a larger plane out of space. Best viewed in color.}
    \label{fig:bfov}
\end{figure*}

\subsection{Representation for the Target Location}\label{seq:representation}
The (r)BBox is the most common and simple way to represent the target object's position in perspective images. It is a rectangular area defined by the rectangle around the target object on the image and denoted as $[cx, cy, w, h, \gamma]$, where $cx, cy$ are object center, $w, h$ are width and height. The rotation angle $\gamma$ of BBox is always zero.  However, these representations would become less accurate to properly constrain the target on the 360$\degree$ image, as depicted in Figure \ref{fig:representation}. The works~\cite{spherical_criteria, PANDORA} for 360$\degree$ object detection show that the BFoV and rBFoV are more appropriate representations on 360$\degree$ images. Basically, we can use the spherical camera model $\mathcal{P}$ to formulate the mathematical relationship between the 2D image in ERP and a continuous 3D unit sphere~\cite{360vo}. 
(r)BFoV is then defined as $[clon, clat, \theta, \phi, \gamma]$ where $[clon, clat]$ are the longitude and latitude coordinates of the object center at the spherical coordinate system, $\theta$ and $\phi$ denote the maximum horizontal and vertical field-of-view angles of the object’s occupation. 
The represented region of (r)BFoV on the 360$\degree$ image is commonly calculated via a tangent plane~\cite{spherical_criteria, PANDORA}, $T(\theta, \phi)\in\mathbb{R}^3$, and formulated as:
\begin{equation}
    I(\mbox{\footnotesize(r)BFoV} \,|\, \Omega) = \mathcal{P}(\mathcal{R}_y(clon)\cdot\mathcal{R}_x(clat)\cdot\mathcal{R}_z(\gamma)\cdot \Omega) ,
\end{equation}
where {$\mathcal{R}$ denotes the 3D rotation along the $y$, $x$, $z$ axes,}
\begin{equation}
    \Omega = T = \begin{bmatrix} \mathbf{X}\\\mathbf{Y}\\\mathbf{Z}\end{bmatrix} =\begin{bmatrix}-tan(\theta/2):tan(\theta/2)\\-tan(\phi/2):tan(\phi/2)\\1\end{bmatrix}.
\end{equation}
The projection function $\mathcal{P}$ to depict the relationship between the 3D camera space $[X, Y, Z]$ and the 2D image space $[u, v]$ is formulated as:
\begin{equation} \label{eq:xyz2uv}
\begin{aligned}
    u &= (\frac{lon}{2\pi} + 0.5) * W = arctan(X/Z) ,\\
    v &= (-\frac{lat}{\pi}+0.5) * H = arctan(\frac{-Y}{\sqrt{X^2+Z^2}}) ,
\end{aligned}
\end{equation}
where $-\pi<lon<\pi$ and $-\pi/2<lat<\pi/2$ denote the longitude and latitude in the spherical coordinate system. W and H are the width and height of the 360$\degree$ image. The unwarped images based on tangent BFoV are distortion-free under the small FoV, as shown in Figure \ref{fig:representation}.

However, this definition has a disadvantage on large FoV and cannot represent the region exceeding 180$\degree$ FoV essentially. With the increasing FoV, the unwarped images from the tangent planes have drastic distortions, shown in the upper row in Figure~\ref{fig:bfov}. This defect limits the application of BFoV on visual object tracking since trackers rely on unwarped images for target searching.  
To address this problem, we extended the definition of BFoV. When the bounding region involves a large FoV, i.e., larger than 90$\degree$, 
the extended BFoV leverages a spherical surface  $S(\theta, \phi)\in\mathbb{R}^3$ instead of a tangent plane to represent the bounding region on the 360$\degree$ image:
\begin{equation}
    S=\begin{bmatrix}cos(\Phi)sin(\Theta) & -sin(\Phi) & cos(\Phi)cos(\Theta)\end{bmatrix}^T,\\
\end{equation}
where $ \Phi\in[-\phi/2, \phi/2], \Theta\in[-\theta/2, \theta/2]$.
Therefore, the corresponding region of extended (r)BFoV on 360$\degree$ image is 
\begin{equation}
    I(\mbox{\footnotesize(r)BFoV} \,|\, \Omega), \quad  \Omega = \begin{cases}
    T(\theta, \phi), &\theta<90\degree, \phi<90\degree\\
    S(\theta, \phi), &otherwise
    \end{cases} .\label{eq:ebfov}
\end{equation}
The comparisons of the boundary on 360$\degree$ images and corresponding unwarped images based on tangent BFoV and the extended BFoV are shown in Figure~\ref{fig:bfov}. For simplicity, BFoV in the following content denotes extended BFoV unless explicitly stated otherwise.

\subsection{360 Tracking Framework}
\label{sec:framework}

\begin{figure*}[!ht]
    \centering
    \includegraphics[width=0.99\linewidth]{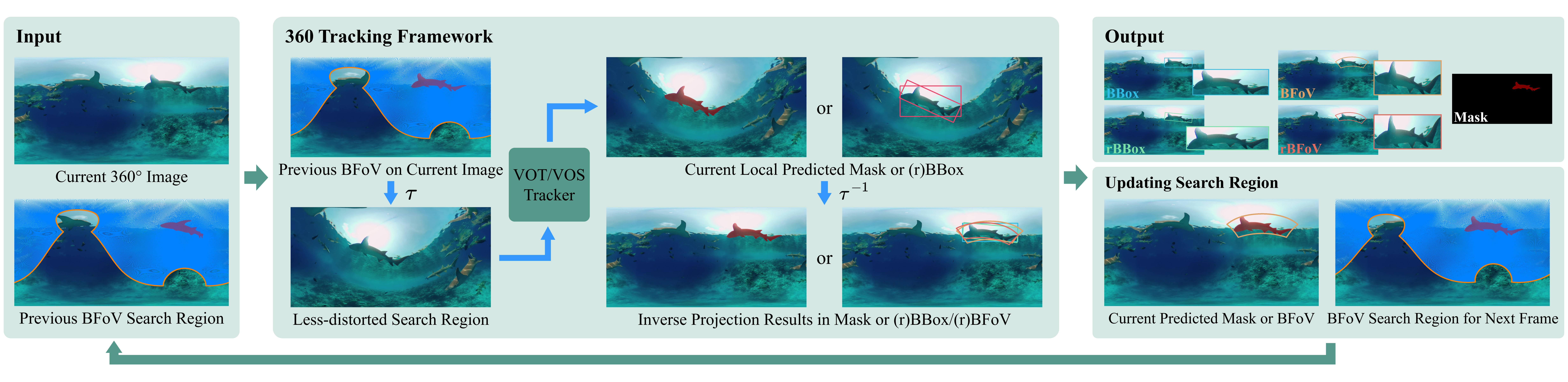}
    \caption{The 360 tracking framework. The framework focuses on extracting a less-distorted search region to enable precise local visual tracking.
    The general framework accommodates arbitrary local visual trackers to generate masks or 4 types of representation. $\tau$ is the remapping operation to extract the less-distorted search region from the input 360$\degree$ image, while $\tau^{-1}$ is its inverse operation.
    }
    \label{fig:360tracking}
\end{figure*}

To conduct omnidirectional tracking and segmentation using an existing local VOT or VOS tracker, we propose a 360 tracking framework, as depicted in Figure~\ref{fig:360tracking}. The framework leverages extended BFoV to address challenges caused by object crossing-border and large distortion on the 360$\degree$ image.

Given an initial BFoV or mask, the framework first calculates the corresponding search region $I$ on the 360$\degree$ image via Eq.~\ref{eq:ebfov}.
Subsequently, by remapping ($\tau$) the 360$\degree$ image using corresponding pixel coordinates recorded in $I$, it extracts a less distorted local image for target identification. Within the extracted image, a local visual tracker then infers a (r)BBox or a segmentation mask. 
Finally, we utilize $I$ to convert the local prediction back to the global bounding region on the 360$\degree$ image via inverse projection $\tau^{-1}$. 
For (r)BBox prediction, we calculate the minimum area (rotated) rectangle on the 360$\degree$ image. For (r)BFoV, we re-project the bounding region's coordinates onto the spherical coordinate system and calculate the maximum bounding FoV. 
As for the final mask in 360$\degree$ images, we dilate the reprojected mask to avoid erosion artifacts resulting from inverse projection. 
To maintain continuous tracking, the BFoV search region for the subsequent frame is updated by the current predicted BFoV or mask. In this manner, the search region follows the motions of the target throughout the entire sequence to achieve object tracking and segmentation. 
It is noted that the framework does not rely on nor affect the network architecture of the tracker, allowing for the adoption of an arbitrary local visual tracker for omnidirectional tracking and segmentation.

{
The 360 tracking framework comprises three essential hyper-parameters: SR Ratio, SR Min, and Max Loss. When utilizing the previously predicted target mask, we extend the search region in the subsequent frame by a scaling factor known as SR Ratio. Moreover, SR Min regulates the minimum BFoV of the search region, preventing the system from becoming fixated on excessively small search areas, particularly when the tracking target is tiny.
Once the target is lost, the framework maintains the search region from the last successful frame for a fixed number of consecutive frames (i.e., Max Loss), allowing the tracker to relocate the target within the same area. If the target remains lost beyond this threshold, the search region is progressively expanded to account for the potential displacement. This expansion continues until the search region either covers the entire 360$\degree$ or a predefined limit on consecutive frames. If the target remains undetected beyond this stage, the framework initiates a full-frame search, ensuring recovery in extreme cases. Once the target is re-tracked, the search region is dynamically updated based on the updated target BFoV, allowing the framework to seamlessly resume normal tracking operations.
}


\section{Benchmark Dataset: 360VOTS}
\label{sec:benchmark}
\subsection{Collection and Annotation}\label{sec4a}
The resources of 360$\degree$ videos in 360VOTS were collected from YouTube or captured using a 360$\degree$ camera. We ranked and filtered hundreds of candidate videos considering tracking difficulty scale and some additional challenging cases. Videos with additional challenges were assigned a higher priority (i.e. distinguishing targets from other highly comparable objects). 
Then, videos were further selected and sampled into sequences with a frame number threshold ($\leq$2400). The relatively stationary frames were further discarded manually. All selected frames were annotated by an interactive segmentation tool~\cite{vot360}.
Considering distribution balance, 120 sequences are finally selected as the 360VOT benchmark and 290 sequences as the 360VOS dataset, sharing 99 sequences for tracking and segmentation benchmarks.

\begin{table}[t]
    \captionsetup{labelsep=newline,justification=centering}
    \caption{
    Attribute description. The 360VOTS contains 13 basic attributes widely used in the existing benchmarks, and 7 distinct challenging attributes as described in the last block of the row.}
    \label{tab:attibute}
    \centering\footnotesize
    \tabcolsep=0.06cm 
    \begin{tabular}{c|m{0.85\linewidth}}
    \hline
    Attr. & Meaning\\
    \hline
    IV & The target is subject to light variation.\\ 
    BC & The background has a similar appearance as the target.\\
    DEF & The target deforms during tracking.\\
    MB & The target is blurred due to motion.\\
    CM & The camera has abrupt motion.\\
    ROT & The target rotates related to the frames. \\ 
    POC & The target is partially occluded.\\
    FOC & The target is fully occluded.\\
    ARC & The ratio of the annotation aspect ratio of the first and the current frame is outside the range [0.5, 2].\\
    SV & The ratio of the annotation area of the first and the current frame is outside the range [0.5, 2].\\
    FM & The motion of the target center between contiguous frames exceeds its own size. \\ 
    LR & The area of the target annotation is less than $1000$ pixels.\\ 
    HR & The area of the target annotation is larger than $500^2$ pixels.\\
    \hline\hline
    SA & The images have stitching artifacts affecting the target object.\\
    CB & The target is crossing the border of the frame and partially appears on the other side.\\
    FMS & The motion angle on the spherical surface of the target center is larger than the last BFoV.\\
    LFoV & The vertical or horizontal FoV of the BFoV is larger than $90\degree$.\\
    LV & The range of the latitude of the target center across the video is larger than $50\degree$.\\
    HL & The latitude of the target center is outside the range $[-60\degree, 60\degree]$, lying in the ``frigid zone".\\
    LD & The target suffers large distortion due to equirectangular projection.\\
    \hline
    \end{tabular}
\end{table}

\subsection{Attribute}\label{sec4b}
Each sequence is annotated with a total of 20 different attributes: illumination variation (IV), background clutter (BC), deformable target (DEF), motion blur (MB), camera motion (CM), rotation (ROT), partial occlusion (POC), full occlusion (FOC), aspect ratio change (ARC), scale variation (SV), fast motion (FM), low resolution (LR), high resolution (HR), stitching artifact (SA), crossing border (CB), fast motion on the sphere (FMS), large FoV (LFoV), latitude variation (LV), high latitude (HL) and large distortion (LD). 

The detailed meaning of each attribute is described in Table~\ref{tab:attibute}. Among them, IV, BC, DEF, MB, CM, ROT, POC, {SA,} and LD attributes are manually labeled, while the others are computed from the annotation results of targets. The distinct features of the 360$\degree$ image are well represented in 360VOTS: \textit{location variations} (FMS, LFoV, and LV), \textit{external disturbances} (SA and LD) and \textit{special imaging} (CB and HL).

\begin{figure*}[t]
    \centering
    \hspace{-0.6cm}
    \captionsetup[subfloat]{labelfont={rm,footnotesize},textfont=footnotesize}
    \subfloat[Attribute histograms.\label{fig:attr-hist}]{\includegraphics[width=0.55\linewidth]{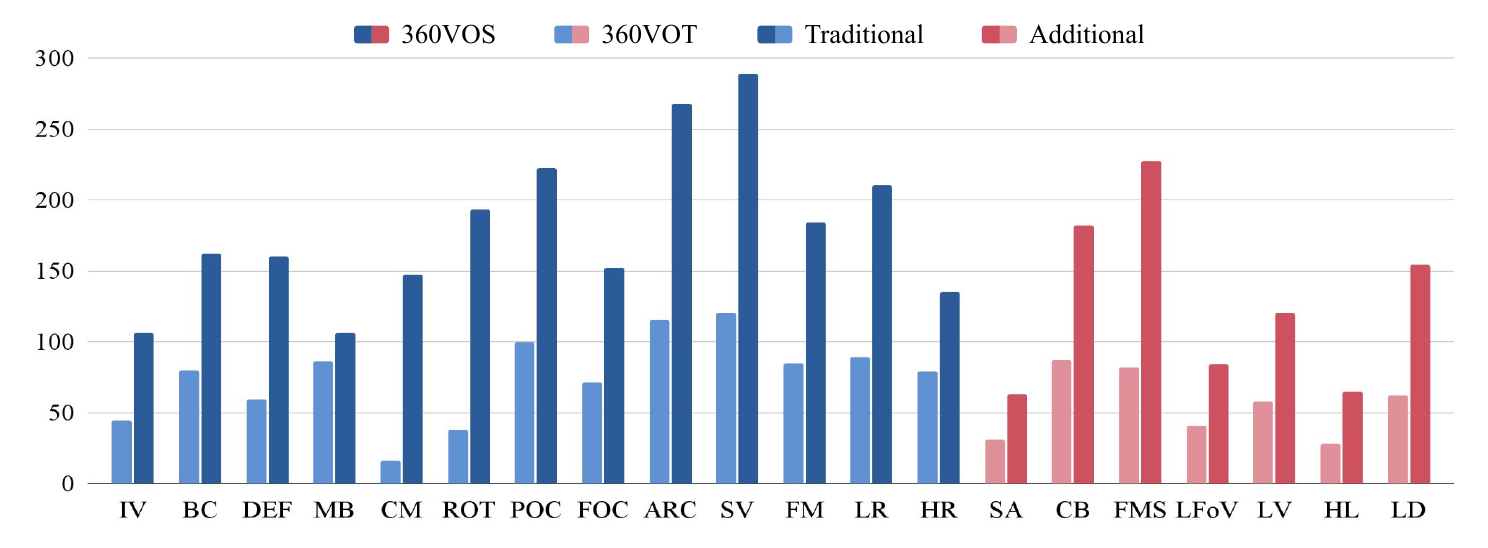}}
    \hspace{0.05cm}
    \subfloat[360VOT dependency wheel.\label{fig:attrwheel-vot}]{\includegraphics[width=0.22\linewidth]{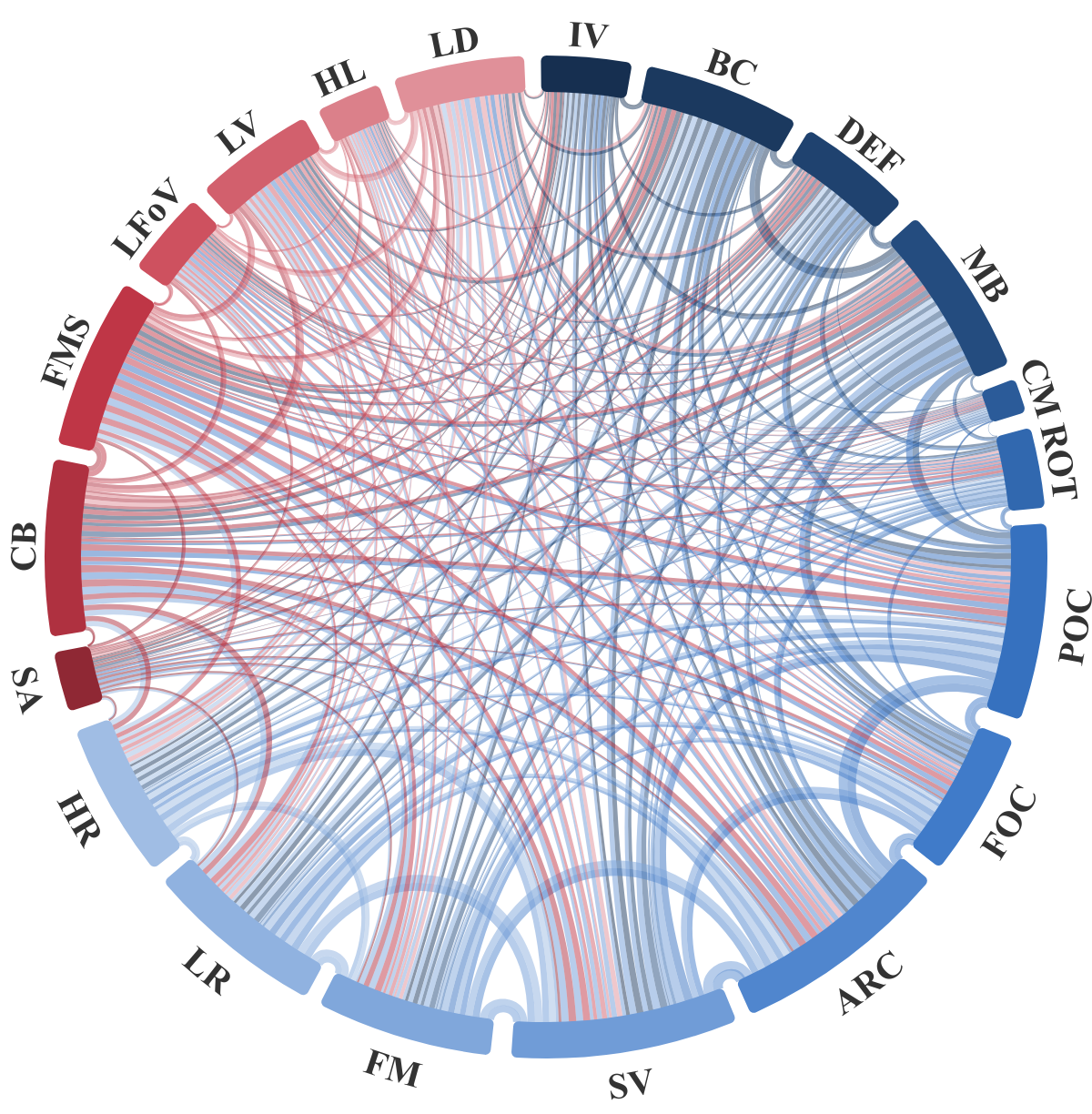}}
    \hspace{0.12cm}
    \subfloat[360VOS dependency wheel.\label{fig:attrwheel-vots}]{\includegraphics[width=0.22\linewidth]{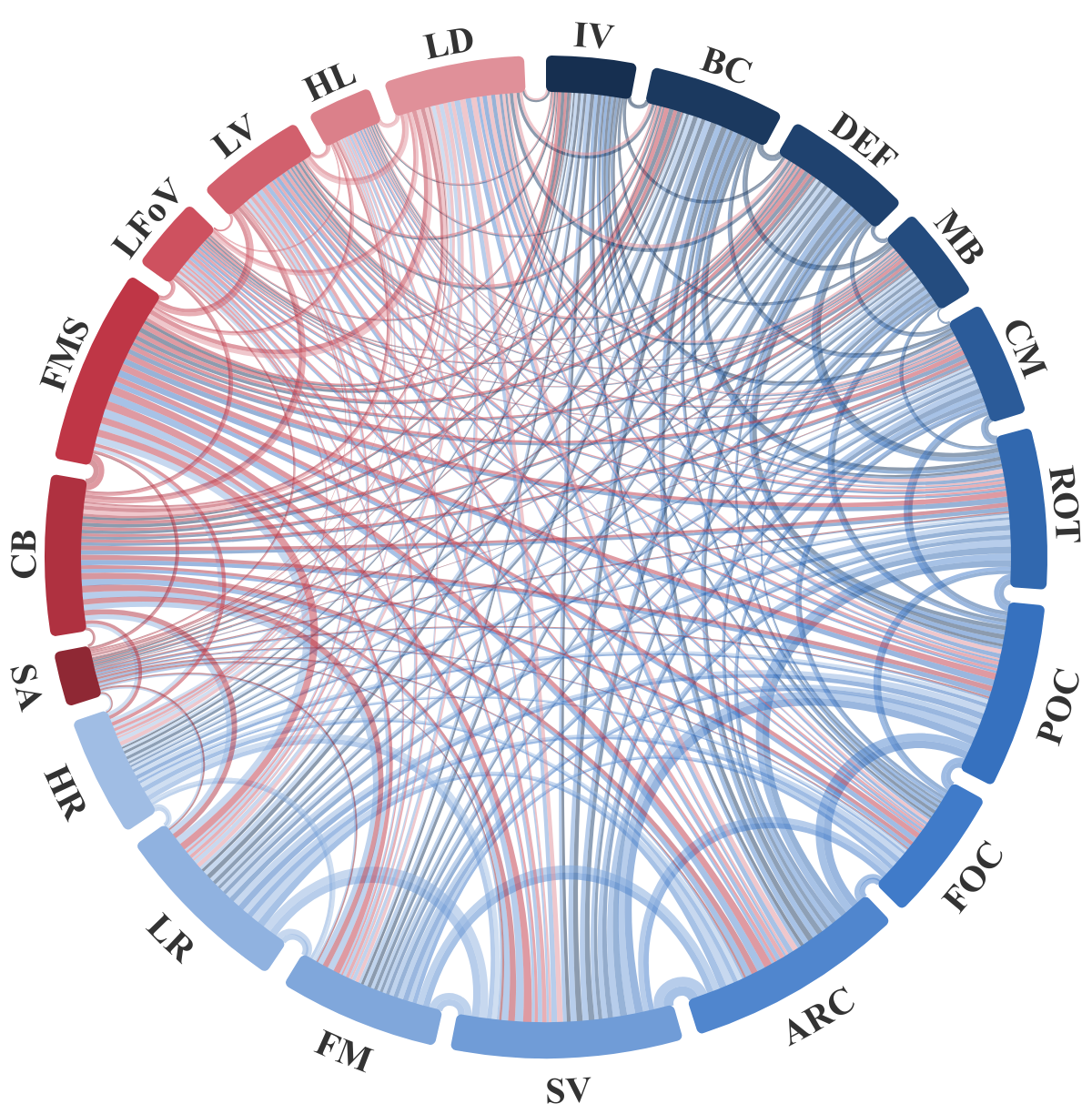}}
    \caption{Attribute distributions and dependencies in 360VOTS. The additional introduced attributes are highlighted in \st{red} palettes, while the commonly used attributes in existing works are highlighted in \nd{blue} palettes. The arc lengths on the dependency wheel measure the dominance of the attributes in the dataset. Thicker connection lines indicate a higher frequency of co-presence between attribute pairs, and vice versa.} 
    \label{fig:attr_distribution}
\end{figure*}

Figure~\ref{fig:attr_distribution} illustrates the comparison between 360VOT and 360VOS in terms of sequence counts and dependencies for each attribute. 
As shown in Figure~\ref{fig:attr-hist}, most attributes have similar proportions in 360VOT and 360VOS. 
In 360VOT, the \textit{scale changes} (ARC and SV) and \textit{motion} (MB and FM) are common challenges illustrated in Figure~\ref{fig:attrwheel-vot}.
By contrast, in 360VOS, the \textit{scale changes} (ARC and SV) remain the predominant features, whereas camera motion (CM) and target rotation (ROT) gain an obvious increase in the number of sequences illustrated in Figure~\ref{fig:attrwheel-vots}.

\subsection{Dataset Statistics of 360VOTS}\label{sec4c}

360VOTS cover four primary categories of \textit{humans}, \textit{animals}, \textit{rigid objects}, and \textit{human \& carrier cases}. {The sub-classes within each primary class are depicted in the supplementary.}
In the majority of sequences, a single target is annotated across all frames, while up to two components of a single target (e.g. bicyclist and bicycle) are annotated separately in several sequences under the \textit{human \& carrier cases} class. 
Overall, the 120 sequences in the 360VOT dataset are grouped into 32 sub-classes and used for evaluation only. The 360VOS dataset has 62 sub-classes and covers more expansive sequences.
In addition, we split the 360VOS dataset into training and test sets with balanced classes of tracking targets, avoiding class bias between the two sets.
Eventually, the test set of 360VOS comprises 98,487 annotated frames across 120 sequences and the training set includes 143,634 annotated frames within 170 sequences.
As a benchmark suitable for long-term tracking and segmentation evaluation, 360VOS is characterized by an average of 821 frames per sequence in the test set and 845 frames per sequence in the training set, respectively, as reported in Table~\ref{tab:benchmark}.


\section{Experiments}\label{sec:exp}
\subsection{Evaluation Metrics}
\label{sec:metrics}
\noindent \textbf{Evaluation on bounding regions.} To conduct the experiments on the tracking results, we use the standard one-pass evaluation (OPE) protocol~\cite{otb100} and measure the success S, precision P, and normalized precision $\overline{P}$~\cite{trackingnet} of the trackers over the test sequences. Success S is computed as the intersection over union (IoU) between the tracking results $B^{tr}$ and the ground truth annotations $B^{gt}$, where the $B^{tr/gt}$ indicates a (r)BBox. The trackers are ranked by the area under curve (AUC), which is the average of the success rates corresponding to the sampled thresholds $[0, 1]$. The precision P is computed as the distance between the results $\mathbf{C}^{tr}$ and the ground truth centers $\mathbf{C}^{gt}$. The trackers are ranked by the precision rate on the specific threshold (i.e., 20 pixels). The normalized precision $\overline{\mbox{P}}$ is scale-invariant, which normalizes the precision P over the size of the ground truth and then ranks the trackers using the AUC for the $\overline{\mbox{P}}$ between 0 and 0.5. For the perspective image using (r)BBox, these metrics can be formulated as:
	\begin{equation}\label{eq:metric1}
		\begin{split}
			\mbox{S}= IoU(B^{gt}, B^{tr}), \,  \mbox{P} = ||\mathbf{C}^{gt}_{xy} - \mathbf{C}^{tr}_{xy}||_2 \\
			\overline{\mbox{P}} = ||diag(B^{gt}, B^{tr})(\mathbf{C}^{gt}_{xy} - \mathbf{C}^{tr}_{xy})||_2 .
		\end{split}
	\end{equation}

However, for 360$\degree$ images, the target predictions may cross the image. To handle this situation and increase the accuracy of BBox evaluation, we introduce dual success S$_{dual}$ and dual precision P$_{dual}$. Specifically, we shift the $B^{gt}$ to the left and right by $W$, the width of 360$\degree$ images, to obtain two temporary ground truth $B^{gt}_l$ and $B^{gt}_r$. Based on the new ground truth, we then calculate extra success S$_{l}$ and S$_{r}$ and precision P$_{l}$ and P$_{r}$ using Eq.~\ref{eq:metric1}. Finally, S$_{dual}$ and P$_{dual}$ are measured by:
\begin{align}\label{eq:metric2}
    \mbox{S}_{dual} &= max\{\mbox{S}_l, \mbox{S}, \mbox{S}_r\}, & \mbox{P}_{dual} &= min\{\mbox{P}_l, \mbox{P}, \mbox{P}_r\} .
\end{align}
S$_{dual}$ and S, as P$_{dual}$ and P, are the same when the annotation does not cross the image border. Similarly, we can compute the dual normalized precision $\overline{\mbox{P}}_{dual}$. 

Since objects suffer significant non-linear distortion in the polar regions due to the equirectangular projection, the distance between the predicted and ground truth centers may be large on the 2D image but they are adjacent on the spherical surface. It means that dual precision P$_{dual}$ is sensitive for 360$\degree$ images. Therefore, we propose a new metric P$_{angle}$, which is measured as the angle precision between the vectors of the ground truth and the tracker results in the spherical coordinate system. The angle precision P$_{angle}$ is formulated as:
\begin{equation}\label{eq:angleP}
    \mbox{P}_{angle} = ||\mathbf{C}^{gt}_{lonlat} - \mathbf{C}^{tr}_{lonlat}||_2 \, .
\end{equation}
The different trackers are ranked with angle precision rate on a threshold, i.e., 3$\degree$. 
Moreover, when target positions are represented by BFoV or rBFoV, we utilize spherical IoU~\cite{uiou} to compute the success metric, denoted as spherical success S$_{sphere}$, while only S$_{sphere}$ and P$_{angle}$ are measured. %

\begin{figure}[t]
    \centering
    \captionsetup[subfloat]{labelfont=rm, textfont=footnotesize}

    \subfloat[Sample 360$\degree$ image with a target elephant and its corresponding mask with perspective pixel-grids.\label{fig:metrics_mrg}]{%
        \includegraphics[width=0.47\linewidth]{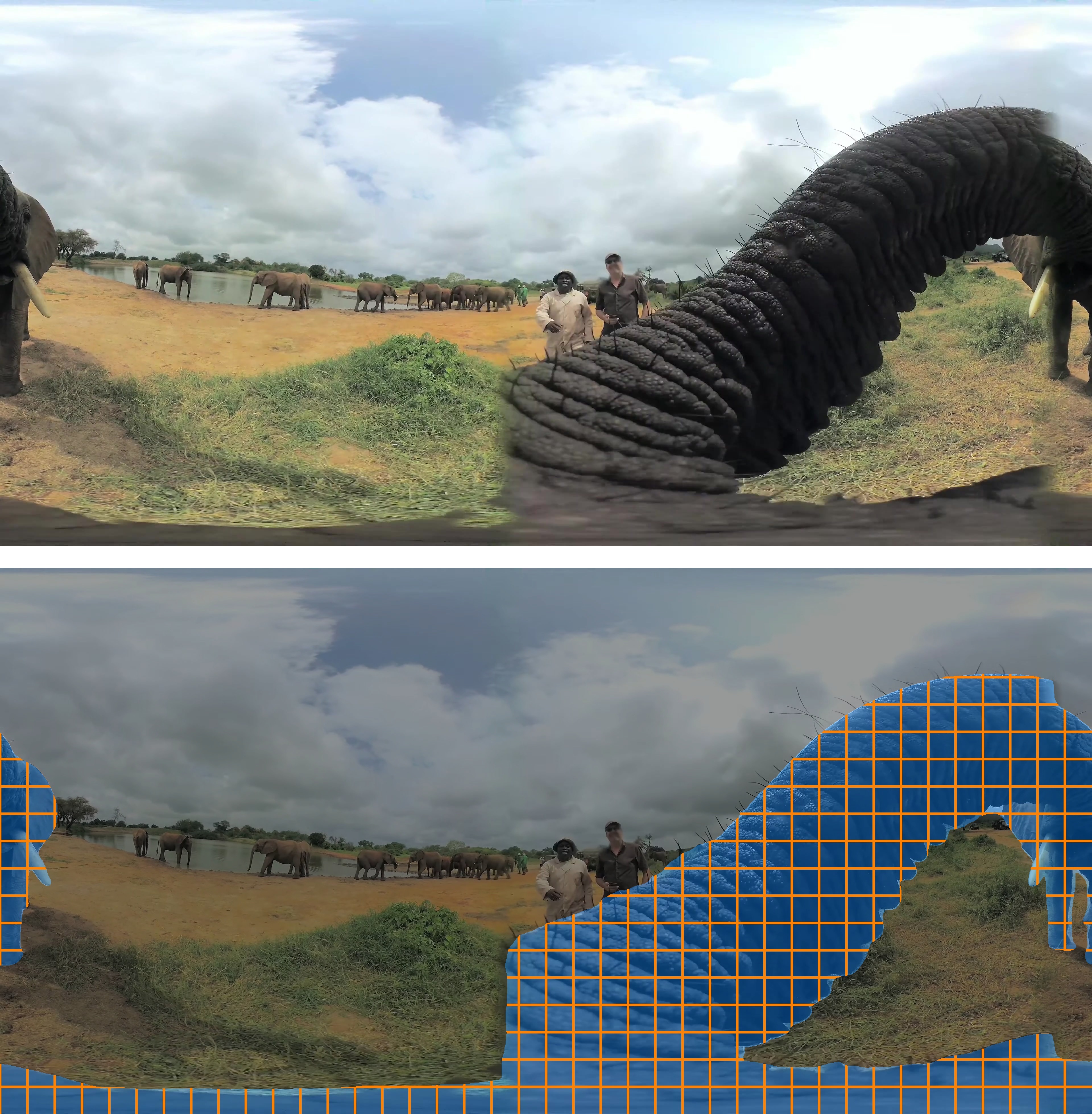}
    }
    \hspace{0.08cm}
    \subfloat[Projection of the mask and grids on a spherical surface, where the size of the grids varies with latitude.\label{fig:metrics_sphere}]{%
        \includegraphics[width=0.48\linewidth]{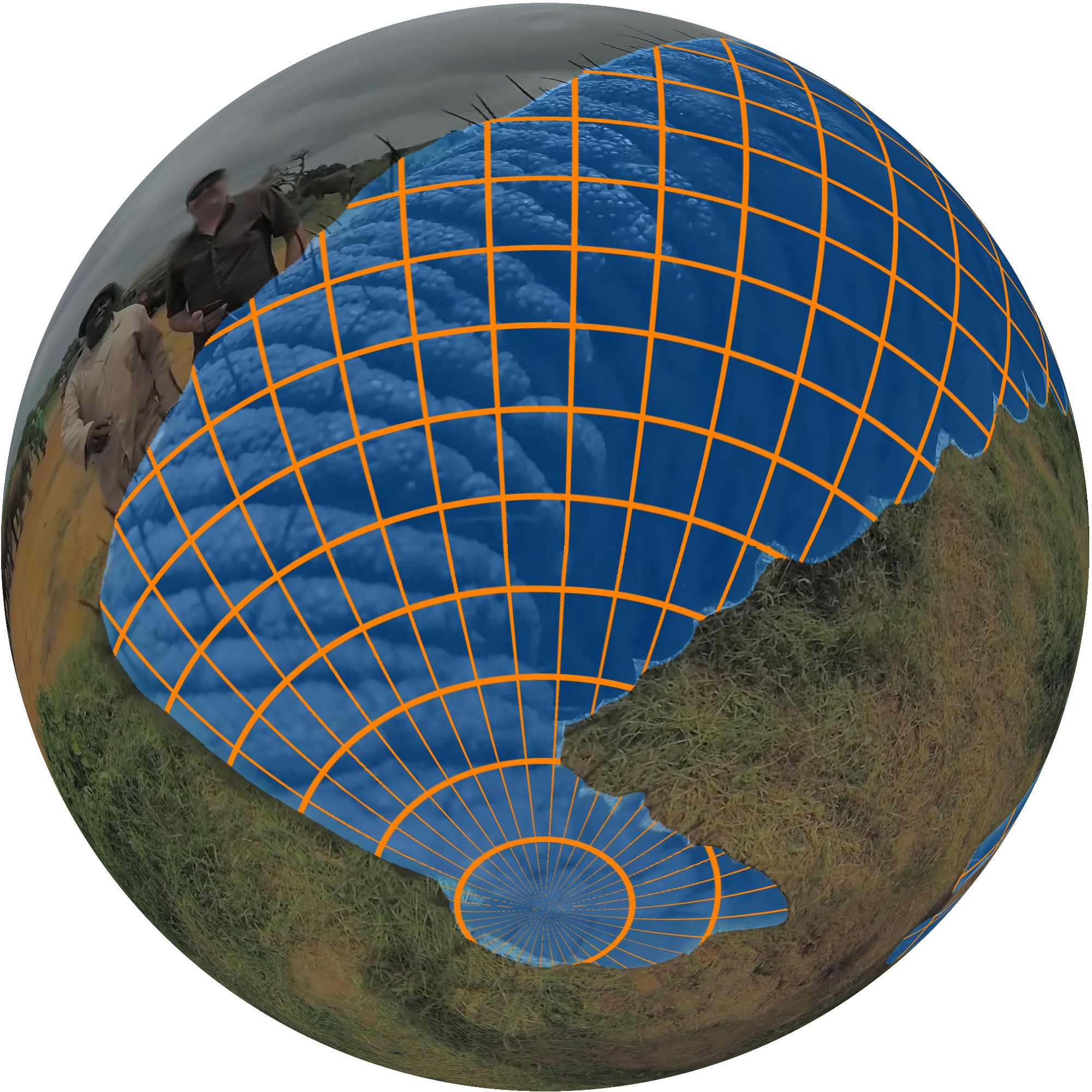}
    }
    \caption{Difference between perspective pixel grids on a 360$\degree$ image and spherical pixel grids on a 3D sphere.}
    \label{fig:metrics_example}
\end{figure}

\noindent \textbf{Evaluation on segmentation.} Region similarity $\mathcal{J}$ and contour accuracy $\mathcal{F}$ \cite{davis2016} are two common metrics in evaluation of segmentation results. Region similarity $\mathcal{J}$ measures the pixel-based IoU between an output segmentation $M^{tr}$ and its ground truth mask $M^{gt}$. $\mathcal{F}$ computes the F1-score by precision and recall on contour $\mbox{P}^c$ and $\mbox{R}^c$ between the contours of $M^{tr}$ and $M^{gt}$. The two standard metrics are defined as:
\begin{align}
\small
\mathcal{J}&=IoU(M^{gt}, M^{tr}), \!\!\!\!\!\!\!\!\!\!\!\!& \mathcal{F}&=\frac{2\mbox{P}^c\mbox{R}^c}{\mbox{P}^c+\mbox{R}^c} . \label{eq:metrics}
\end{align}


When a target is situated in high-latitude areas (i.e., close to the top or bottom) on 360$\degree$ images, its segmentation is usually stretched due to the spherical distortion, leading to issues of overestimation in the metrics of $\mathcal{J}$ and $\mathcal{F}$. Figure~\ref{fig:metrics_mrg} shows a sample target of the mentioned case, as well as its \textcolor{metmsk}{mask} and \textcolor{metgrd}{pixel grids} in the perspective view (i.e., parallel with the 360$\degree$ image boundaries) are marked separately. 
To handle this issue, we re-project the 360$\degree$ image on a 3D sphere as shown in Figure~\ref{fig:metrics_sphere}.
The spherical pixel grids near the polar regions contribute less weight to the segmentation than other regions.
Considering the weights varying along the latitude, we apply pre-computed spherical weights $\mathfrak{W}$ on the standard metrics $\mathcal{J}$ and $\mathcal{F}$ obtaining two spherical metrics, i.e., spherical region similarity $\mathcal{J}_{sphere}$ and spherical contour accuracy $\mathcal{F}_{sphere}$.


Specifically, in a 360$\degree$ image with width $W$ and height $H$, the spherical weight measuring the importance of a specific pixel $[u, v]$ can be defined by the corresponding surface area on a unit sphere (radius $r=1$), formulated as:
{
\begin{equation}\label{eq:Ws}
    \mathfrak{W}_{u,v}=\|\int_{\theta-\Delta\theta/2}^{\theta+\Delta\theta/2}\int_{\phi-\Delta\phi/2}^{\phi+\Delta\phi/2}\sin\phi\; d\phi d\theta \| ,
\end{equation}
}
{where $[\theta, \phi]$ is the longitude and latitude of the spherical coordinate which can be computed from $[u, v]$ via Eq.~\ref{eq:xyz2uv}, $\Delta\theta=2\pi/W$ and $\Delta\phi=\pi/H$.}

With the pre-computed spherical weights, spherical region similarity $\mathcal{J}_{sphere}$ is defined by:
\begin{equation}\label{eq:Js}
    \mathcal{J}_{sphere}=IoU(M^{gt}\odot \mathfrak{W}, M^{tr}\odot \mathfrak{W}) .
\end{equation}
For evaluating contour accuracy on a sphere $\mathcal{F}_{sphere}$, we adopt the bipartite graph matching \cite{fmatch} as in \cite{davis2016}. The spherical contour accuracy $\mathcal{F}_{sphere}$ is measured by:
\begin{equation}\label{eq:metric5}
    \begin{split}
        \mbox{P}_{sphere}^c&=\frac{\sum \Psi(c(M^{tr}) | c(M^{gt}))\odot \mathfrak{W}}{\sum c(M^{tr})\odot \mathfrak{W}}, \\
        \mbox{R}_{sphere}^c&=\frac{\sum \Psi(c(M^{gt}) | c(M^{tr}))\odot \mathfrak{W}}{\sum c(M^{gt})\odot \mathfrak{W}}, \\
        \mathcal{F}_{sphere}&=\frac{2\mbox{P}_{sphere}^c\mbox{R}_{sphere}^c}{\mbox{P}_{sphere}^c+\mbox{R}_{sphere}^c},
    \end{split}
\end{equation}
where the matching operation $\Psi(\cdot | \cdot)$, the contour of a mask $c(\cdot)$, and the element-wise multiplication $\odot$.

\subsection{VOT Evaluation}\label{sec:vot-evaluation}
\noindent \textbf{VOT baseline trackers.} {We evaluated 24 state-of-the-art visual object trackers on 360VOT, as shown in Table~\ref{tab:result}.} The compared methods can be roughly classified into three groups: transformer trackers, Siamese trackers, and other deep learning-based trackers.
{Specifically, the transformer trackers contain ARTrack~\cite{artrack}, 
HIPTrack~\cite{hiptrack}, OSTrack~\cite{ostrack}, LoRAT~\cite{lorat}, Stark~\cite{stark}, ToMP~\cite{tomp}, MixFormer~\cite{mixformer}, SimTrack~\cite{simtrack} and AiATrack~\cite{aiatrack}.}
The Siamese trackers include SiamDW~\cite{SiamDW}, SiamMask~\cite{siammask}, SiamRPNpp~\cite{siamrpn++}, SiamBAN~\cite{siamban}, AutoMatch~\cite{automatch}, Ocean~\cite{ocean} and SiamX~\cite{siamx2022}. For other deep trackers,  UDT~\cite{UDT}, Meta-SDNet~\cite{meta}, MDNet~\cite{mdnet}, ECO~\cite{eco}, ATOM~\cite{atom}, KYS~\cite{kys}, DiMP~\cite{dimp}, PrDiMP~\cite{prdimp} were evaluated. We used the official implementation, trained models, and default configurations to ensure a fair comparison among trackers. In addition, we developed a new baseline AiATrack-360 that combines the transformer tracker AiATrack~\cite{aiatrack} with our 360 tracking framework. We also adapted a different kind of tracker SiamX~\cite{siamx2022} with our framework, named SiamX-360, to verify the generality of the proposed framework.

\begin{table*}[t]
    \centering\footnotesize
    \setlength{\tabcolsep}{1pt} 
    \def\imgw{0.19}
    \begin{tabular}{cccccc}
    {\rotatebox{90}{\parbox{0.07\linewidth}{\centering BBox}}}& \includegraphics[width=\imgw\linewidth]{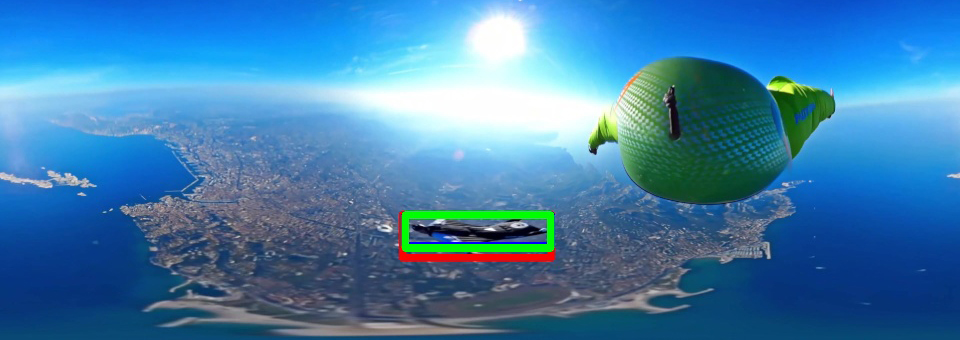}&
    \includegraphics[width=\imgw\linewidth]{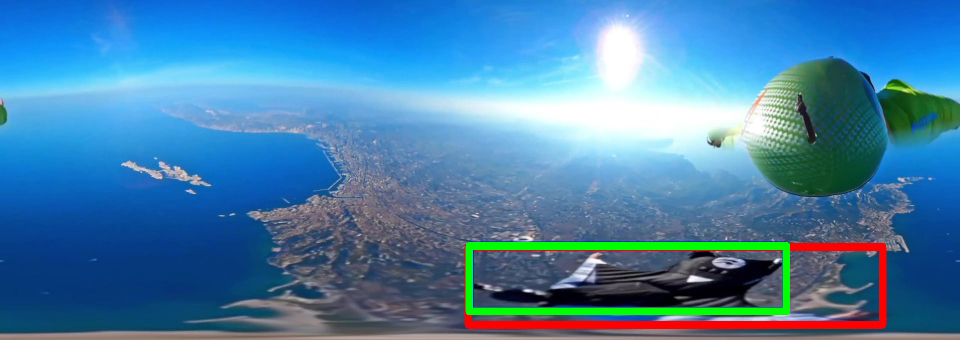}&
    \includegraphics[width=\imgw\linewidth]{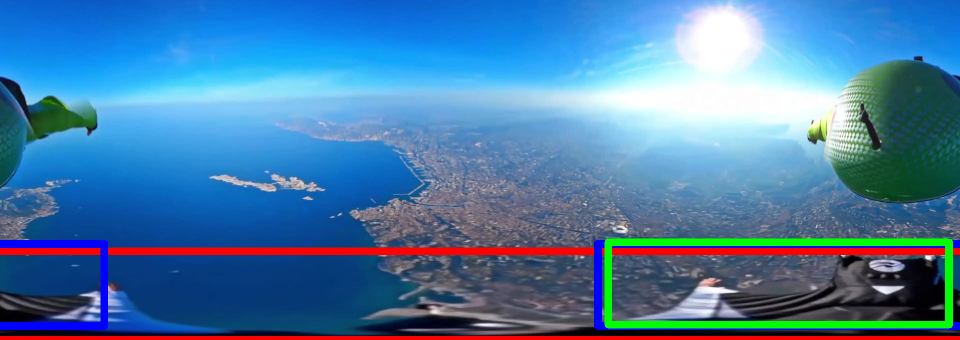}&
    \includegraphics[width=\imgw\linewidth]{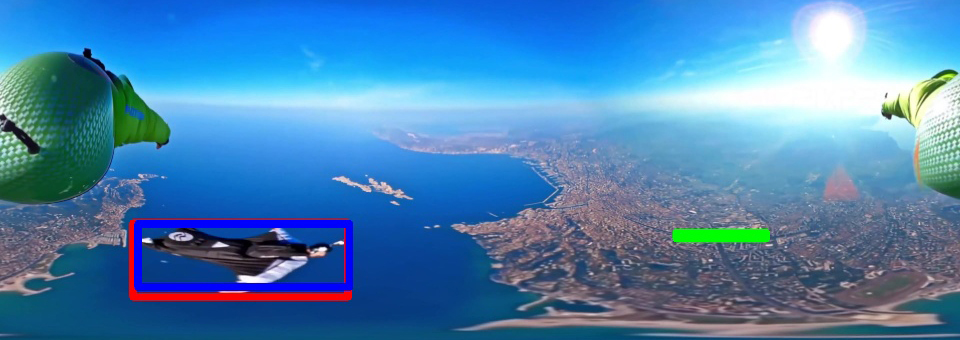}&
    \includegraphics[width=\imgw\linewidth]{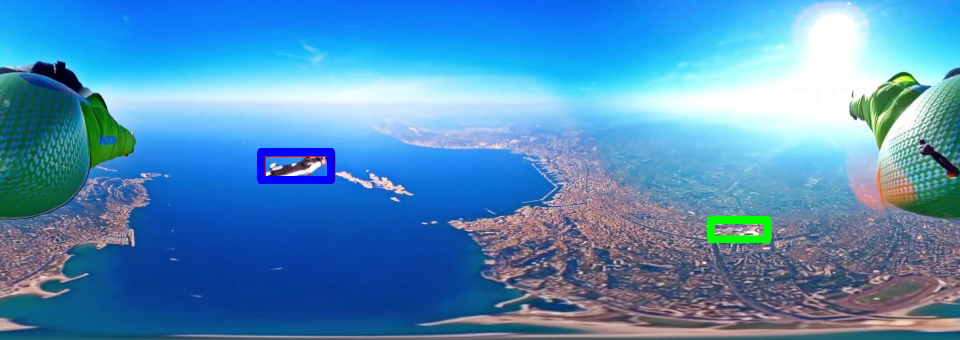}
    \vspace{-0.1cm}\\
    {\rotatebox{90}{\parbox{0.07\linewidth}{\centering rBBox}}}& \includegraphics[width=\imgw\linewidth]{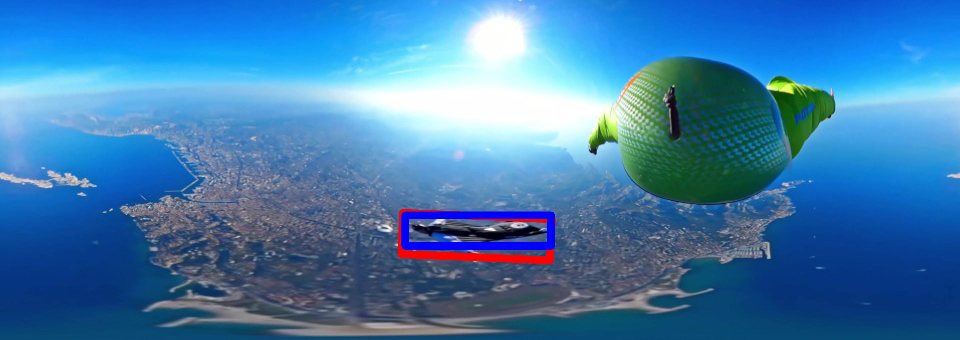}&
    \includegraphics[width=\imgw\linewidth]{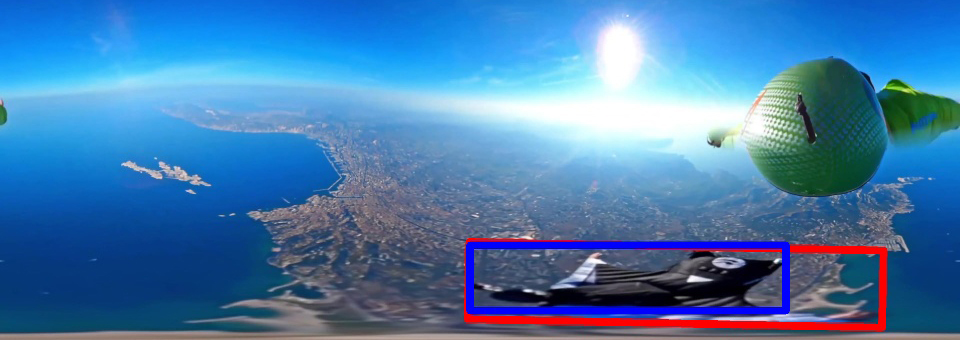}&
    \includegraphics[width=\imgw\linewidth]{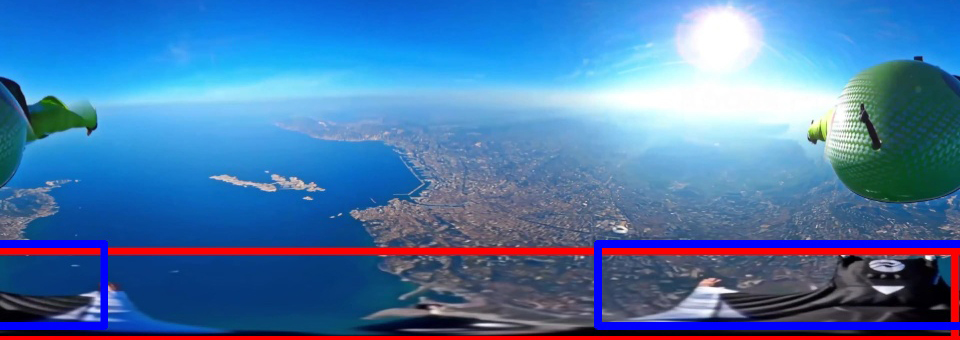}&
    \includegraphics[width=\imgw\linewidth]{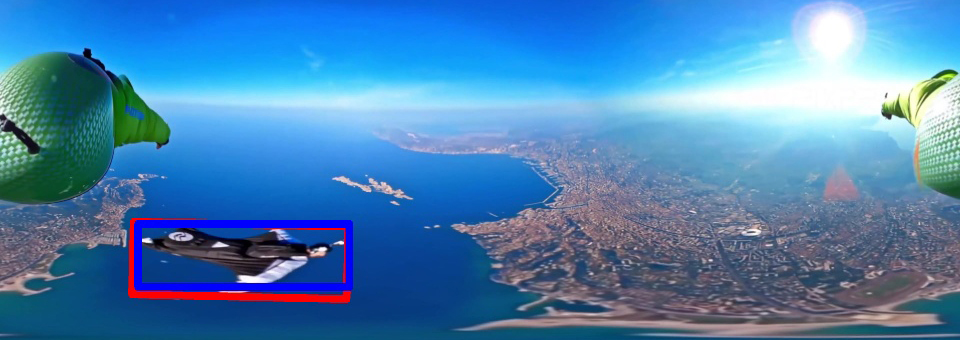}&
    \includegraphics[width=\imgw\linewidth]{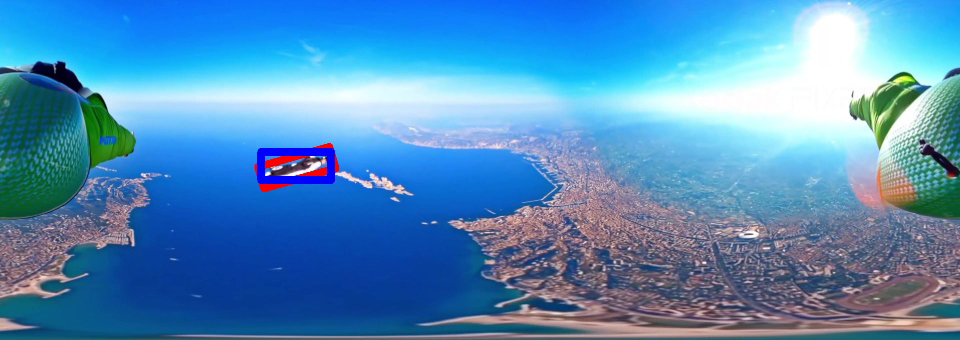} 
    \vspace{-0.1cm}\\
    {\rotatebox{90}{\parbox{0.07\linewidth}{\centering BFoV}}}&\includegraphics[width=\imgw\linewidth]{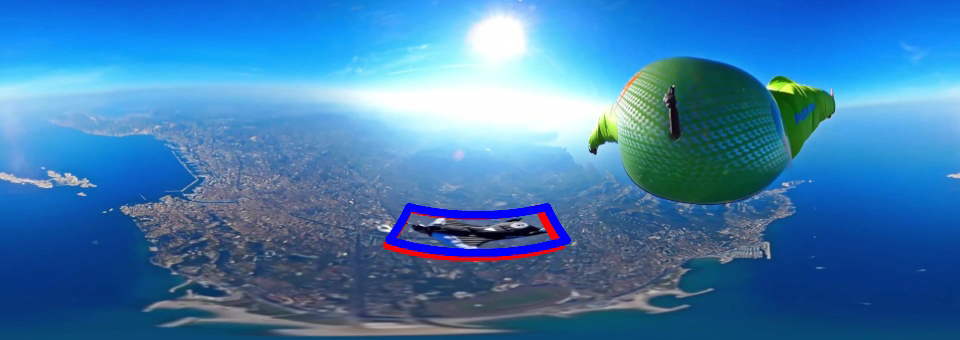}&
    \includegraphics[width=\imgw\linewidth]{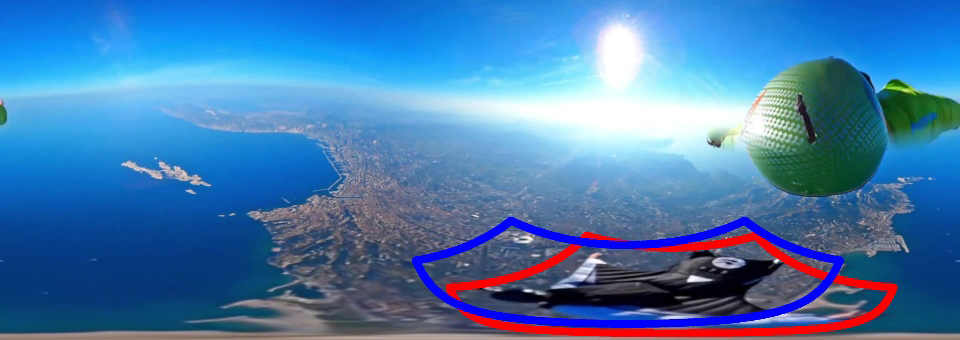}&
    \includegraphics[width=\imgw\linewidth]{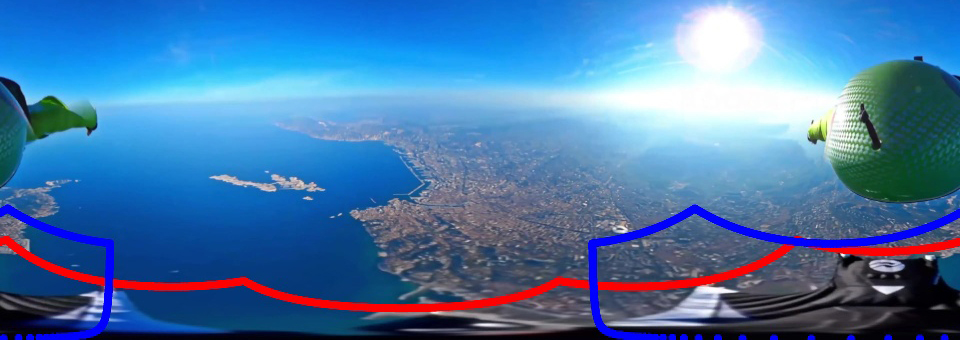}&
    \includegraphics[width=\imgw\linewidth]{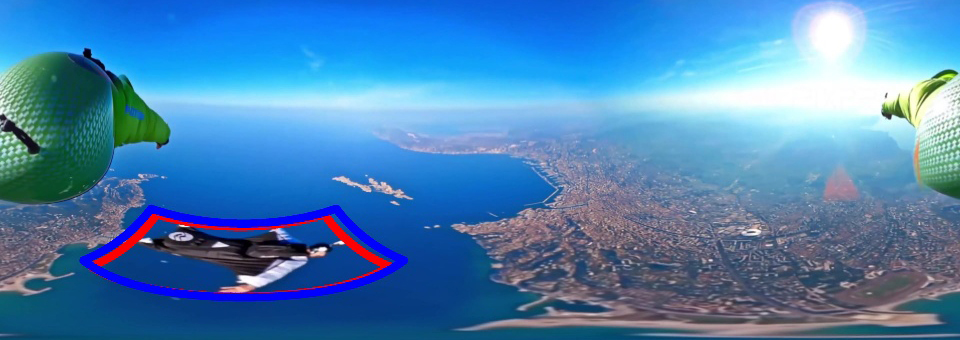}&
    \includegraphics[width=\imgw\linewidth]{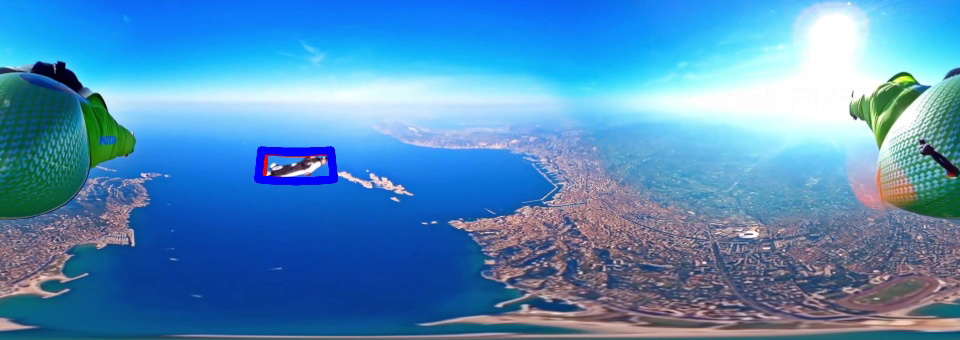}
    \vspace{-0.1cm}\\
    {\rotatebox{90}{\parbox{0.07\linewidth}{\centering rBFoV}}}&\includegraphics[width=\imgw\linewidth]{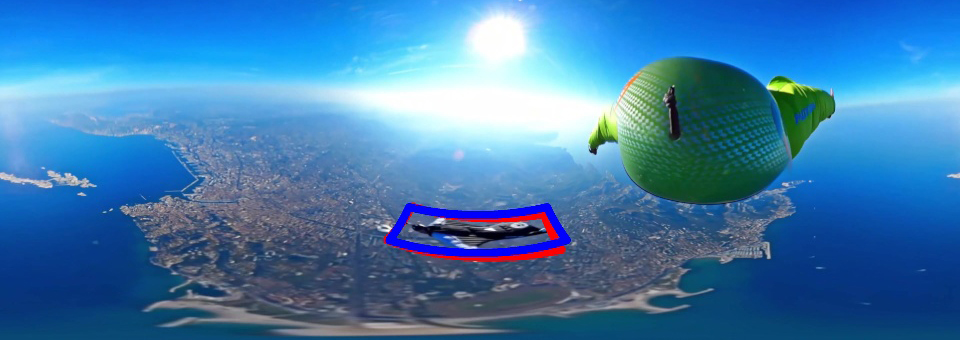}&
    \includegraphics[width=\imgw\linewidth]{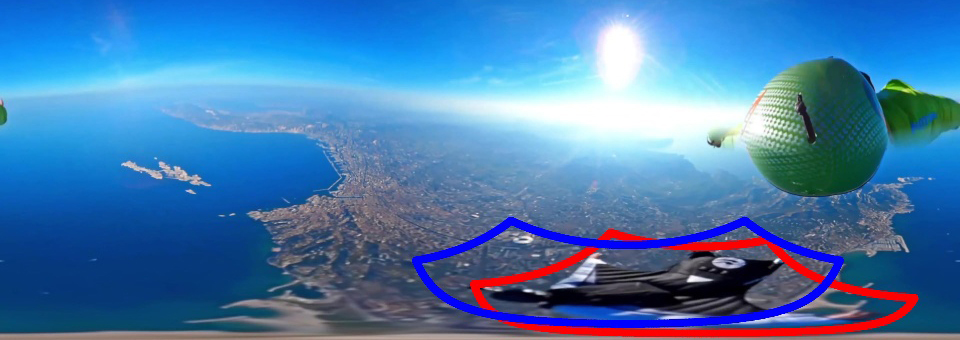}&
    \includegraphics[width=\imgw\linewidth]{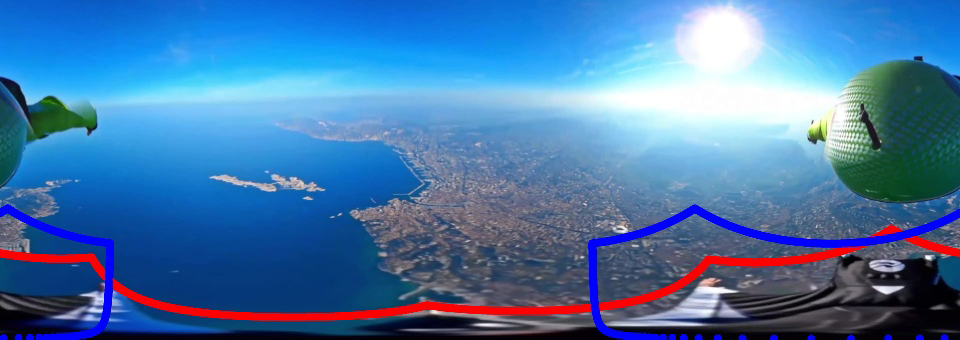}&
    \includegraphics[width=\imgw\linewidth]{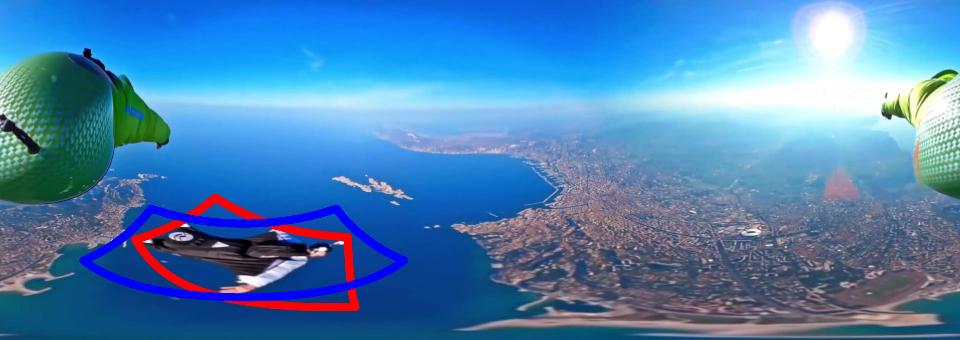}&
    \includegraphics[width=\imgw\linewidth]{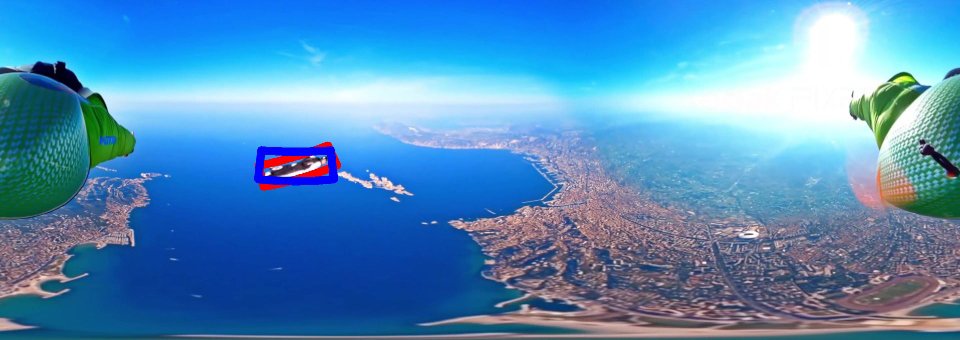} 
    \vspace{-0.1cm}\\
    \end{tabular}
    
    \captionof{figure}{Qualitative tracking results of the baseline on different representations of 360VOT. \textcolor{red}{Red} denotes the ground truth, and \textcolor{blue}{blue} denotes the results of AiATrack-360. The \textcolor{green}{green} in the first row denotes the results of AiATrack.}
    \label{fig:qual_result}
\end{table*}

\renewcommand{\arraystretch}{0.98}
\begin{table}[t]
    \captionsetup{labelsep=newline,justification=centering}
    \caption{
    {Overall performance on 360VOT BBox in terms of dual success, dual precision, dual normalized precision, and angle precision.\\ Bold \nd{blue} indicates the best results in the tracker group. Bold \st{red} indicates the best results overall. 
    }}
    \vspace{-0.1cm}
    \centering
    \footnotesize
    \begin{tabular}{lcccc}
    \toprule
	\multirow{2}{*}{VOT Tracker}& \multicolumn{4}{c}{360VOT BBox}\\ \cmidrule(lr){2-5}
	&\makecell{$\mbox{S}_{dual}$\\\footnotesize(AUC)} &$\mbox{P}_{dual}$ &\makecell{$\overline{\mbox{P}}_{dual}$\\\footnotesize(AUC)} &$\mbox{P}_{angle}$\\
	\midrule            
    UDT\cite{UDT}             &0.104  &0.075  &0.117  &0.098\\
    Meta-SDNet\cite{meta}     &0.131  &0.097  &0.164  &0.136\\
    MDNet\cite{mdnet}         &0.150  &0.106  &0.188  &0.143\\
    ECO\cite{eco}             &0.175  &0.130  &0.212  &0.179\\
    ATOM\cite{atom}           &0.252  &0.216  &0.286  &0.266\\
    KYS\cite{kys}             &0.286  &0.245  &0.312  &0.296\\
    DiMP\cite{dimp}           &0.290  &0.247  &0.315  &0.299\\
    PrDiMP\cite{prdimp}       &\nd{0.341}  &\nd{0.292}  &\nd{0.371}  &\nd{0.347}\\
    \midrule                      
    SiamDW\cite{SiamDW}       &0.156  &0.116  &0.190  &0.156\\
    SiamMask\cite{siammask}   &0.189  &0.161  &0.220  &0.203\\
    SiamRPNpp\cite{siamrpn++} &0.201  &0.175  &0.233  &0.213\\
    SiamBAN\cite{siamban}     &0.205  &0.187  &0.242  &0.227\\
    AutoMatch\cite{automatch} &0.208  &0.202  &0.261  &0.248\\
    Ocean\cite{ocean}         &0.240  &0.223  &0.287  &0.264\\
    SiamX\cite{siamx2022}     &\nd{0.302}  &\nd{0.265}  &\nd{0.331}  &\nd{0.315}\\
    \midrule
    ARTrack\cite{artrack}     &0.380  &0.354  &0.396  &0.395\\
    Stark\cite{stark}         &0.381  &0.356  &0.403  &0.408\\
    ToMP\cite{tomp}           &0.393  &0.352  &0.421  &0.413\\
    MixFormer\cite{mixformer} &0.395  &0.378  &0.417  &0.424\\
    SimTrack\cite{simtrack}   &0.400  &0.373  &0.421  &0.424\\
    AiATrack\cite{aiatrack}   &0.405  &0.369  &0.427  &0.423\\
    HIPTrack\cite{hiptrack}   &0.440  &0.415  &0.463  &0.468\\
    OSTrack\cite{ostrack}     &0.447  &0.433  &0.472  &0.484\\
    LoRAT\cite{lorat}         &\nd{0.461}  &\nd{0.468}  &\nd{0.487}  &\nd{0.503}\\
    \hline & \\[\dimexpr-\normalbaselineskip+3pt]
    SiamX-360       &0.391  &0.365  &0.430  &0.425\\
    AiATrack-360    &\st{0.534}  &\st{0.506}  &\st{0.563}  &\st{0.574}\\
    \bottomrule
    \end{tabular}
    \vspace{-0.2cm}

    \label{tab:result}
\end{table}

\begin{table}[h]
    \captionsetup{labelsep=newline,justification=centering}
    \caption{Tracking performance based on other annotations of 360VOT using 360 tracking framework. }
    \vspace{-0.1cm}
    \centering
    \footnotesize
    \setlength{\tabcolsep}{4pt} 
    \begin{tabular}{lcccc }
    \toprule
    \multirow{2}{*}{VOT Tracker}& \multicolumn{4}{c}{360VOT rBBox}\\
    \cmidrule(lr){2-5}
    &$\mbox{S}_{dual}$\footnotesize{(AUC)} &$\mbox{P}_{dual}$ &$\overline{\mbox{P}}_{dual}$\footnotesize{(AUC)}&$\mbox{P}_{angle}$\\
    \midrule
    SiamX-360     &0.205 &0.278 &0.278 &0.327\\
    AiATrack-360  &0.362 &0.449 &0.516 &0.535 \\
    \midrule
     &\multicolumn{2}{c}{360VOT BFoV}& \multicolumn{2}{c}{360VOT rBFoV} \\
     \cmidrule(lr){2-3}\cmidrule(lr){4-5}
     &$\mbox{S}_{sphere}$\footnotesize{(AUC)} &$\mbox{P}_{angle}$&$\mbox{S}_{sphere}$\footnotesize{(AUC)}&$\mbox{P}_{angle}$\\
     \midrule
    SiamX-360  &0.262 &0.327 &0.243 &0.323\\
    AiATrack-360  &0.548 &0.564 &0.426 &0.530 \\
    \bottomrule
    \end{tabular}    
    \label{tab:result2}
    \vspace{-0.2cm}
\end{table}

\noindent\textbf{Overall performance on BBox.} Existing trackers take the BBox of the first frame to initialize the tracking, and the inference results are also in the form of BBox. 
Table \ref{tab:result} shows comparison results among four groups of trackers, i.e., other, Siamese, transformer baselines, and the adapted trackers for each block in the table. According to the quantitative results, PrDiMP~\cite{prdimp}, SiamX~\cite{siamx2022} and AiATrack-360 perform best in their group of trackers. Owing to the powerful network architectures, the \textit{transformer} trackers generally outperform other groups of the compared trackers. After AiATrack integrates our proposed framework, AiATrack-360 achieves a significant performance increase of 12.9\%, 13.7\%, 13.6\% and 15.1\% in terms of S$_{dual}$, P$_{dual}$, $\overline{\mbox{P}}_{dual}$ and P$_{angle}$ respectively. AiATrack-360 outperforms all other trackers with a big performance gap. Compared to SiamX, SiamX-360 is improved by 8.9\% S$_{dual}$, 10\% P$_{dual}$, 9.6\% $\overline{\mbox{P}}_{dual}$ and 11\% P$_{angle}$, which is comparable with other transformer trackers. Although the performance gains of AiATrack-360 and SiamX-360 are different, the difference validates the effectiveness and generalization of our 360 tracking framework on 360$\degree$ visual object tracking. They can serve as a new baseline for future comparison. 


\noindent\textbf{Performance based on other annotations.} Apart from BBox, we provide additional ground truth, including rBBox and (r)BFoV. As our 360 tracking framework can estimate approximate rBBox and (r)BFoV from local BBox predictions, we additionally evaluate the performances of SiamX-360 and AiATrack-360 on these three representations (Table~\ref{tab:result2}). Compared with the results on BBox (Table~\ref{tab:result}), the performance on rBBox declines vastly. SiamX-360 and AiATrack-360 only achieve 0.205 and 0.362 S$_{dual}$ respectively. By contrast, the evaluation of (r)BFoV has more reasonable and consistent numbers. 
In addition, we display visual results by AiATrack-360 and AiAtrack in Figure \ref{fig:qual_result}. 
AiATrack-360 can follow and localize targets in challenging cases. Compared with (r)BBox, (r)BFoV can bind targets accurately with fewer irrelevant areas.
From the extensive evaluation, we observe that (r)BFoV is beneficial for object localization in omnidirectional scenes. 
\subsection{VOS Evaluation} 
\label{sec:vos-evaluation}
\noindent \textbf{VOS baseline trackers.} On 360VOS benchmark, we tested 16 state-of-the-art video object segmentation trackers. Among them, we included five memory-based methods that have demonstrated impressive results in VOS tasks: STM\cite{stm}, GMVOS\cite{gevos}, STCN\cite{stcn}, XMem\cite{xmem}, and XMem++\cite{xmem2}. Additionally, we evaluated unified VOS methods across segmentation and tracking namely RTS\cite{rts}, UNICORN\cite{unicorn}, and TarViS\cite{tarviS}, transformer-based methods including AOT\cite{aot} and DeAOT\cite{deaot}, and other learning-based methods including AFB-URR\cite{afb}, CFBI\cite{cfbi}, CFBI+\cite{cfbip}, LWL\cite{lwl}, TBD\cite{tbd}, and JOINT\cite{joint}. 
In the experiments, some listed trackers had excessive GPU and RAM demands when handling a long sequence (i.e., 2400 frames) of 360VOS. For memory-based trackers STCN\cite{stcn}, GMVOS\cite{gevos}, and STM\cite{stm}, we limited their memory clip length to 16. As for trackers AOT\cite{aot} and DeAOT\cite{deaot} that integrate long-term memory, we only kept the latest 32 clips of their long-term memory.
Furthermore, we combined XMem\cite{xmem} with our 360 tracking framework as a \textit{framework baseline} XMem-360. To verify the efficacy of the 360VOS training set, we retrained XMem \cite{xmem} to obtain a \textit{retrained baseline} XMem$^\star$ and evaluated it without our framework. Finally, we applied the retrained model to the 360 tracking framework, denoted as XMem-360$^\star$.

\begin{table}[t]
    \captionsetup{labelsep=newline,justification=centering}
    \caption{
    Segmentation performance on 360VOS in terms of standard and spherical region similarity and contour accuracy. 
    Bold \nd{blue} indicates the best results in existing trackers, while bold \st{red} indicates the best results overall.
    }
    \centering
    \footnotesize
    \begin{tabular}{lcccc}
    \toprule
        \multirow{2}{*}{VOS Tracker}& \multicolumn{4}{c}{360VOS}\\ \cmidrule(lr){2-5}
         & $\mathcal{J}$ & $\mathcal{F}$ & $\mathcal{J}_{sphere}$ & $\mathcal{F}_{sphere}$\\
    \midrule   
        STM\cite{stm} & 0.364 & 0.438 & 0.366 & 0.439 \\ 
        UNICORN\cite{unicorn} & 0.339 & 0.469 & 0.340 & 0.470 \\ 
        TarVis\cite{tarviS} & 0.324 & 0.411 & 0.325 & 0.414 \\
        AFB-URR\cite{afb} & 0.389 & 0.481 & 0.392 & 0.483 \\
        GMVOS\cite{gevos} & 0.431 & 0.522 & 0.435 & 0.526 \\ 
        CFBI\cite{cfbi} & 0.412 & 0.513 & 0.414 & 0.516 \\ 
        CFBI+\cite{cfbip} & 0.428 & 0.532 & 0.430 & 0.534 \\ 
        DeAOT\cite{deaot} & 0.447 & 0.578 & 0.448 & 0.579 \\ 
        AOT\cite{aot} & 0.471 & 0.597 & 0.473 & 0.598 \\ 
        LWL\cite{lwl} & 0.469 & 0.567 & 0.471 & 0.569 \\ 
        TBD\cite{tbd} & 0.474 & 0.598 & 0.477 & 0.600 \\ 
        JOINT\cite{joint} & 0.487 & 0.587 & 0.488 & 0.588 \\ 
        RTS\cite{rts} & 0.540 & 0.645 & 0.542 & 0.647 \\ 
        STCN\cite{stcn} & 0.550 & 0.667 & 0.552 & 0.669 \\ 
        XMem\cite{xmem} & 0.560 & 0.662 & 0.562 & 0.663 \\ 
        XMem++\cite{xmem2} & \nd{0.579} & \nd{0.692} & \nd{0.581} & \nd{0.693} \\ 
    \midrule 
        Xmem$^\star$&0.596 & 0.703 & 0.597 & 0.704 \\
        XMem-360& 0.656 & 0.780 & 0.658 & 0.782 \\
        XMem-360$^\star$& \st{0.676} & \st{0.799} & \st{0.677} & \st{0.801} \\
    \bottomrule
    \end{tabular}
    \label{tab:vostrackers}
    \vspace{-0.15cm}
\end{table}

{
\renewcommand{\arraystretch}{0.97}
\begin{table}[t]
    \captionsetup{labelsep=newline,justification=centering}
    \caption{
    Tracking performance of adapting VOS trackers on the 360VOT benchmark. (VOT) indicates the (r)BBox and (r)BFoV results of the trackers are converted from their original output masks. In the same group of performance, bold \st{red} highlights the best results, while bold \nd{blue} highlights the second results.}
    \centering
    \footnotesize
    \setlength{\tabcolsep}{3pt} 
    \begin{tabular}{lcccc }
    \toprule
    \multirow{2}{*}{Tracker}& \multicolumn{4}{c}{360VOT BBox}\\
    \cmidrule(lr){2-5}
    &$\mbox{S}_{dual}$\footnotesize{(AUC)} &$\mbox{P}_{dual}$ &$\overline{\mbox{P}}_{dual}$\footnotesize{(AUC)}&$\mbox{P}_{angle}$\\
    \midrule
    SiamX-360     &0.391 &0.365 &0.430 &0.425\\
    AiATrack-360  &0.534 &0.506 &0.563 &0.574 \\
    \midrule
    (VOT)~XMem\cite{xmem} &0.432 &0.425 &0.460 &0.465\\
    (VOT)~XMem$^\star$ &0.523 &0.529 &0.553 &0.576\\
    (VOT)~XMem-360 &\nd{0.535} &\nd{0.534} &\nd{0.566} &\nd{0.588}\\
    (VOT)~XMem-360$^\star$ &\st{0.583} &\st{0.588} &\st{0.607} &\st{0.637}\\
    \midrule \midrule
    \multirow{2}{*}{Tracker}& \multicolumn{4}{c}{360VOT rBBox}\\
    \cmidrule(lr){2-5}
    &$\mbox{S}_{dual}$\footnotesize{(AUC)} &$\mbox{P}_{dual}$ &$\overline{\mbox{P}}_{dual}$\footnotesize{(AUC)}&$\mbox{P}_{angle}$\\
    \midrule
    SiamX-360     &0.205 &0.278 &0.278 &0.327\\
    AiATrack-360  &\nd{0.362} &0.449 &0.516 &0.535 \\
    \midrule
    (VOT)~XMem\cite{xmem} &0.288 &0.411 &0.430 &0.457\\
    (VOT)~XMem$^\star$ &0.347 &0.511 &0.517 &0.564\\
    (VOT)~XMem-360 &0.357 &\nd{0.512} &\nd{0.524} &\nd{0.574}\\
    (VOT)~XMem-360$^\star$ &\st{0.387} &\st{0.566} &\st{0.565} &\st{0.624}\\
    \midrule \midrule
     \multirow{2}{*}{Tracker}&\multicolumn{2}{c}{360VOT BFoV}& \multicolumn{2}{c}{360VOT rBFoV} \\
     \cmidrule(lr){2-3}\cmidrule(lr){4-5}
     &$\mbox{S}_{sphere}$\footnotesize{(AUC)} &$\mbox{P}_{angle}$&$\mbox{S}_{sphere}$\footnotesize{(AUC)}&$\mbox{P}_{angle}$\\
     \midrule 
    SiamX-360  &0.262 &0.327 &0.243 &0.323\\
    AiATrack-360  &\nd{0.548} &0.564 &0.426 &0.530 \\
    \midrule
    (VOT)~XMem\cite{xmem} &0.435 &0.465 &0.414 &0.458 \\
    (VOT)~XMem$^\star$ &0.523 &0.575 &0.498 &0.568 \\
    (VOT)~XMem-360 &0.541 &\nd{0.585} &\nd{0.514} &\nd{0.577} \\
    (VOT)~XMem-360$^\star$ &\st{0.594} &\st{0.635} &\st{0.565} &\st{0.626} \\
    \bottomrule
    \end{tabular}
    \label{tab:result3}
    \vspace{-0.15cm}
\end{table}
}

\begin{table*}[t]
    \centering\small
    \footnotesize
    \setlength{\tabcolsep}{1pt} 
    \def\imgw{0.19}
    \def\imgh{0.085}
    \begin{tabular}{cccccc}

    {\rotatebox{90}{\parbox{0.09\linewidth}{\centering \scriptsize GT}}}& \includegraphics[width=\imgw\linewidth,height=\imgh\linewidth]{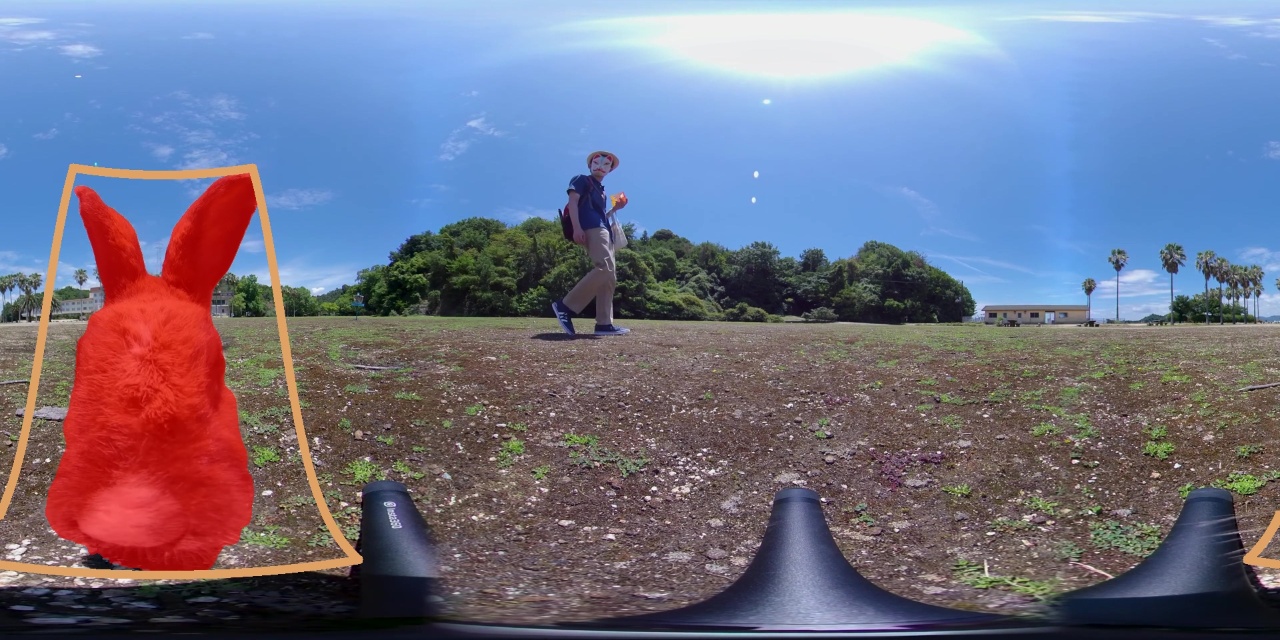}&
    \includegraphics[width=\imgw\linewidth,height=\imgh\linewidth]{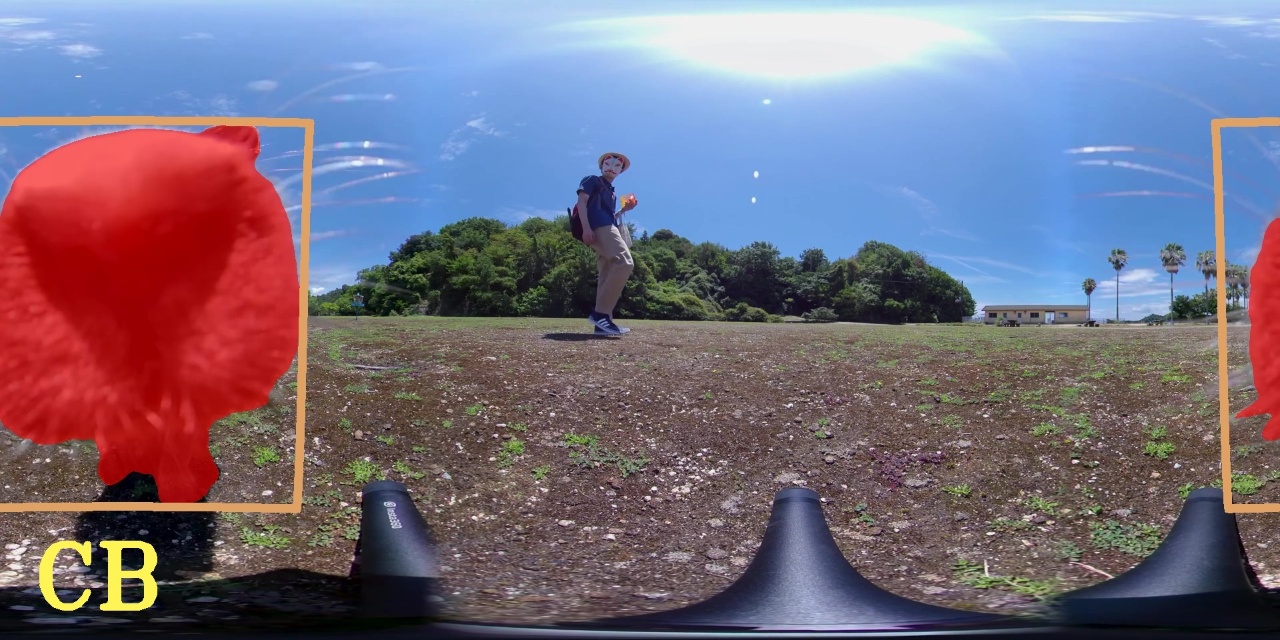}&
    \includegraphics[width=\imgw\linewidth,height=\imgh\linewidth]{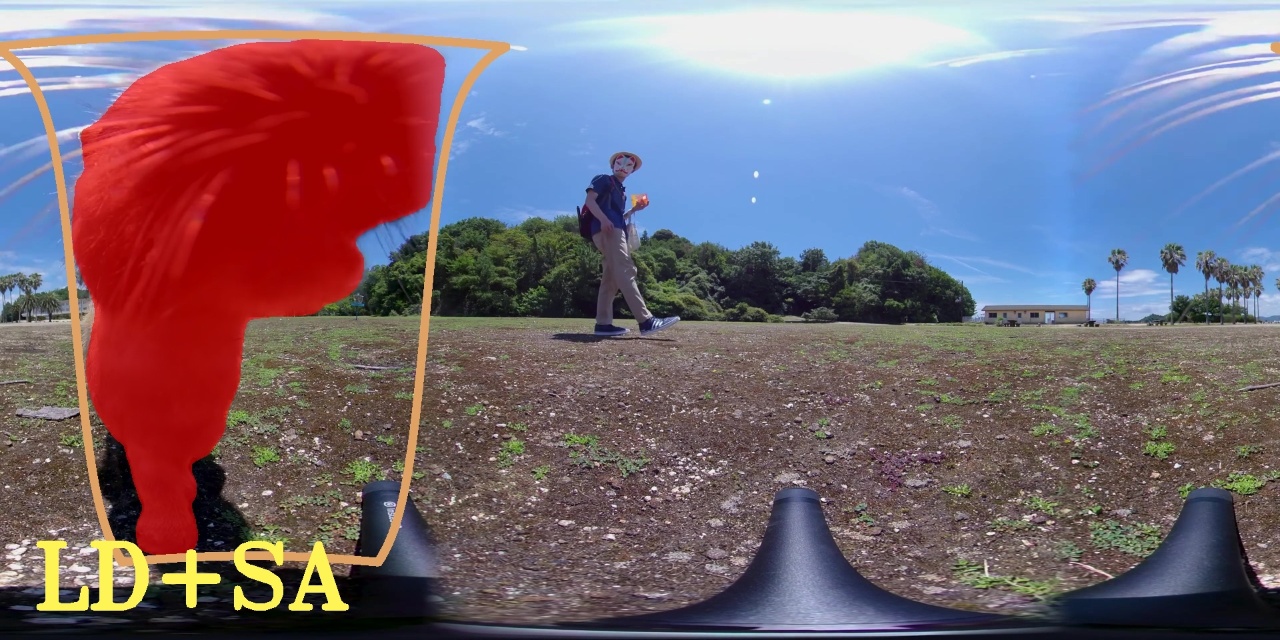}&
    \includegraphics[width=\imgw\linewidth,height=\imgh\linewidth]{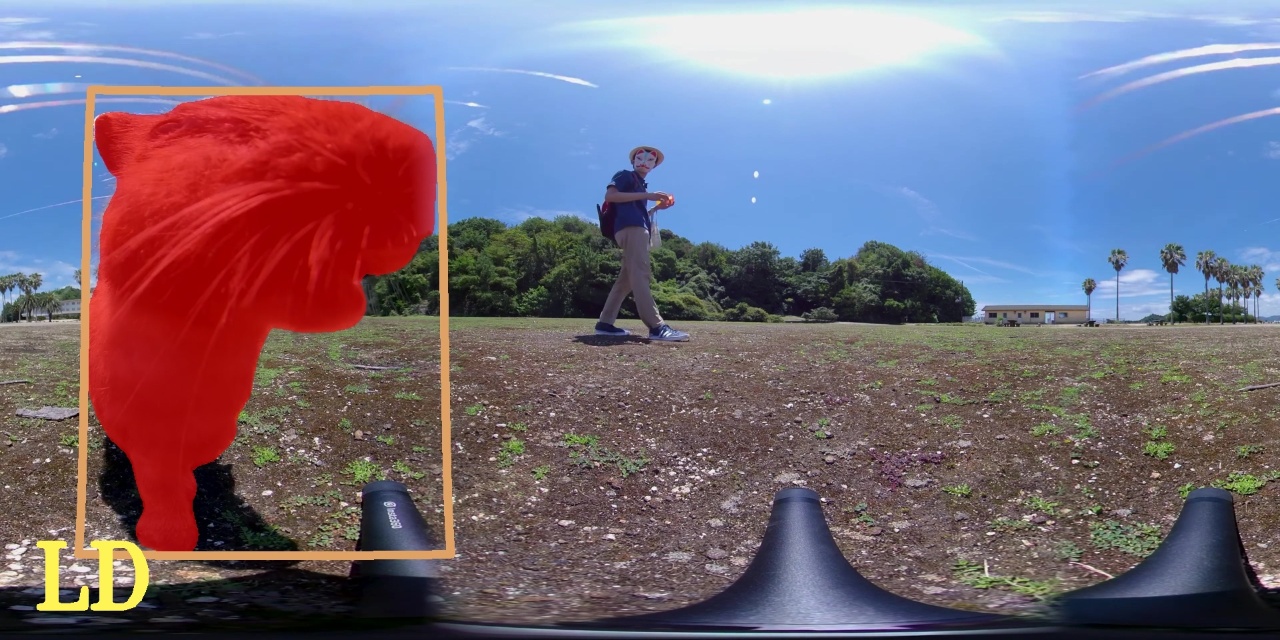}&
    \includegraphics[width=\imgw\linewidth,height=\imgh\linewidth]{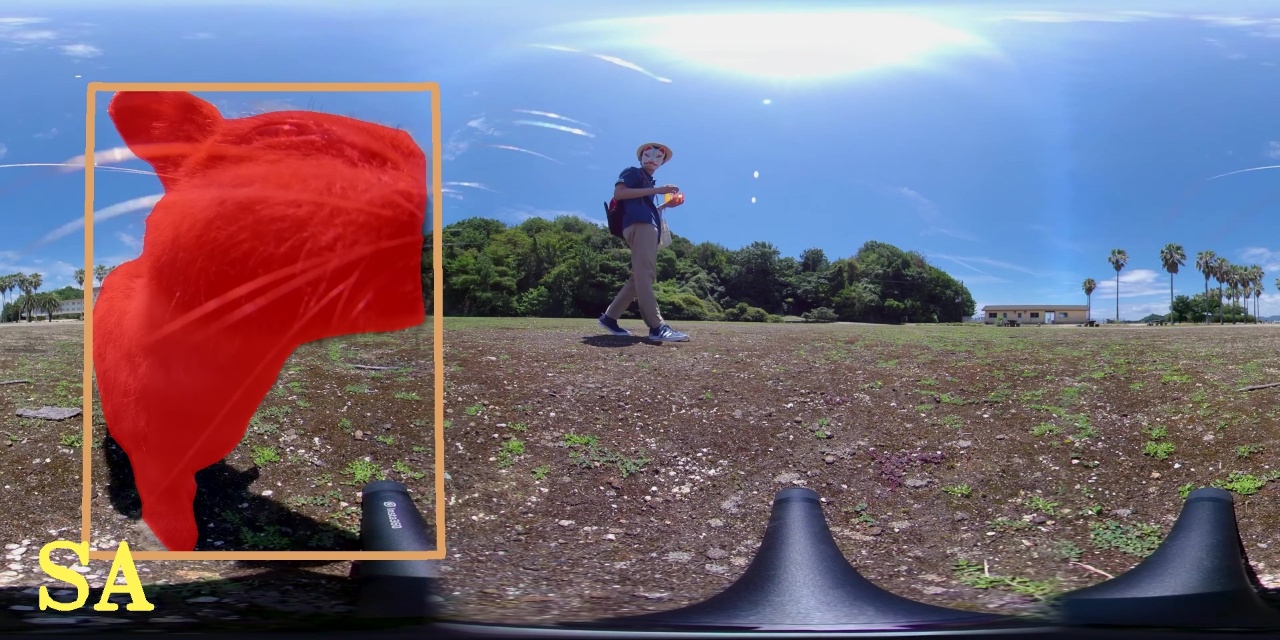}
\vspace{-0.2cm}\\
    {\rotatebox{90}{\parbox{0.09\linewidth}{\centering \scriptsize XMem}}}& \includegraphics[width=\imgw\linewidth,height=\imgh\linewidth]{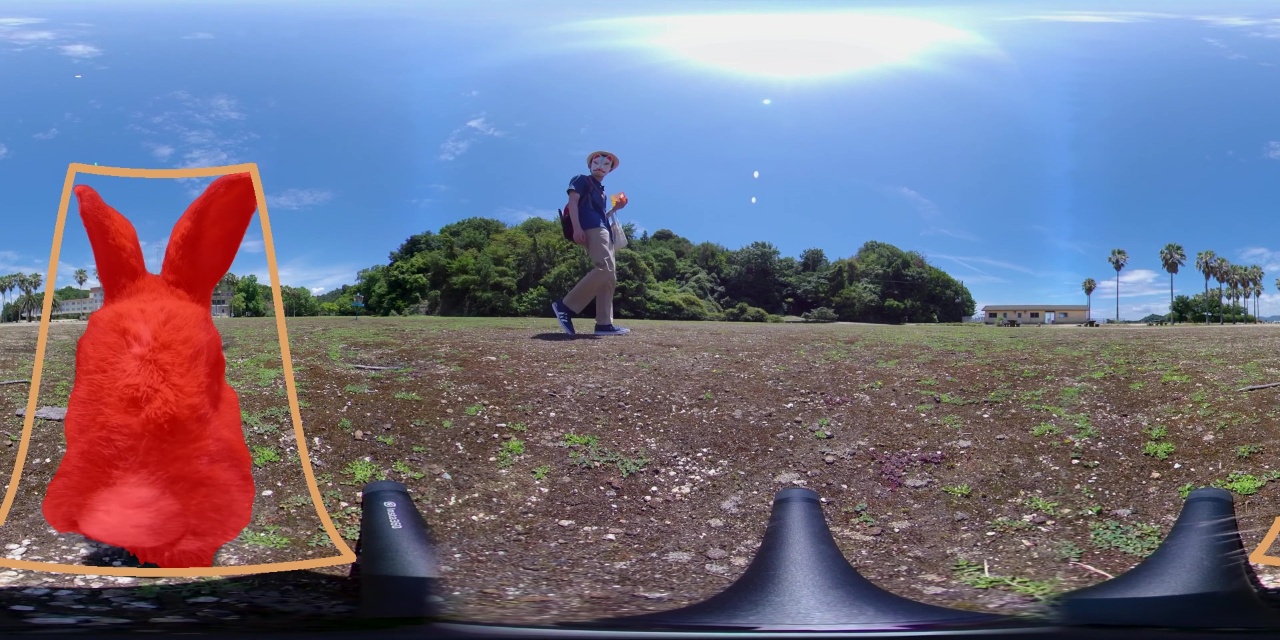}&
    \includegraphics[width=\imgw\linewidth,height=\imgh\linewidth]{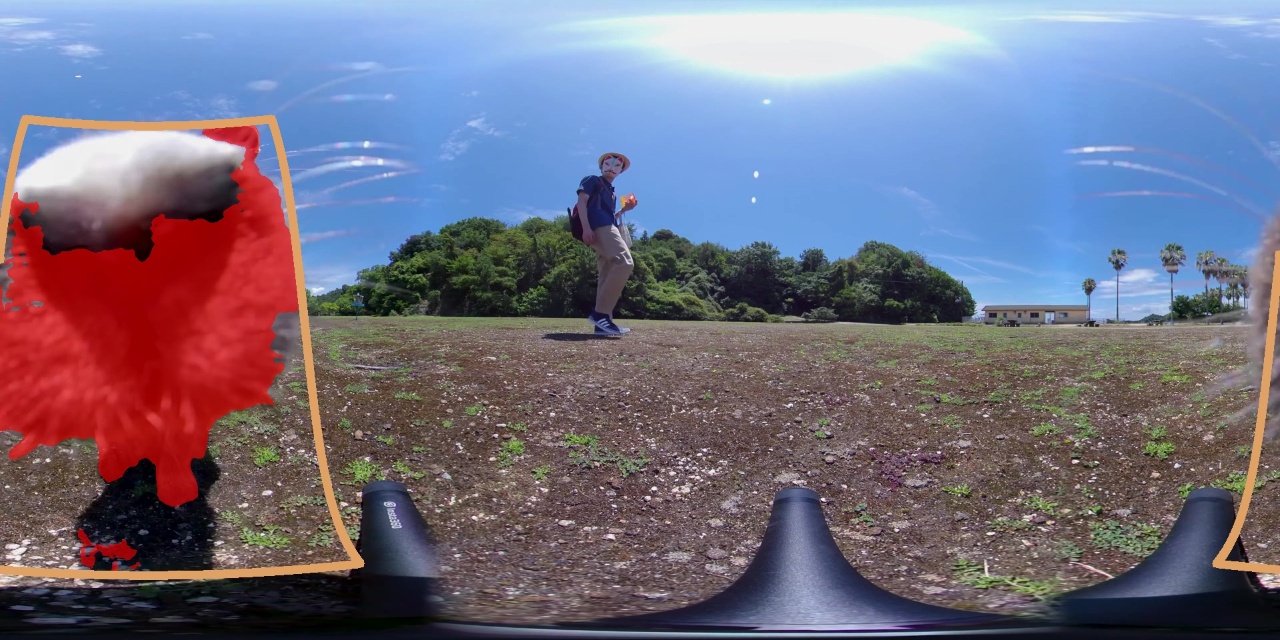}&
    \includegraphics[width=\imgw\linewidth,height=\imgh\linewidth]{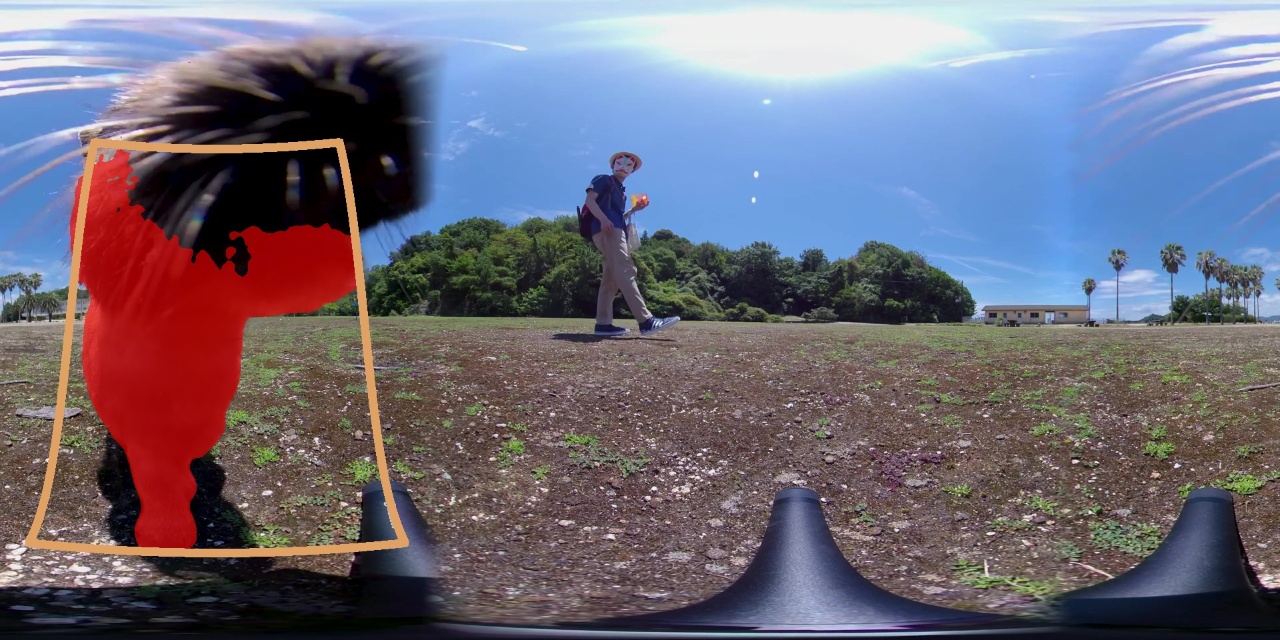}&
    \includegraphics[width=\imgw\linewidth,height=\imgh\linewidth]{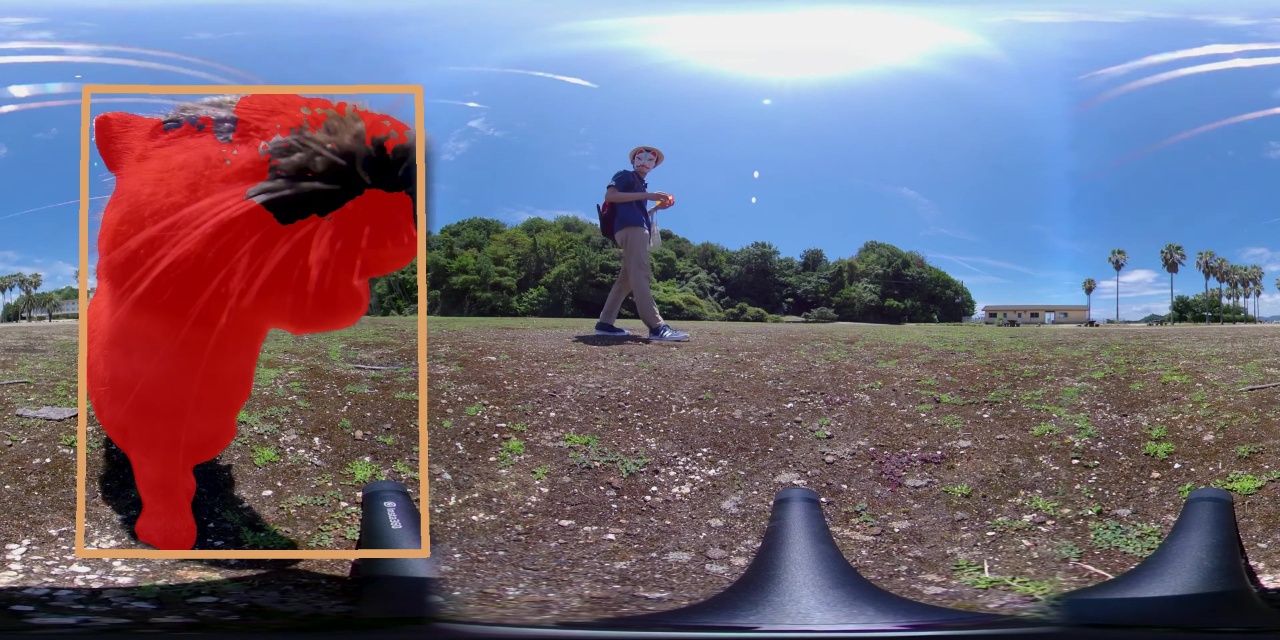}&
    \includegraphics[width=\imgw\linewidth,height=\imgh\linewidth]{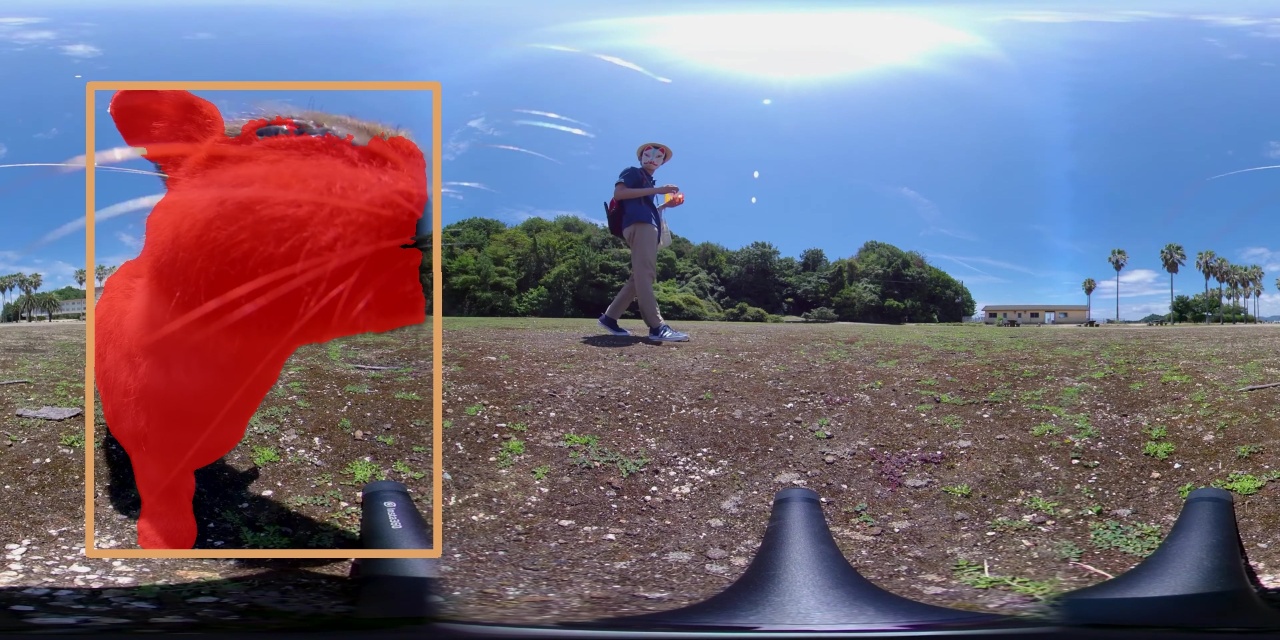} 
\vspace{-0.2cm}\\
    {\rotatebox{90}{\parbox{0.09\linewidth}{\centering \scriptsize XMem$^\star$}}}&\includegraphics[width=\imgw\linewidth,height=\imgh\linewidth]{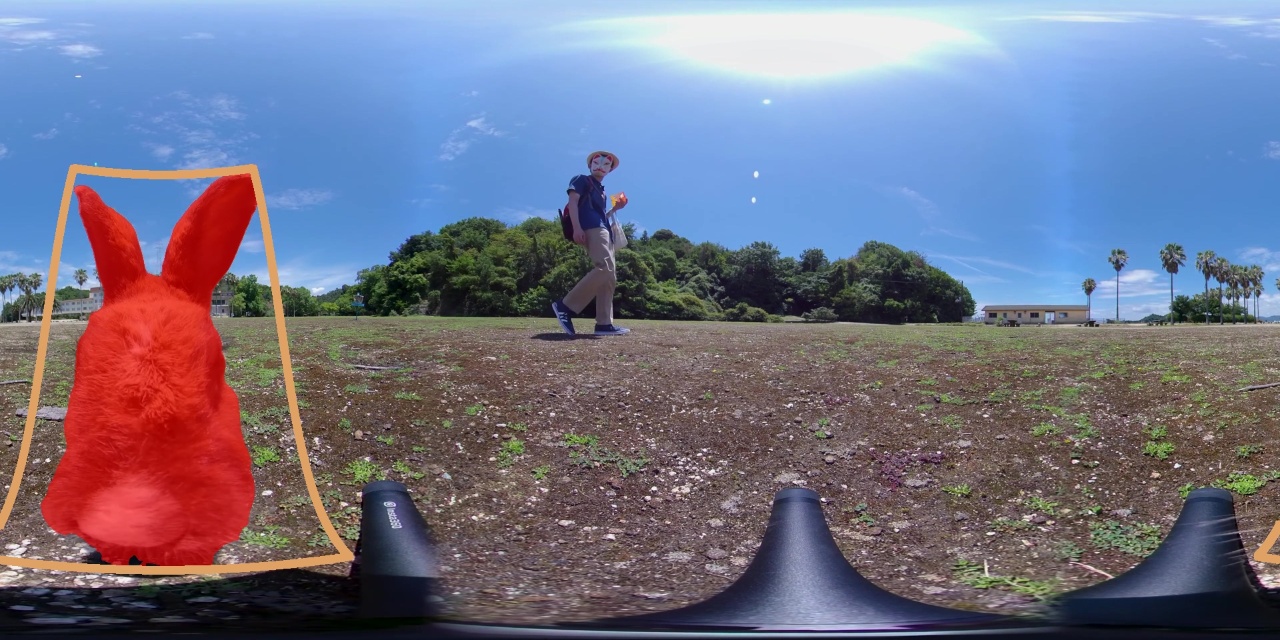}&
    \includegraphics[width=\imgw\linewidth,height=\imgh\linewidth]{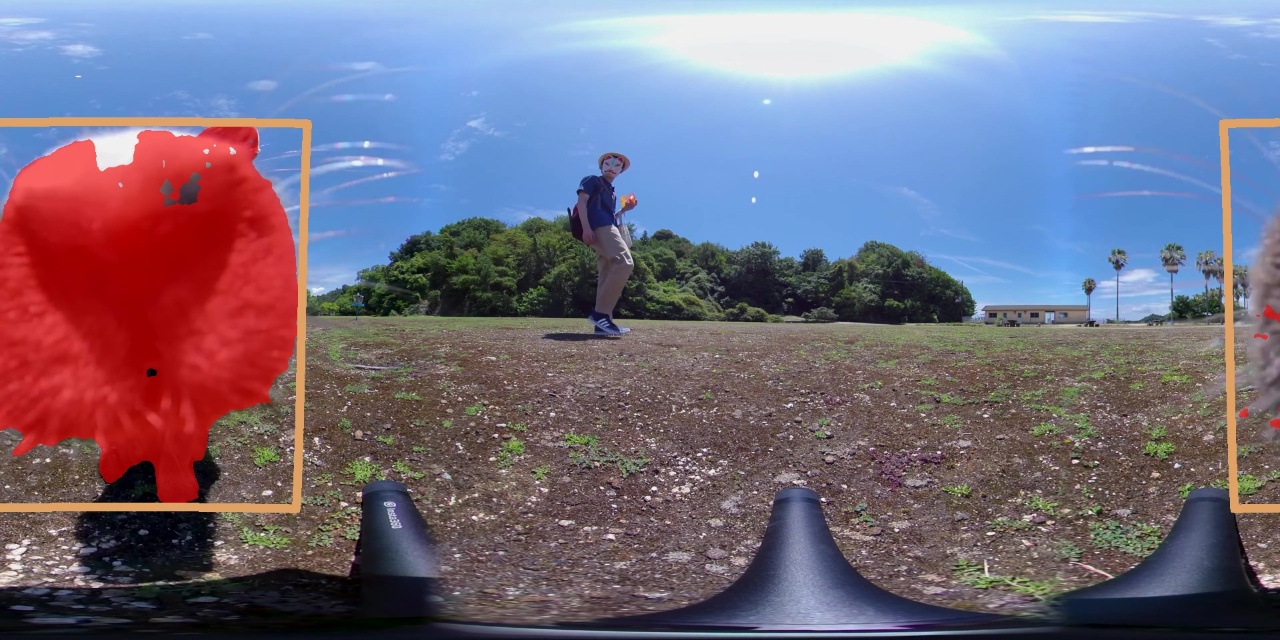}&
    \includegraphics[width=\imgw\linewidth,height=\imgh\linewidth]{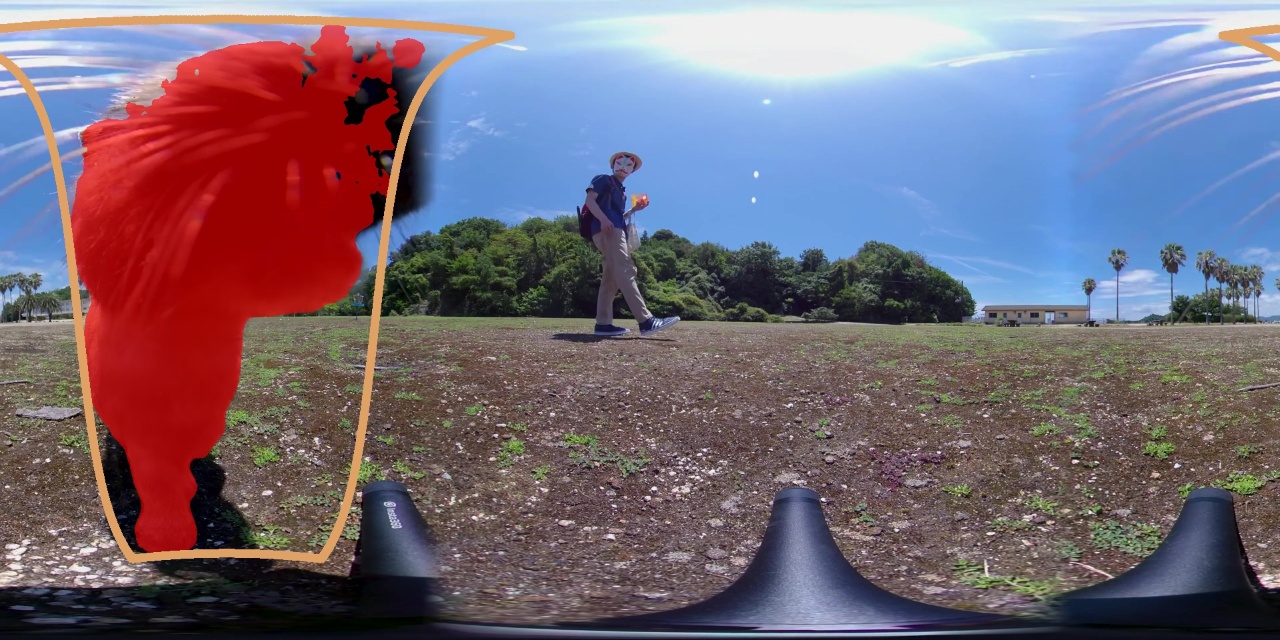}&
    \includegraphics[width=\imgw\linewidth,height=\imgh\linewidth]{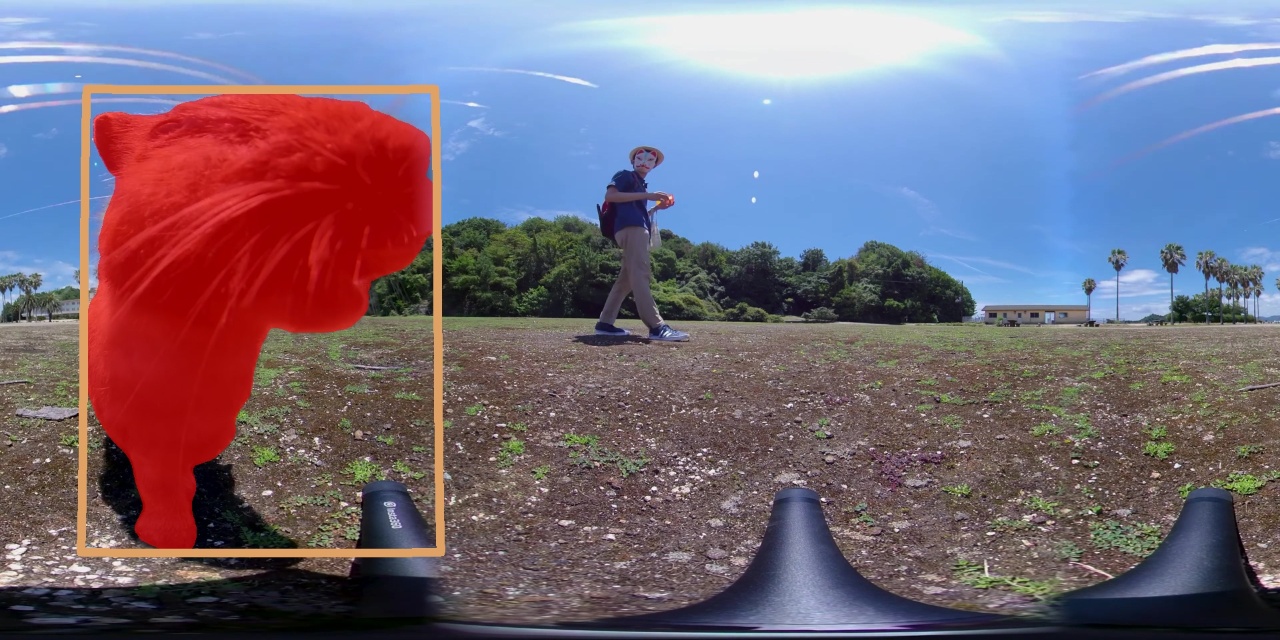}&
    \includegraphics[width=\imgw\linewidth,height=\imgh\linewidth]{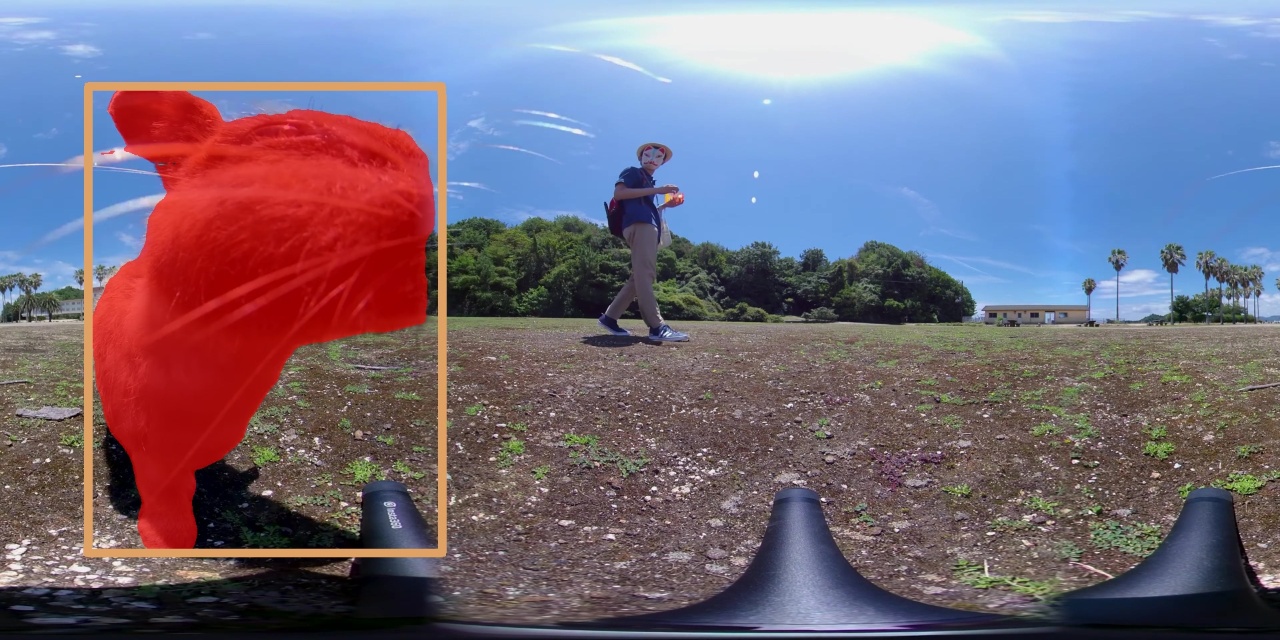}
\vspace{-0.2cm}\\
    {\rotatebox{90}{\parbox{0.09\linewidth}{\centering \scriptsize XMem-360}}}&\includegraphics[width=\imgw\linewidth,height=\imgh\linewidth]{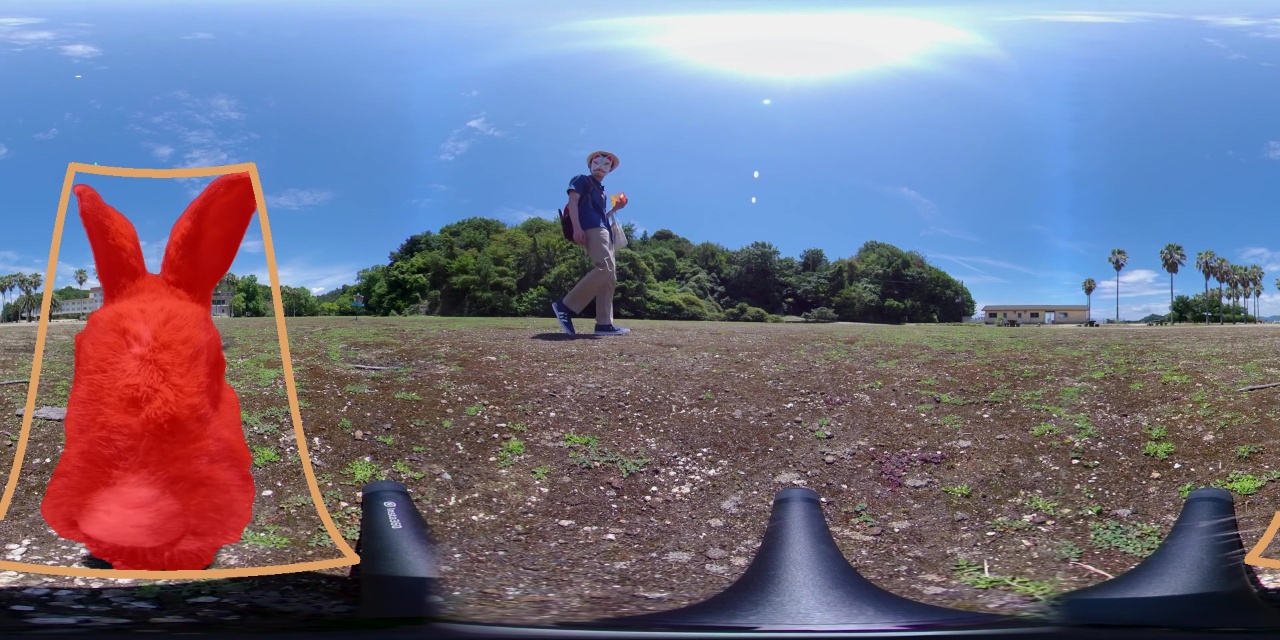}&
    \includegraphics[width=\imgw\linewidth,height=\imgh\linewidth]{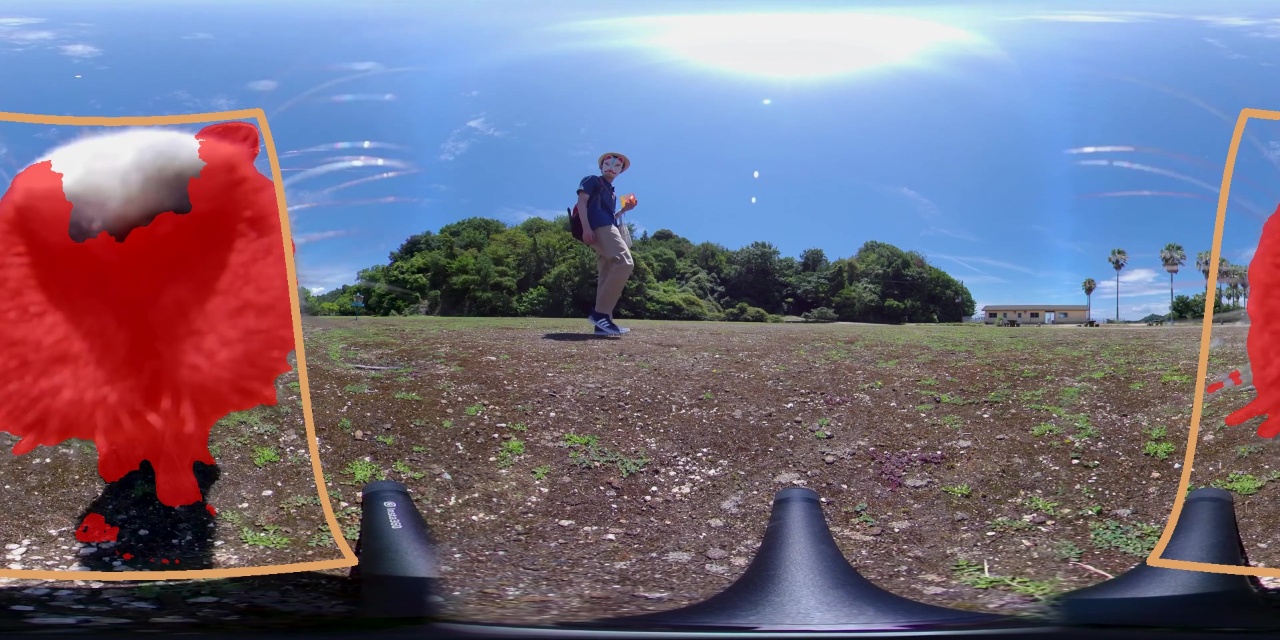}&
    \includegraphics[width=\imgw\linewidth,height=\imgh\linewidth]{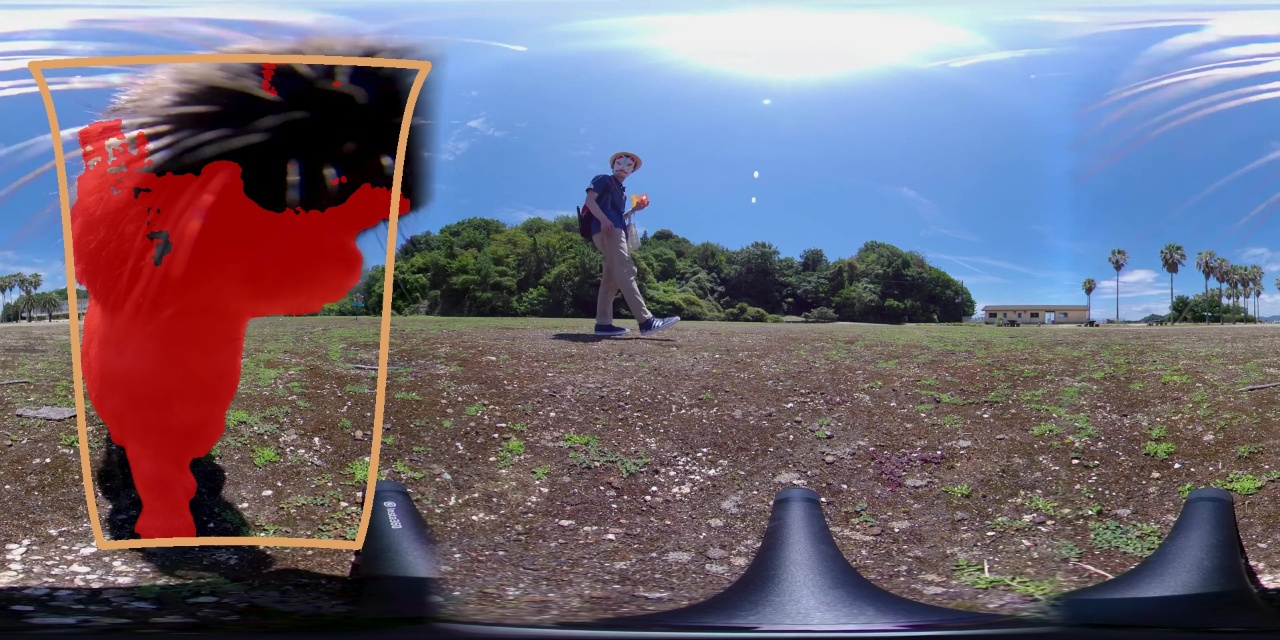}&
    \includegraphics[width=\imgw\linewidth,height=\imgh\linewidth]{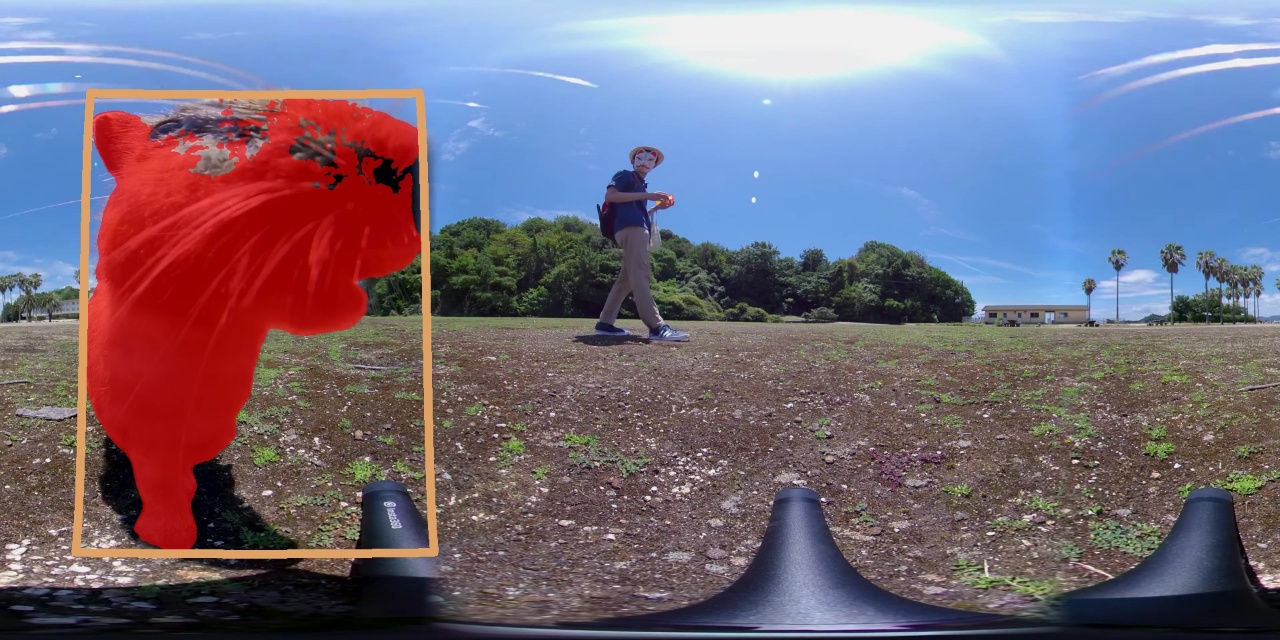}&
    \includegraphics[width=\imgw\linewidth,height=\imgh\linewidth]{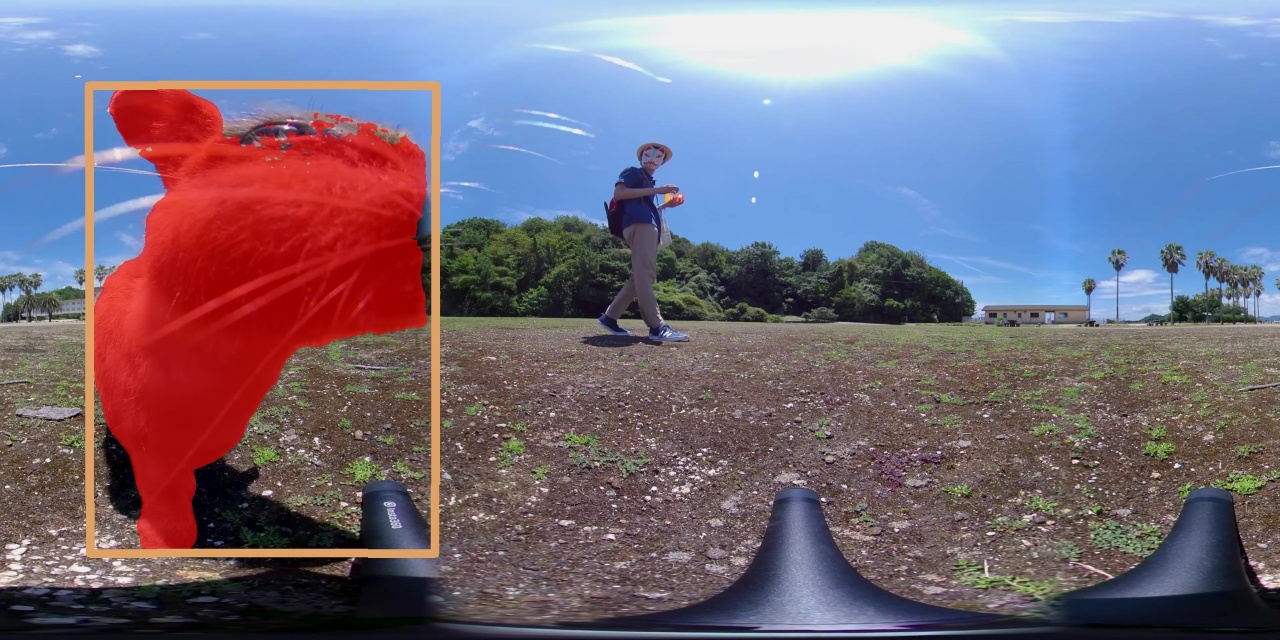}
 \vspace{-0.2cm}   \\
    {\rotatebox{90}{\parbox{0.09\linewidth}{\centering \scriptsize XMem-360$^\star$}}}&\includegraphics[width=\imgw\linewidth,height=\imgh\linewidth]{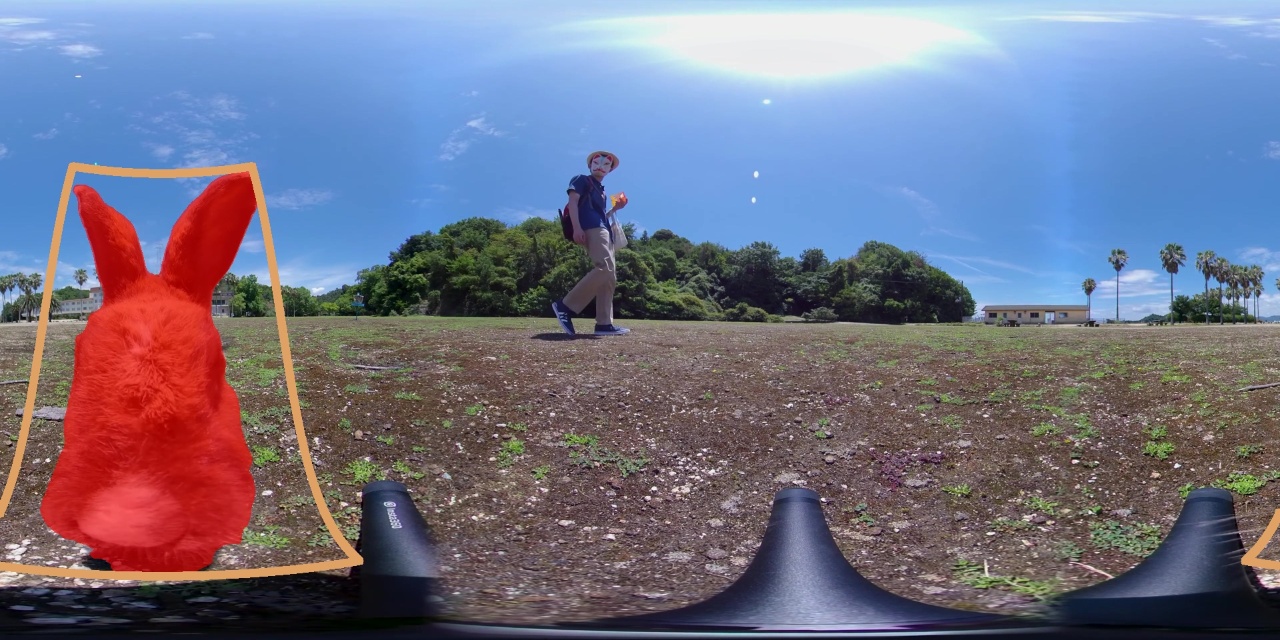}&
    \includegraphics[width=\imgw\linewidth,height=\imgh\linewidth]{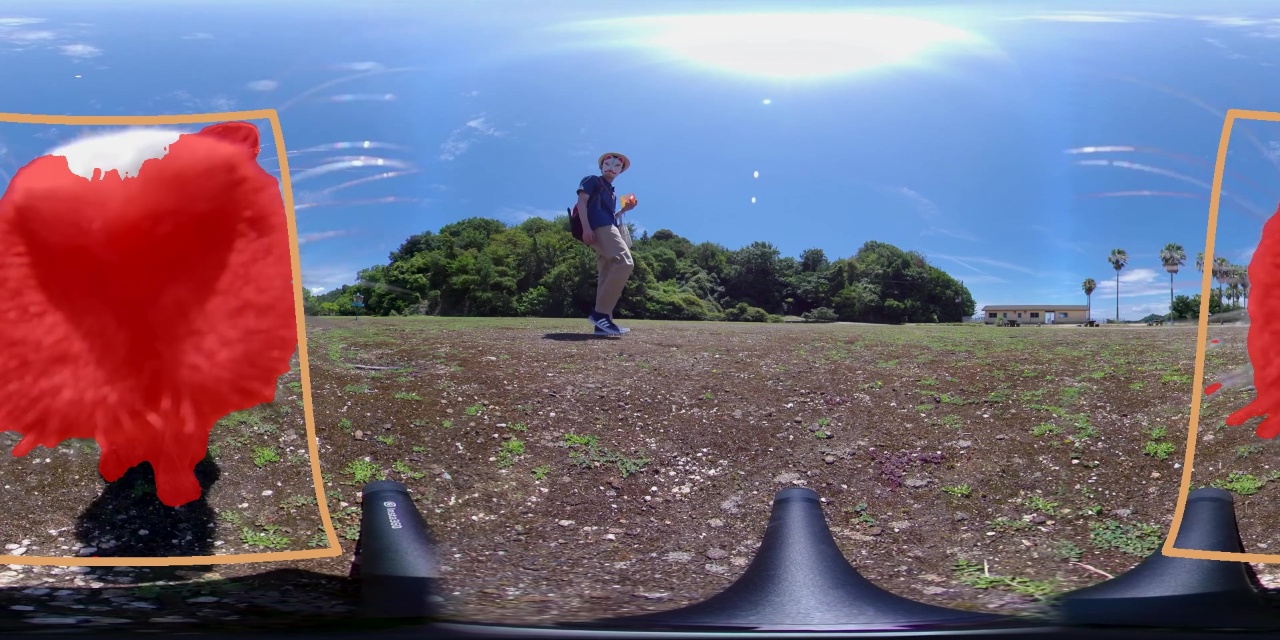}&
    \includegraphics[width=\imgw\linewidth,height=\imgh\linewidth]{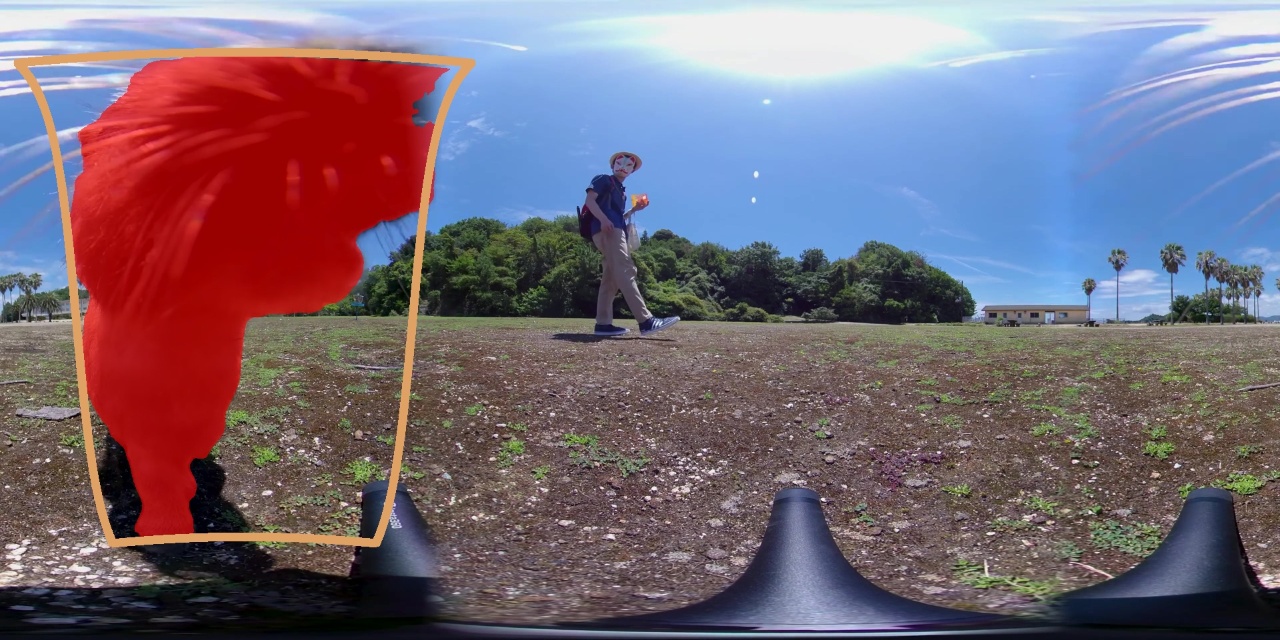}&
    \includegraphics[width=\imgw\linewidth,height=\imgh\linewidth]{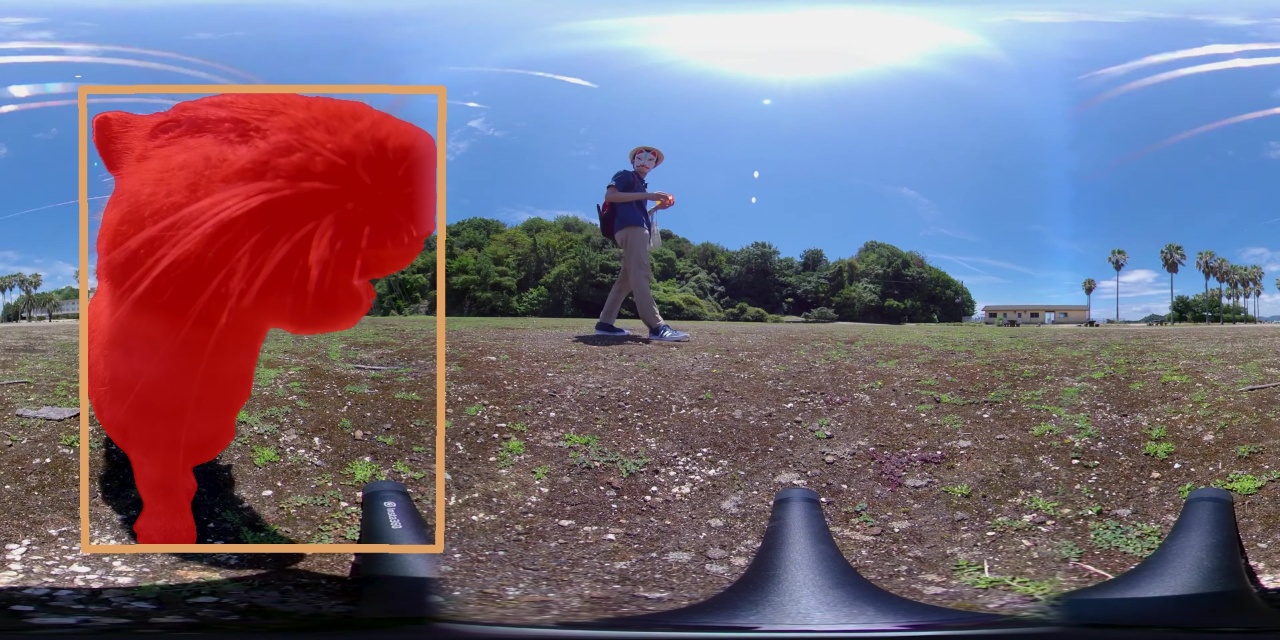}&
    \includegraphics[width=\imgw\linewidth,height=\imgh\linewidth]{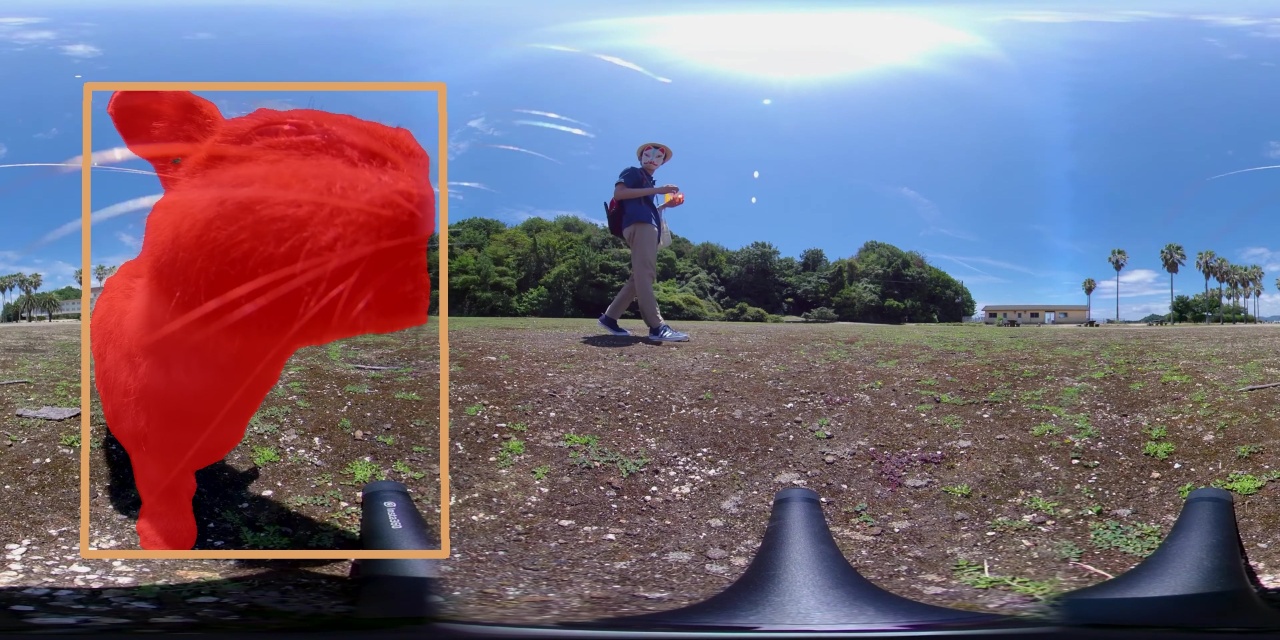} \vspace{-0.2cm}\\
    \end{tabular}
    \captionof{figure}{Comparison between the ground truth (GT) and the results from XMem\cite{xmem} and our three new baselines.
    The output \textcolor{red}{mask}s and \textcolor{bfov}{BFoV}s are shown with distinct colors.
    The 360 tracking framework enhances performance in addressing specific challenges, such as crossing border (CB) and large distortion (LD), while the retraining strategy (i.e. XMem*, XMem-360*) yields noticeable improvements on large distortion and stitching artifact (SA).
    }
    \label{fig:vos_result}
\end{table*}

\noindent \textbf{Performance on 360VOS.} 
In each testing sequence, the mask was initialized on the first frame, and all trackers subsequently inferred masks for the following frames.
According to the quantitative results reported in Table~\ref{tab:vostrackers}, memory-based methods such as STCN\cite{stcn}, XMem\cite{xmem}, and XMem++\cite{xmem2} outperform other existing trackers when dealing with omnidirectional scenes. 
Particularly, XMem++~\cite{xmem2} achieve the top performance among the 16 existing trackers, with scores of 0.579, 0.692, 0.570, and 0.687 in terms of $\mathcal{J}$, $\mathcal{F}$, $\mathcal{J}_{\text{sphere}}$, and $\mathcal{F}_{\text{sphere}}$, respectively. 
{
With our proposed framework, both XMem-360 and XMem-360$^\star$ demonstrate significant improvements over all other trackers. 
The visualized results are presented in Figure~\ref{fig:vos_result}.
}

\begin{figure*}[h]
    \centering
    \captionsetup[subfloat]{labelfont={rm,footnotesize},textfont=footnotesize} 
    \vspace{-0.3cm}
    \subfloat[Search Region Ratio]{%
        \includegraphics[width=0.33\linewidth]{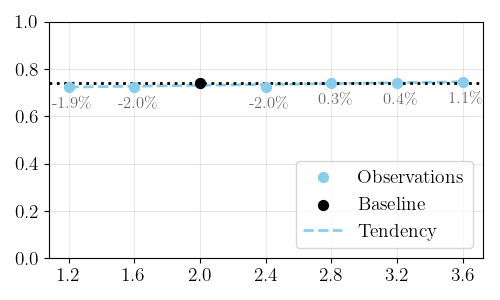}%
        \label{fig:ablmain1}%
    }
    \hfill
    \subfloat[Min Search Region]{%
        \includegraphics[width=0.33\linewidth]{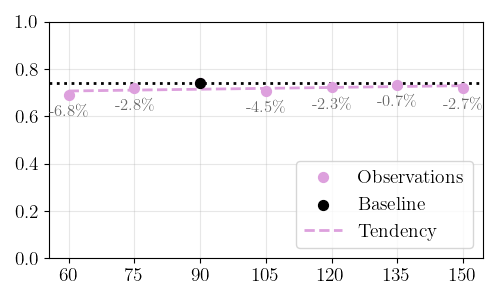}%
        \label{fig:ablmain2}%
    }
    \hfill
    \subfloat[Max Frames Loss]{%
        \includegraphics[width=0.33\linewidth]{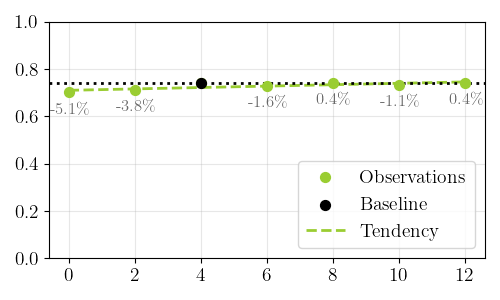}%
        \label{fig:ablmain3}%
    }
    \caption{
    {Ablation studies of the 360 tracking framework. The baseline, represented by the black dot in each plot, corresponds to XMem-360$^\star$ with specific settings. The performance is measured by average spherical metrics $(\mathcal{J} \& \mathcal{F})_{sphere}$. Other observations in each setting group are marked in different colored dots. The tendency lines are computed using the least squares method based on the baseline and all observations.}
    }
    \label{fig:ablmain}
\end{figure*}

\noindent \textbf{Tracking performance of VOS trackers on 360VOT.}
To explore the potential for cross-domain methodological enhancements that integrate the strength of VOS and VOT methods, we conducted additional experiments to evaluate the tracking performance of VOS trackers on the 360VOT benchmark.
In the implementation, we converted the VOS outputs of XMem\cite{xmem} and the 3 new baselines to obtain the tracking results in terms of (r)BBox and (r)BFoV. The comparisons with the pure VOT trackers (i.e., AiATrack-360 and SiamX-360) are demonstrated in Table~\ref{tab:result3}. 
XMem\cite{xmem} exhibits commendable performance in VOT tasks surpassing SiamX-360.
Compared to AiATrack-360, XMem-360 shows comparable performance while the tracking precision is higher. For example, the rBBox angle precision P$_{angle}$ of XMem-360 outperforms AiATrack-360 by 3.9\%. It indicates an outstanding capability of VOS trackers in handling objects with complex orientations within 360$\degree$ videos.
This is further evidenced by an obvious performance gain in the rBFoV evaluation.
Moreover, XMem-360$^\star$ achieves consistent advancements across all tracking metrics, proving the effectiveness of the training set.

\noindent \textbf{Robustness of 360 tracking framework.}
We have conducted ablation experiments to analyze the impact of different settings on our proposed 360 tracking framework in Section~\ref{sec:framework}, including 3 key parameters: SR Ratio, SR Min, and Max Loss. 
As demonstrated in Figure \ref{fig:ablmain}, the performance variations are within a narrow range (e.g., $\pm 5\%$) apart from extreme cases, showing that our framework is stable and robust across a wide range of parameter values. The baseline setting provides the best trade-off between accuracy and robustness. The extreme settings (e.g., SR Min = 60°), result in a larger drop in performance (e.g., $-6.8\%$) due to overly restrictive search regions. 
Max Loss indicates how many frames the search region remains unchanged after the tracker loses the target.  
Disabling this mechanism entirely (e.g., Max Loss = 0) results in more significant performance degradation (e.g., $-5.1\%$), highlighting the importance of the search region updating strategy in our framework.

\subsection{Efficacy of 360VOTS Training Sets}\label{sec:efficacy}
{
360VOTS fills in the absence of training data for omnidirectional tracking, as mentioned in Section \ref{sec4c}. Firstly, we retrained XMem~\cite{xmem} on a combined dataset that consists of 360VOTS training sets, YouTubeVOS\cite{ytvos2018}, and DAVIS\cite{davis2017}. According to Table~\ref{tab:vostrackers}, the retrained baseline XMem$^\star$ show a noticeable improvement of 1.7\%, 1.1\%, 1.4\%, and 1.2\% compared to XMem++\cite{xmem2} in terms of $\mathcal{J}$, $\mathcal{F}$, $\mathcal{J}_{\text{sphere}}$, and $\mathcal{F}_{\text{sphere}}$.
After integrating the retrained model with our 360 tracking framework, XMem-360$^\star$ achieves the best performance. The new training data enhances models in addressing the intricate appearance variance of tracking targets in 360VOS, illustrated in Figure~\ref{fig:vos_result}. 
Besides retraining a VOS tracker, we converted the segmentation masks to obtain VOT ground truth and fine-tuned the pretrained model of LoRAT\cite{lorat} on 360VOTS training sets only for 10 epochs. 
The finetuned VOT tracker  
LoRAT$^\star$ achieves 0.495, 0.504, 0.511, and 0.526 in terms of S$_{dual}$, P$_{dual}$, $\overline{\mbox{P}}_{dual}$, and P$_{angle}$ in 360VOT, which outperforms LoRAT\cite{lorat} 7.38\%, 7.69\%, 4.93\%, and 4.57\% accordingly.
Even though only conducted rough finetuning, LoRAT$^\star$ shows competitive performance compared to AiATrack-360.
The consistent advancements of XMem$^\star$ and LoRAT$^\star$ verify the efficacy of training data in both 360VOS and 360VOT, matching the objectives on the newly proposed datasets.
}

\section{Discussion and Conclusion} 
In this paper, we explore omnidirectional visual object tracking and segmentation. We take advantage of BFoV to represent target positions on 360$\degree$ images and extract less-distorted searching regions. The BFoV representation thus becomes the wheel in our 360 tracking framework that can adopt arbitrary classical local visual trackers to perform visual object tracking or segmentation in omnidirectional videos. The integration of existing algorithms into our 360 tracking framework demonstrates promising results in handling the intricate challenges posed by 360VOTS.
In addition, we introduced a comprehensive benchmark dataset to facilitate the study of both omnidirectional visual object tracking and segmentation. The newly available training data also provides an effective way to enhance the tracker's performance. 

Nevertheless, there still remains a big room for improvement and we want to discuss some promising directions here. One of them is the development of long-term omnidirectional tracking algorithms. The trackers enhanced by our tracking framework are still classified as short-term trackers technically. As target occlusion is a noticeable attribute of 360VOTS, the long-term tracker capable of target relocalization can perform better. 
Another promising direction is exploring new network architectures. The trackers exploiting network architectures~\cite{spherenet, deepsphere} tailored for omnidirectional images may be able to extract better features and correlations for robust VOT and VOS.

By introducing new representations, new datasets, new metrics, and new benchmarks, we envisage 360VOTS will not only catalyze further research on omnidirectional visual object tracking and segmentation but also promote new applications across a broad spectrum of computer vision and robotics fields.

\noindent\textbf{Acknowledgement.}
{\small
This research is partially supported by an internal grant from HKUST (R9429) and the Innovation and Technology Support Programme of the Innovation and Technology Fund (Ref: ITS/200/20FP).}

%

\bibliographystyle{IEEEtran}
\bibliography{bibs}

\vspace{-0.08cm}
\begin{IEEEbiography}[{\includegraphics[width=1in,height=1.25in,clip,keepaspectratio]{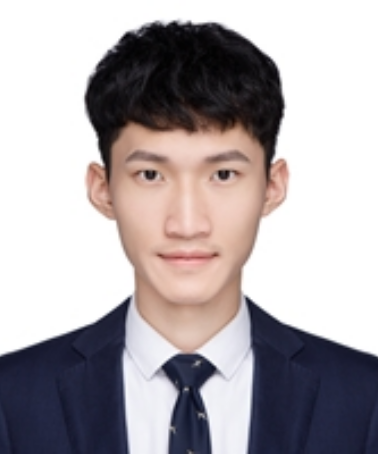}}]{Yinzhe Xu} received the B.S. degree in Integrative Systems and Design from the Hong Kong University of Science and Technology. He is currently working toward the Ph.D. degree at the same university. His research interests include omnidirectional vision, object tracking and segmentation, and SLAM.
\end{IEEEbiography}
\vspace{-0.8cm}
\begin{IEEEbiography}[{\includegraphics[width=1in,height=1.25in,clip,keepaspectratio]{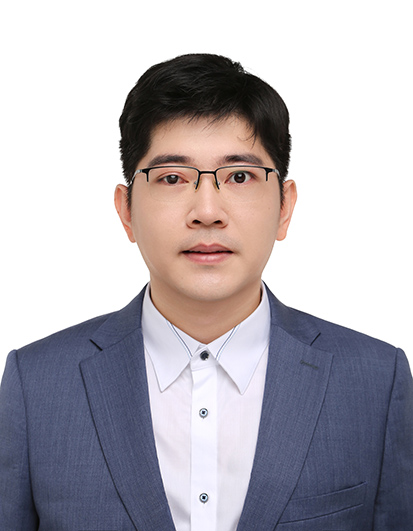}}]{Huajian Huang} was a postdoctoral fellow at the Hong Kong University of Science and Technology (HKUST). He obtained his Ph.D. degree in Computer Science and Engineering at HKUST. Before the PhD, He received his B.Eng at the Sun Yat-Sen University, China. His research focuses on omnidirectional perception, visual localization, mapping, SLAM, and 3D vision. Personal Page: https://huajianup.github.io/
\end{IEEEbiography}
\vspace{-0.8cm}
\begin{IEEEbiography}[{\includegraphics[width=1in,height=1.25in,clip,keepaspectratio]{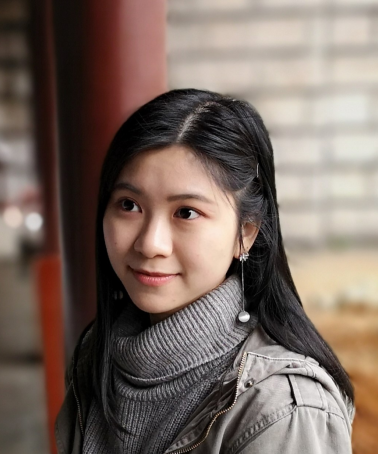}}]{Yingshu Chen} completed her Ph.D. in Computer Science and Engineering at the Hong Kong University of Science and Technology.
She received the B.Eng in Digital Media Technology from Zhejiang University and the M.Sc in Computer Science from the University of Hong Kong. Her research interests cover computer vision and computer graphics for computational design, including 2D and 3D style transfer.
\end{IEEEbiography}
\vspace{-0.8cm}
\begin{IEEEbiography}[{\includegraphics[width=1in,height=1.25in,clip,keepaspectratio]{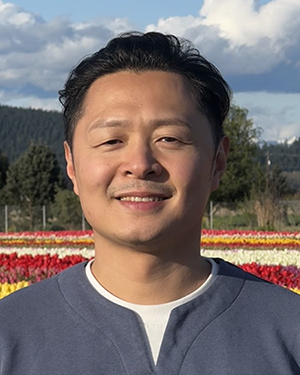}}]{Sai-Kit Yeung} is a Professor at the Division of Integrative Systems and Design, the Department of Computer Science and Engineering and the Department of Ocean Science at HKUST. Before joining HKUST, he was an Assistant Professor at the Singapore University of Technology and Design (SUTD) and founded the Vision, Graphics and Computational Design Group. During his time at SUTD, he was also a Visiting Assistant Professor at the Computer Science Department at Stanford University and the Computer Science and Artificial Intelligence Laboratory at MIT. Prior to that, he had been a Postdoctoral Scholar in the Department of Mathematics, UCLA.
Prof. Yeung’s research interests include 3D vision and graphics, content generation, fabrication, novel computational techniques and integrative systems for marine-related problems. He has published extensively in premier computer vision and graphics venues, including numerous full oral papers in CVPR, ICCV, ECCV, and AAAI. His work has received best paper honorable mention awards at ICCP 2015 and 3DV 2016. 
\end{IEEEbiography}

\vfill
\clearpage

\appendix

\subsection{Tracking on 360$\degree$ Video}
\noindent \textbf{Definition of Rotation in (r)BFoV.}
In Section \ref{seq:representation} of the main paper, (r)BFoV is defined as $[clon, clat, \theta, \phi, \gamma]$ where $[clon, clat]$ are the longitude and latitude coordinates of the object center at the spherical coordinate system, $\theta$ and $\phi$ denote the maximum horizontal and vertical field-of-view angles of the object’s occupation. The represented region of (r)BFoV on the 360$\degree$ image is formulated as:
\begin{equation*}
    I(\mbox{\footnotesize(r)BFoV} \,|\, \Omega) = \mathcal{P}(\mathcal{R}_y(clon)\cdot\mathcal{R}_x(clat)\cdot\mathcal{R}_z(\gamma)\cdot \Omega) ,
\end{equation*}
where $\mathcal{R}$ denotes the 3D rotation along the $y,x,z$ axis, $\Omega$ equals $T(\theta, \phi)$ here. The 3D rotations are defined by:
\begin{align*}
    \mathcal{R}_y(clon)&=\begin{bmatrix} cos(clon) & 0 &sin(clon)\\ 0& 1&0 \\-sin(clon) &0& cos(clon) \end{bmatrix},\\
    \mathcal{R}_x(clat)&=\begin{bmatrix} 1 & 0 &0\\ 0& cos(clat)&-sin(clat) \\0 &sin(clat)& cos(clat) \end{bmatrix},\\
    \mathcal{R}_z(\gamma)&= \begin{bmatrix} cos\gamma & -sin\gamma &0\\ sin\gamma& cos\gamma&0 \\0 &0& 1 \end{bmatrix}.
\end{align*}
\\

\noindent \textbf{Search Region Updating Strategy.}
VOT and VOS trackers may lose the target object and later re-track it back, particularly in challenging scenarios such as occlusion, fast motion, and border-crossing. To address the challenges, our framework incorporates a robust and adaptive search region updating strategy. 

For further clarification, the search region updating strategy introduced in Section \ref{sec:framework} of the main paper is outlined as pseudo-code in Algorithms \ref{alg:search_region_update}. This strategy balances efficiency and robustness by maintaining a smaller search region during short-term losses while progressively expanding it for longer-term gaps. 
\\

\begin{algorithm}[t]
\caption{Search Region Updating Strategy}
\label{alg:search_region_update}
\begin{algorithmic}[*]
\REQUIRE Predicted mask in current search region (\textit{mask}), previous search region (\textit{prev\_region}), maximum consecutive loss frames (\textit{max\_loss\_frames}), current loss count (\textit{loss\_count})
\STATE Initialize valid search region flag: $\textit{valid} \leftarrow \text{False}$
\IF{$\textit{mask}$ is valid}
    \STATE Computing the BFoV of \textit{mask}
    \STATE Updating the search region by the computed BFoV
    \STATE Reset $\textit{loss\_count} \leftarrow 0$
    \STATE $\text{valid} \leftarrow \text{True}$
\ELSE
    \IF{$\textit{loss\_count} < \textit{max\_loss\_frames}$}
        \STATE Maintain previous search region $\textit{prev\_region}$
        \STATE Increment loss count: $\textit{loss\_count} \leftarrow \textit{loss\_count} + 1$
        \STATE $\textit{valid} \leftarrow \text{True}$
    \ELSIF{$\textit{loss\_count} < 2 \times \textit{max\_loss\_frames}$}
        \STATE Gradually expand the search region based on $\textit{prev\_region}$
        \STATE Increment loss count: $\textit{loss\_count} \leftarrow \textit{loss\_count} + 1$
        \STATE $\textit{valid} \leftarrow \text{True}$
    \ELSE
        \STATE Default to a full-frame search
        \STATE $\textit{valid} \leftarrow \text{False}$
    \ENDIF
\ENDIF
\end{algorithmic}
\end{algorithm}

\begin{figure*}[t]
    \centering
    \includegraphics[width=\linewidth]{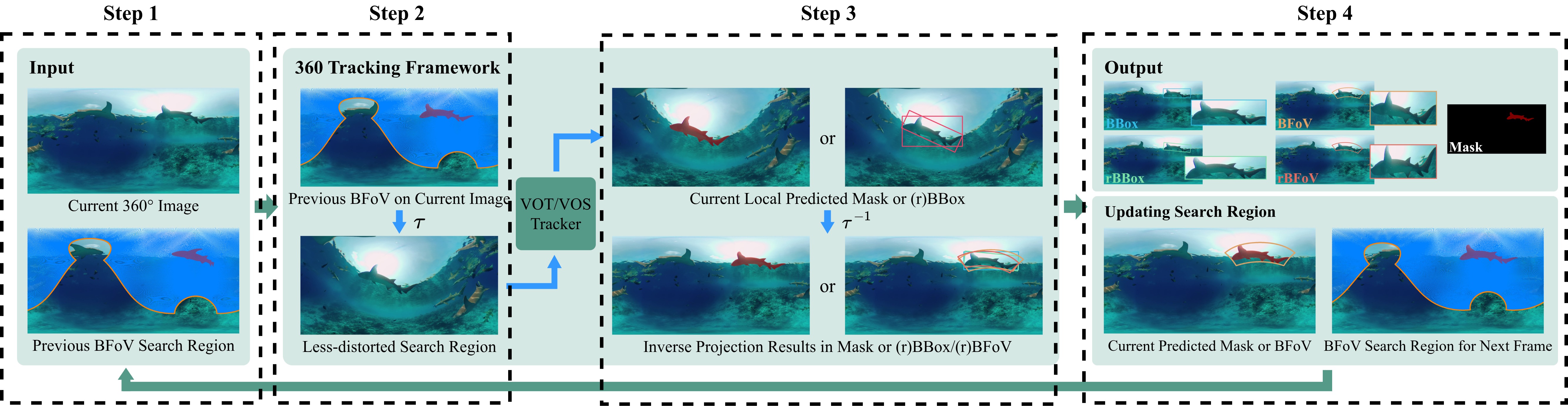}
    \caption{The 360 tracking framework that is introduced in Section \ref{sec:framework} and Figure \ref{fig:360tracking} of the main paper. The framework is separated into 4 steps here to show the intermediate visual examples in Figure \ref{fig:360fw-iter}.
    }
    \label{fig:360fw}
\end{figure*}

\begin{figure}[t]
    \centering
    \includegraphics[width=0.99\linewidth]{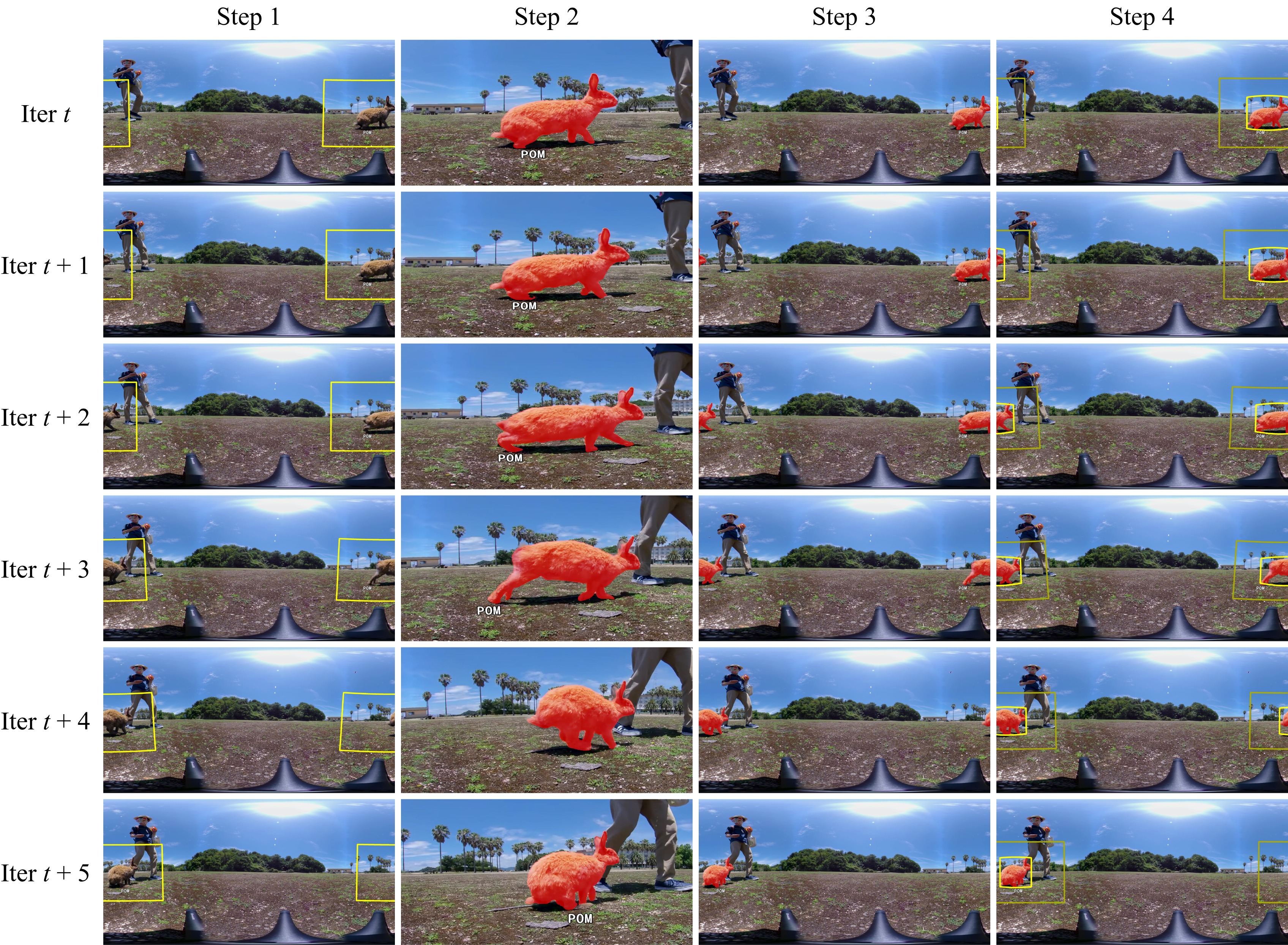}
    \caption{Intermediate visual examples of the 360 tracking framework when processing CB attributes. Sequential frames from iteration $t$ to $t+6$ are selected, during which the target object crosses the frame border. The four-step process (Figure \ref{fig:360fw}) is demonstrated: (1) The search region from the previous frame is visualized on the current frame. (2) The tracker estimates the target mask within the search region. (3) The estimated mask is reprojected onto the original frame. (4) The updated target mask and next frame’s search region are shown. These steps ensure seamless tracking across frame borders.\\
    Included sequence: Test-096.
    }
    \label{fig:360fw-iter}
\end{figure}

\begin{figure}[t]
    \centering
    \captionsetup{type=figure}
    \def\imgw{0.32}
    \def\imgh{0.16}
    
    \subfloat{\includegraphics[width=\imgw\linewidth, height=\imgh\linewidth]{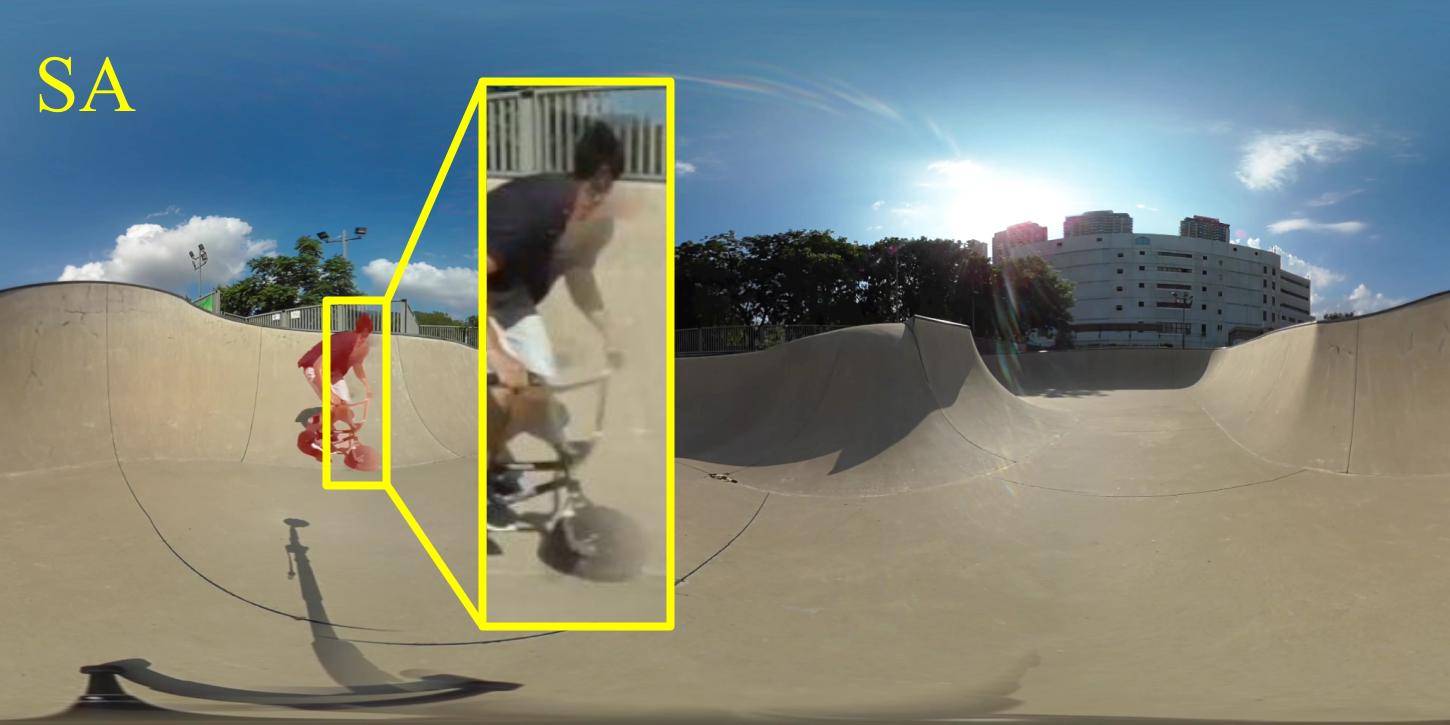}}\!
    \subfloat{\includegraphics[width=\imgw\linewidth, height=\imgh\linewidth]{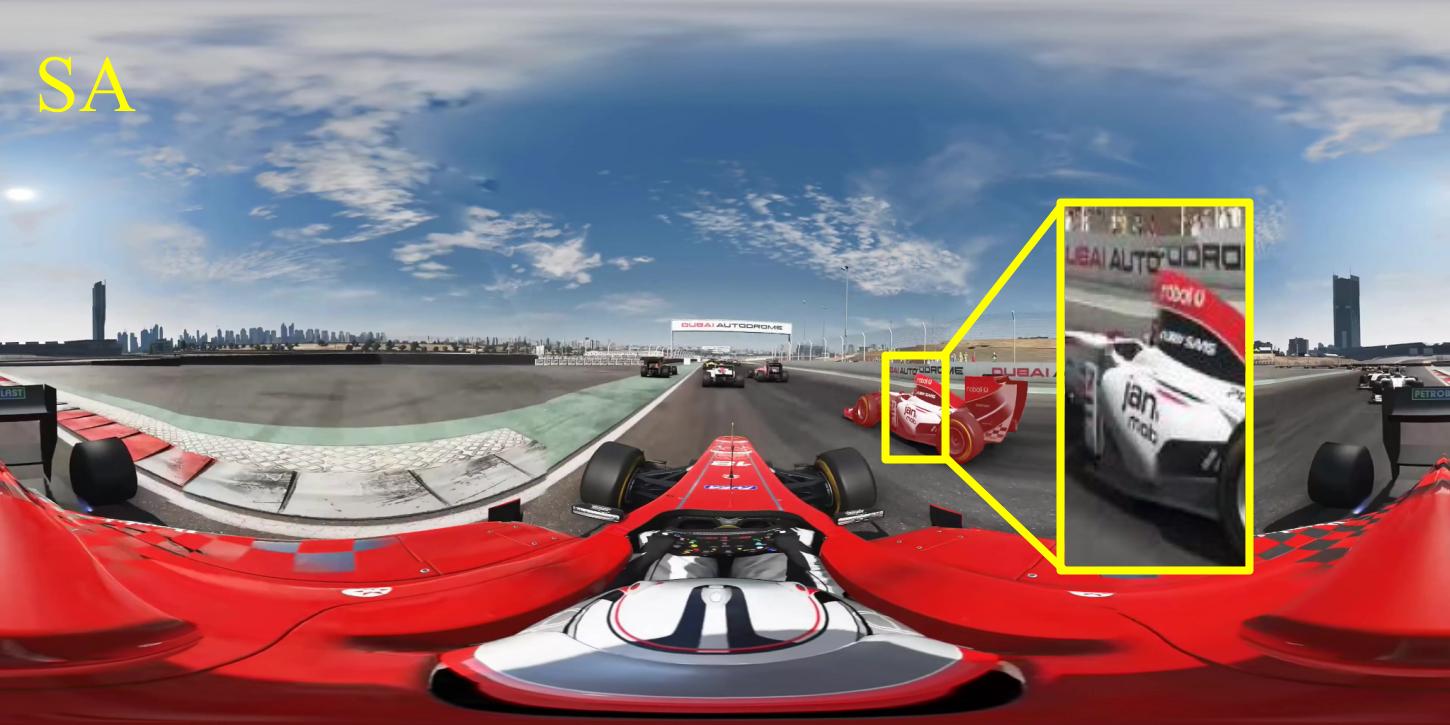}}\!
    \subfloat{\includegraphics[width=\imgw\linewidth, height=\imgh\linewidth]{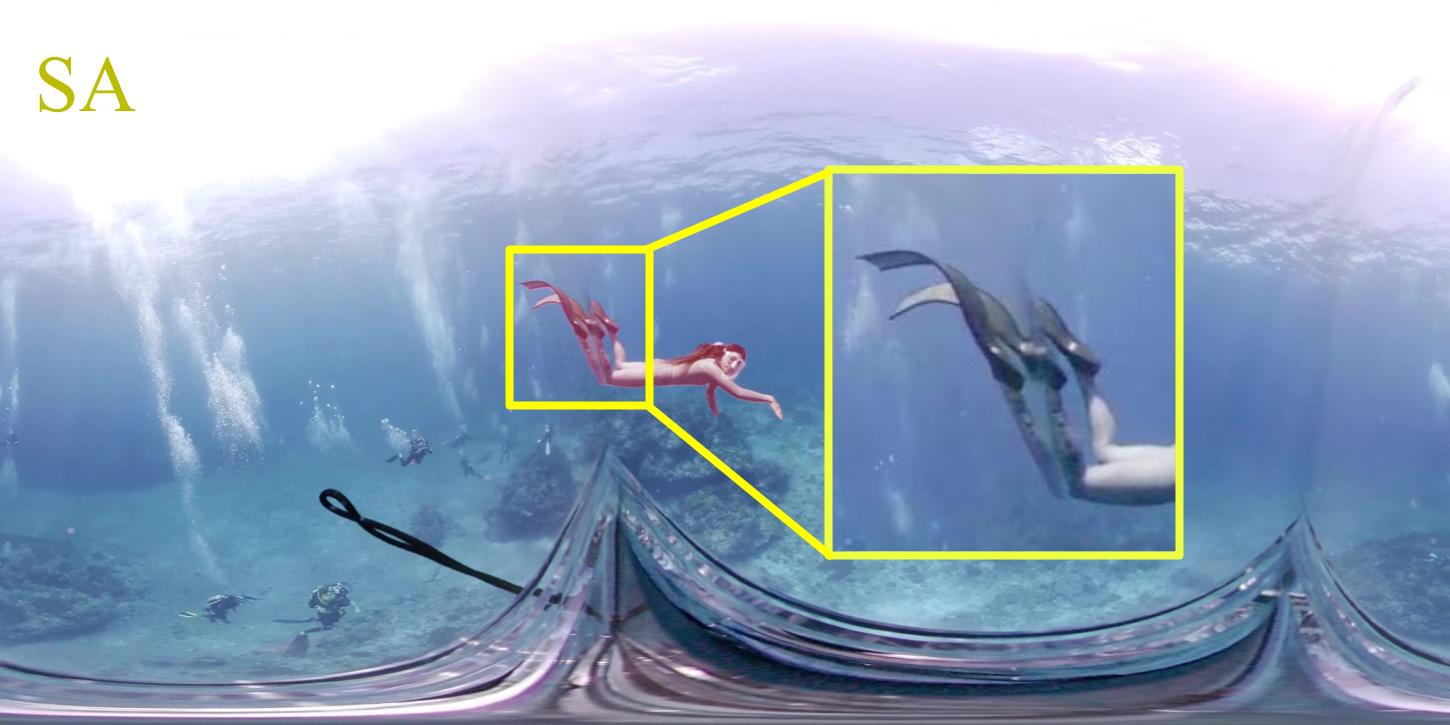}}
    \vspace{-0.9em}    \\
    \subfloat{\includegraphics[width=\imgw\linewidth, height=\imgh\linewidth]{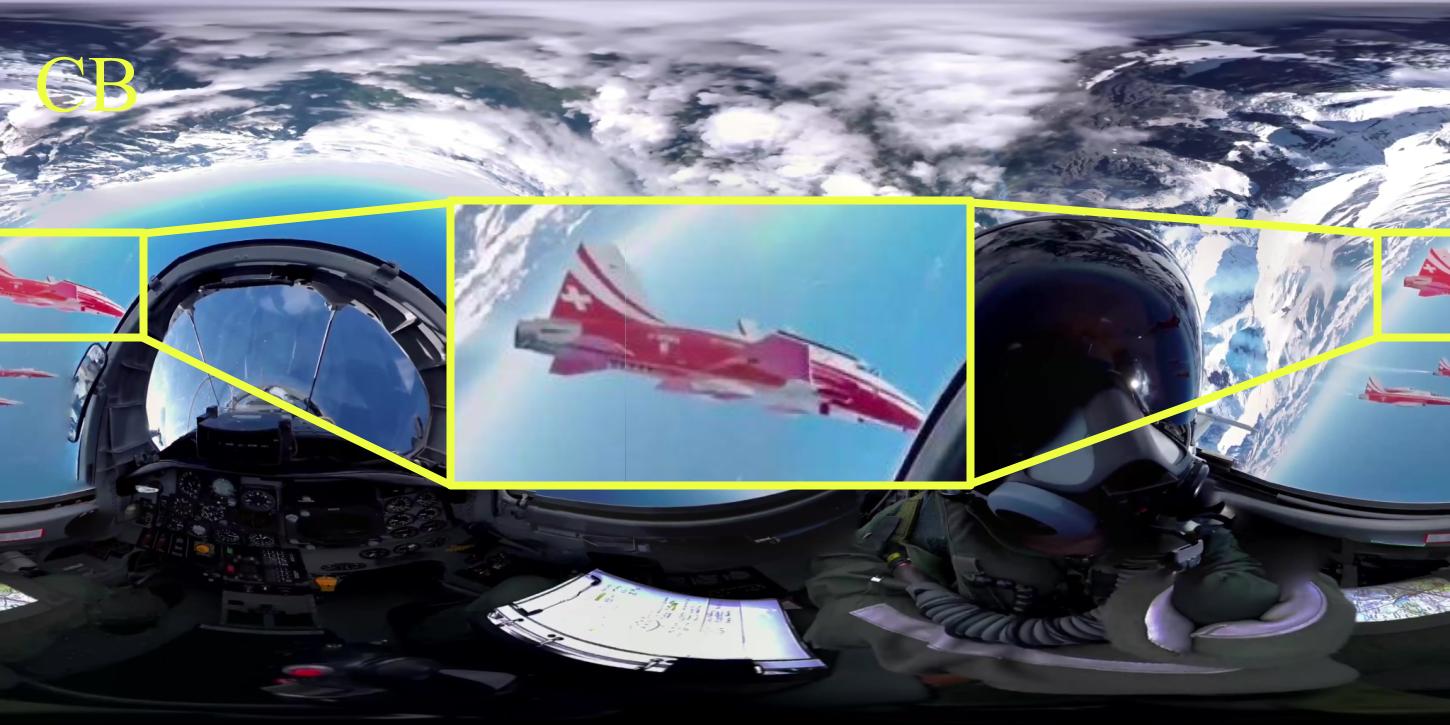}}\!
    \subfloat{\includegraphics[width=\imgw\linewidth, height=\imgh\linewidth]{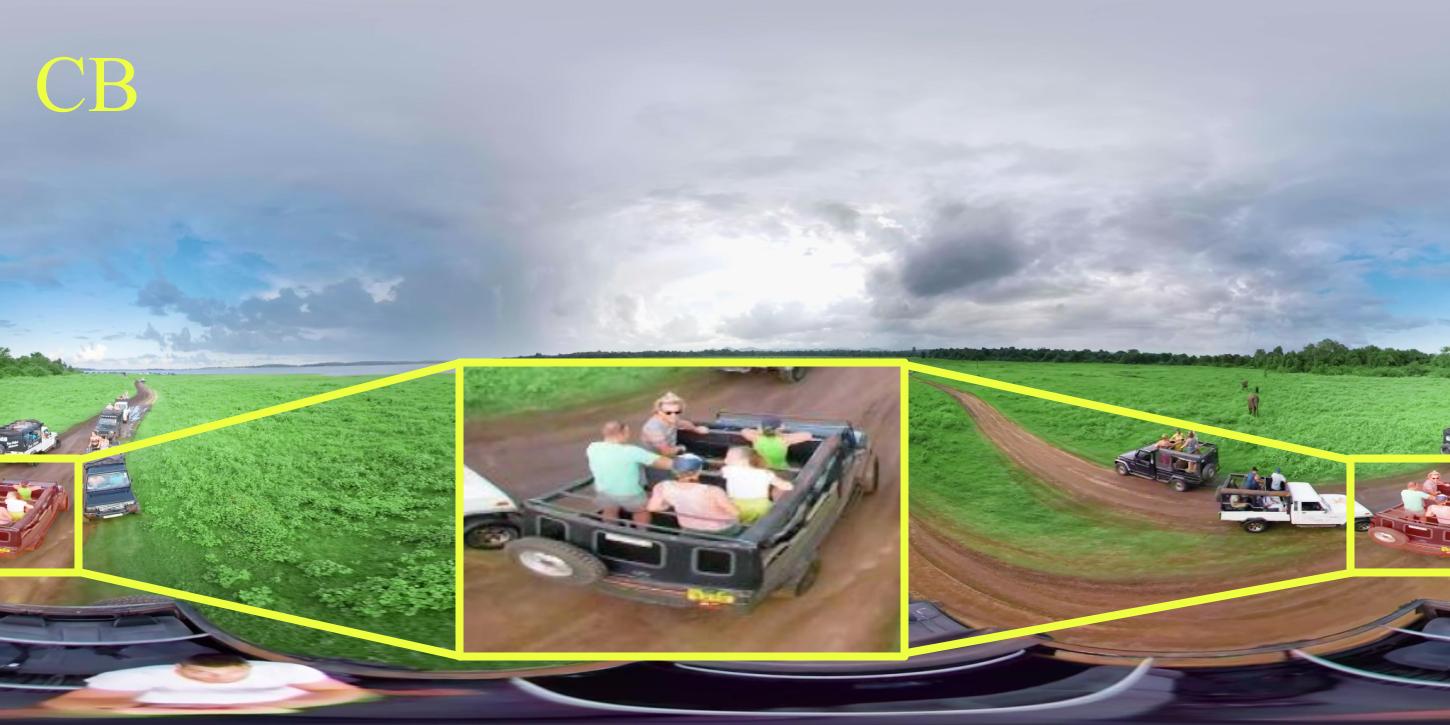}}\!
    \subfloat{\includegraphics[width=\imgw\linewidth, height=\imgh\linewidth]{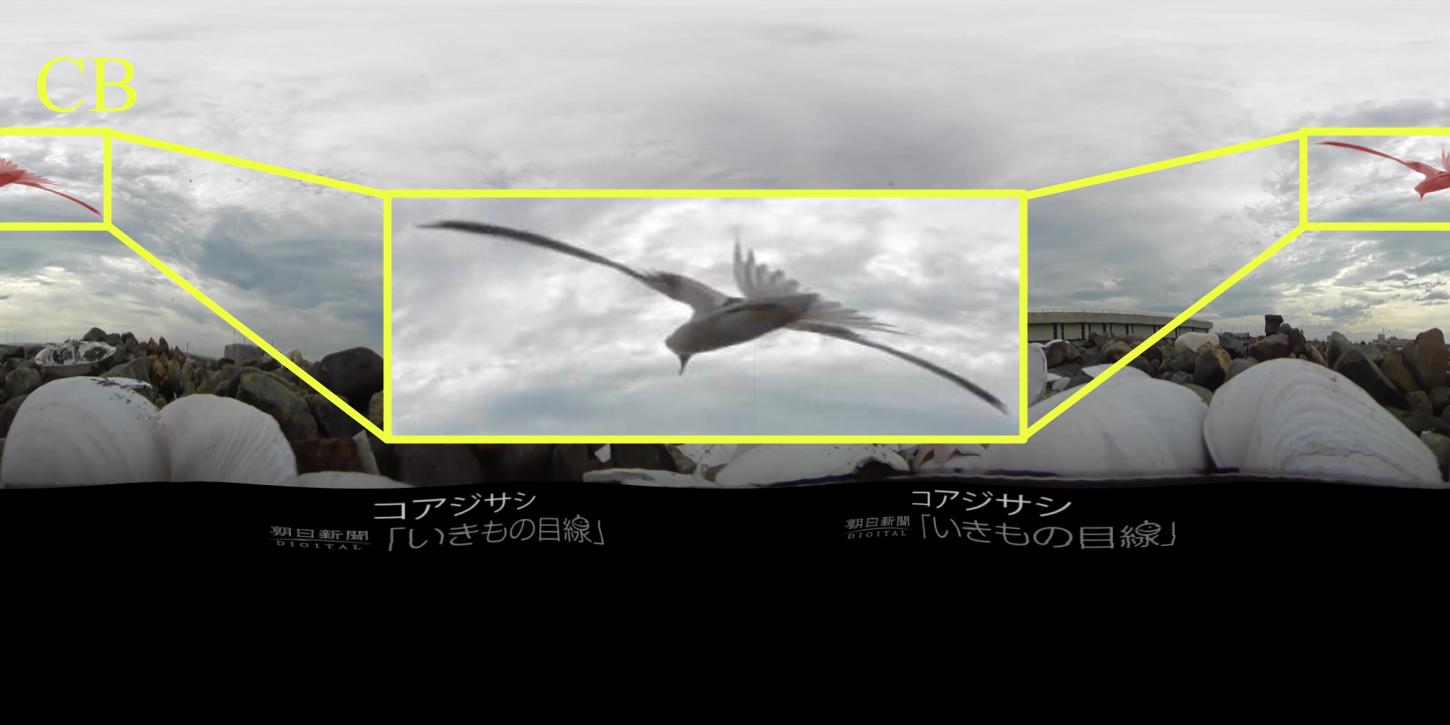}}
    \vspace{-0.9em}    \\
    \subfloat{\includegraphics[width=\imgw\linewidth, height=\imgh\linewidth]{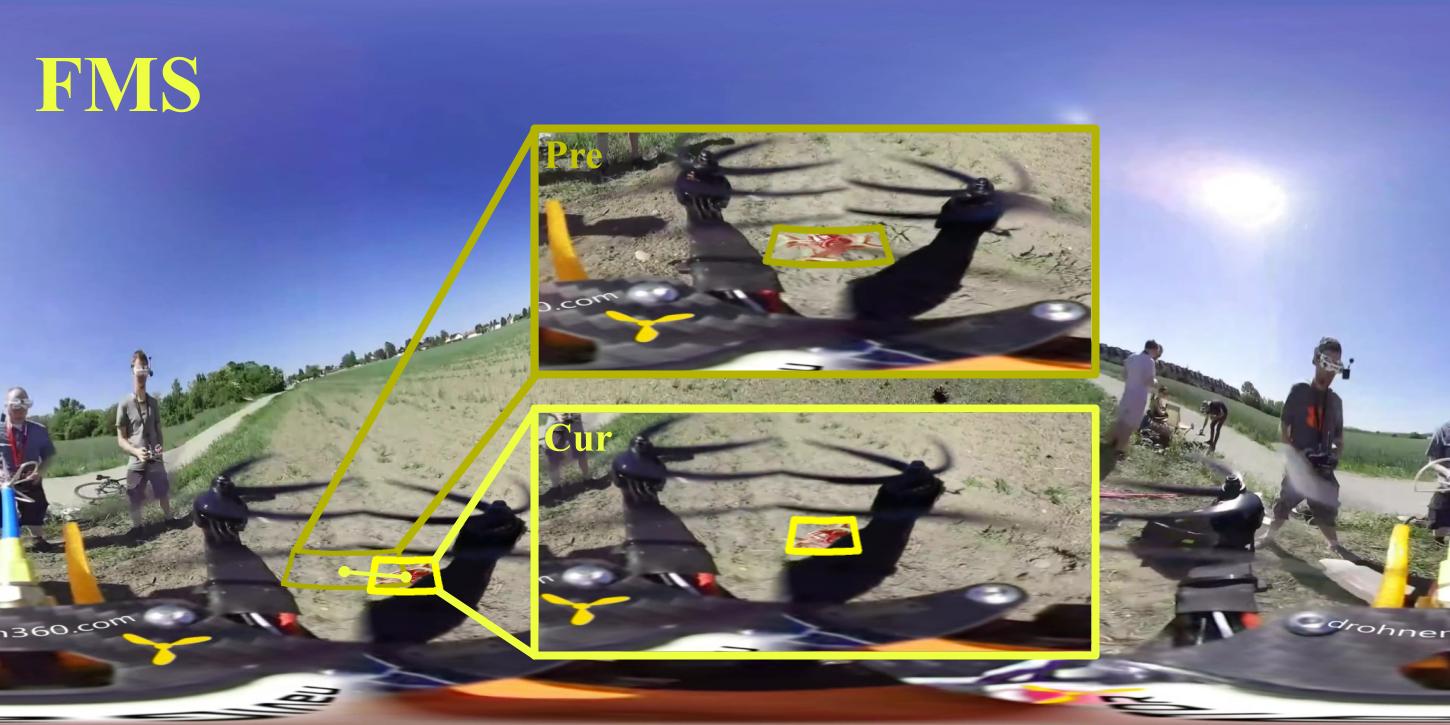}}\!
    \subfloat{\includegraphics[width=\imgw\linewidth, height=\imgh\linewidth]{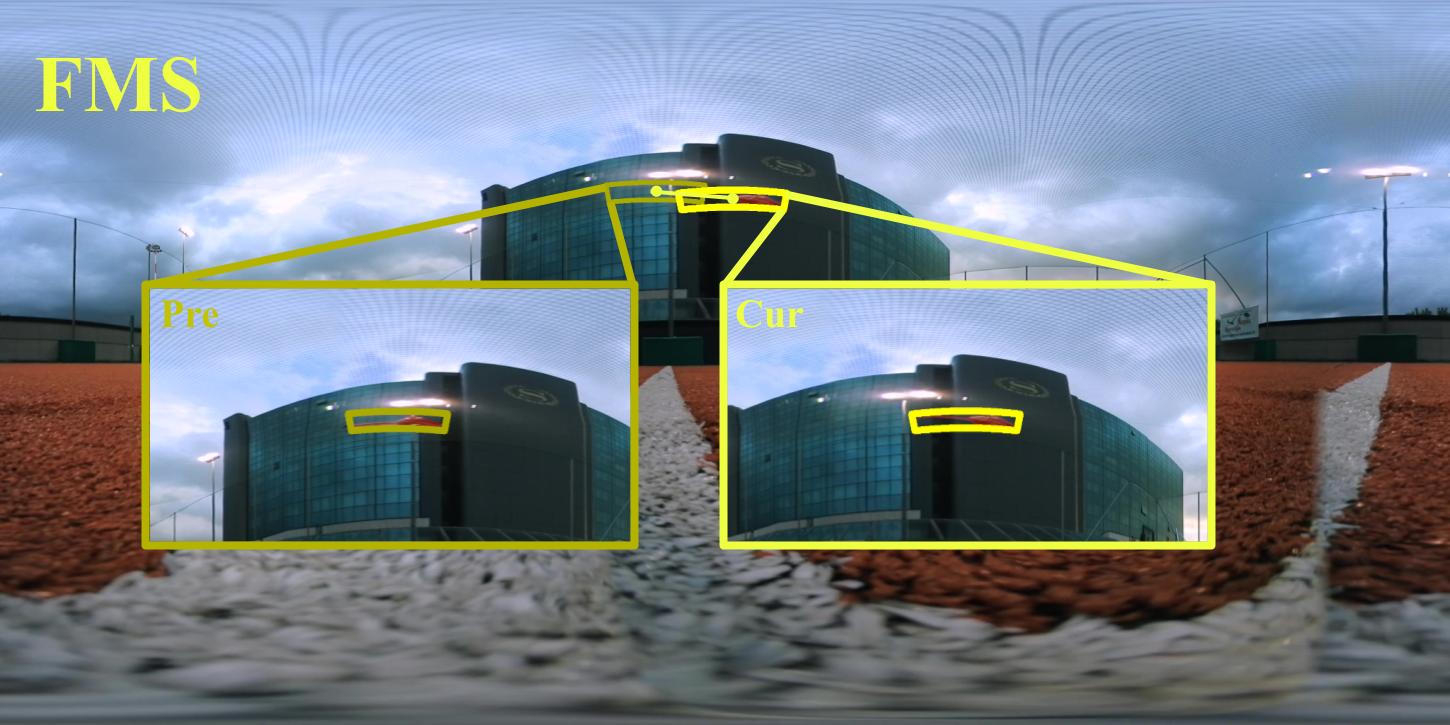}}\!
    \subfloat{\includegraphics[width=\imgw\linewidth, height=\imgh\linewidth]{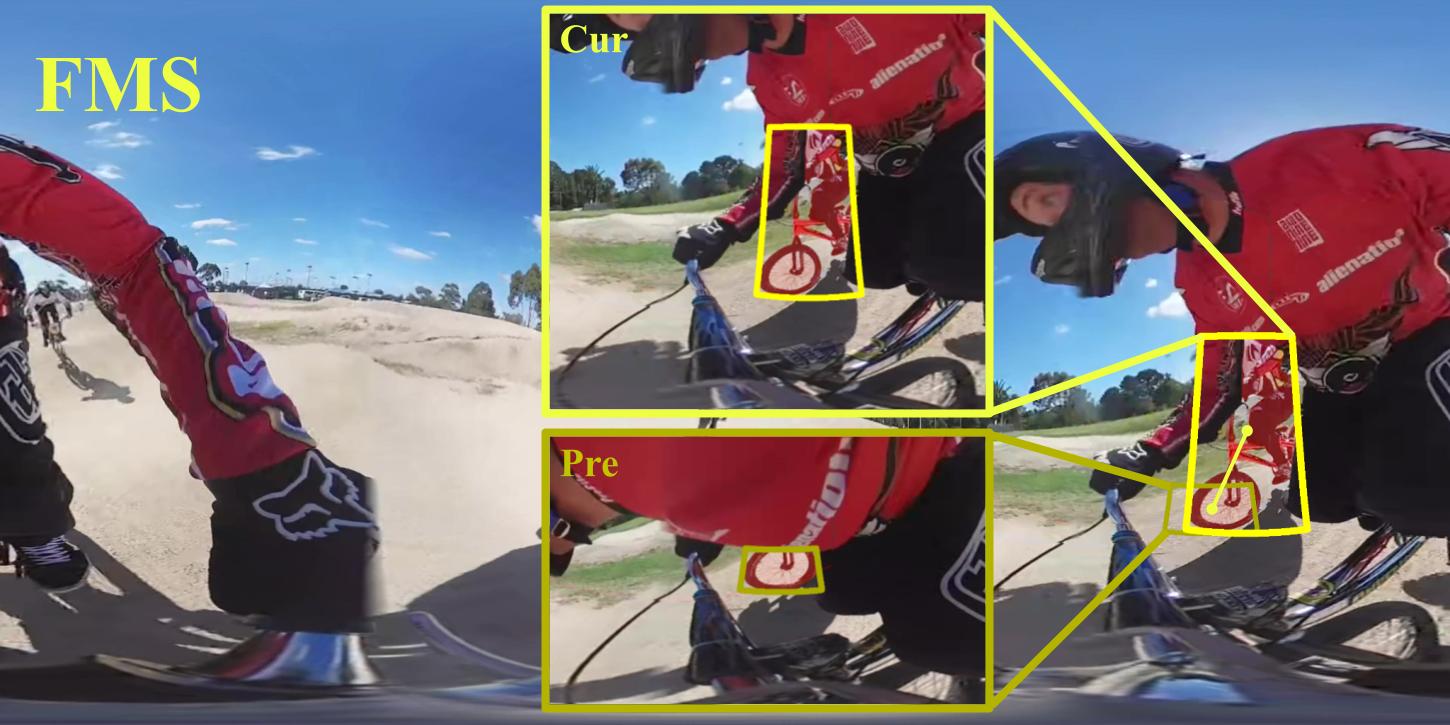}}
    \vspace{-0.9em}    \\
    \subfloat{\includegraphics[width=\imgw\linewidth, height=\imgh\linewidth]{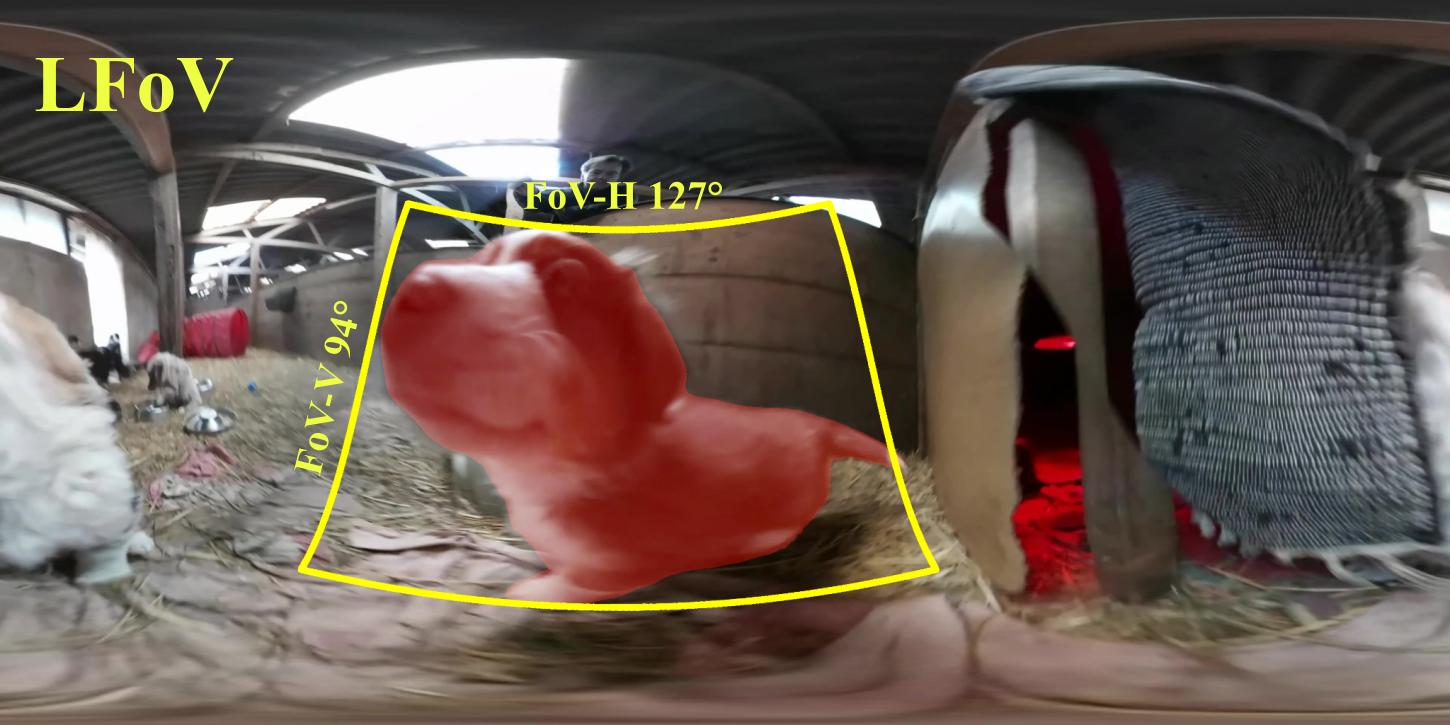}}\!
    \subfloat{\includegraphics[width=\imgw\linewidth, height=\imgh\linewidth]{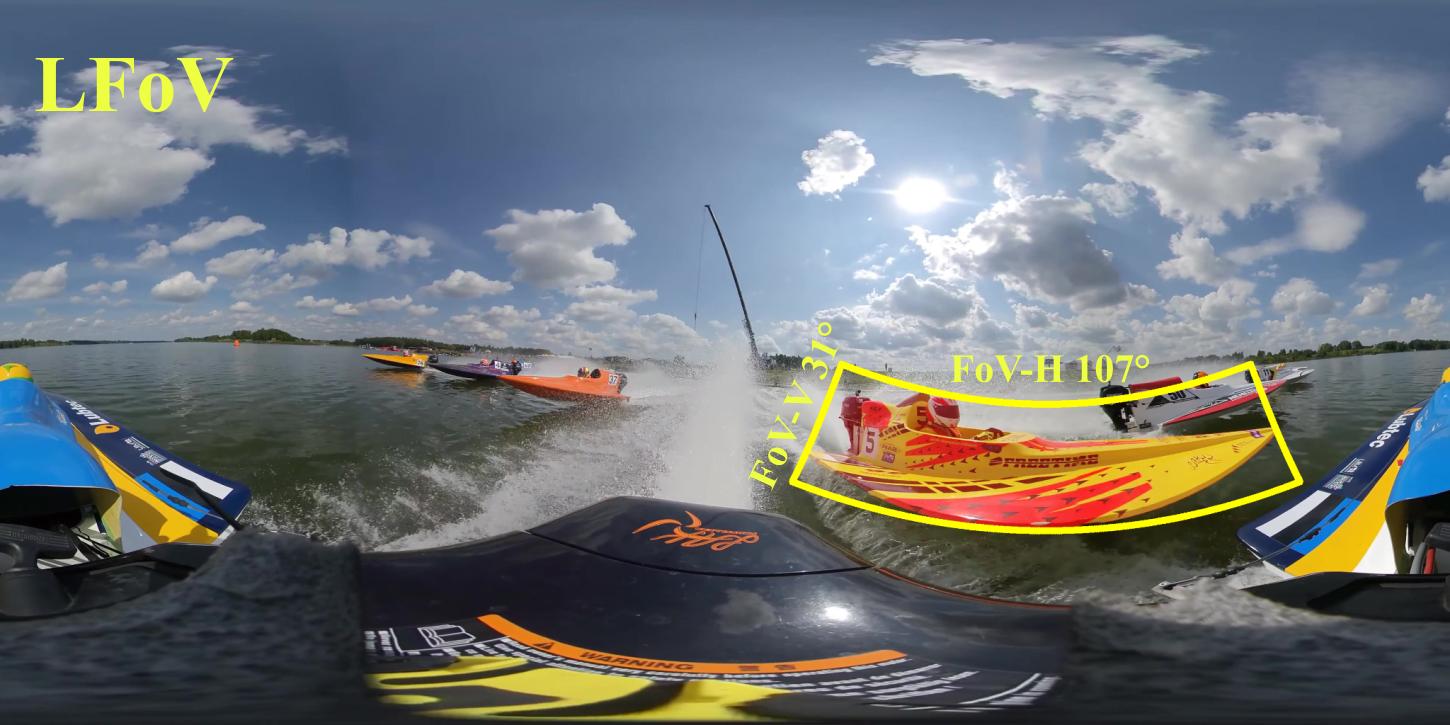}}\!
    \subfloat{\includegraphics[width=\imgw\linewidth, height=\imgh\linewidth]{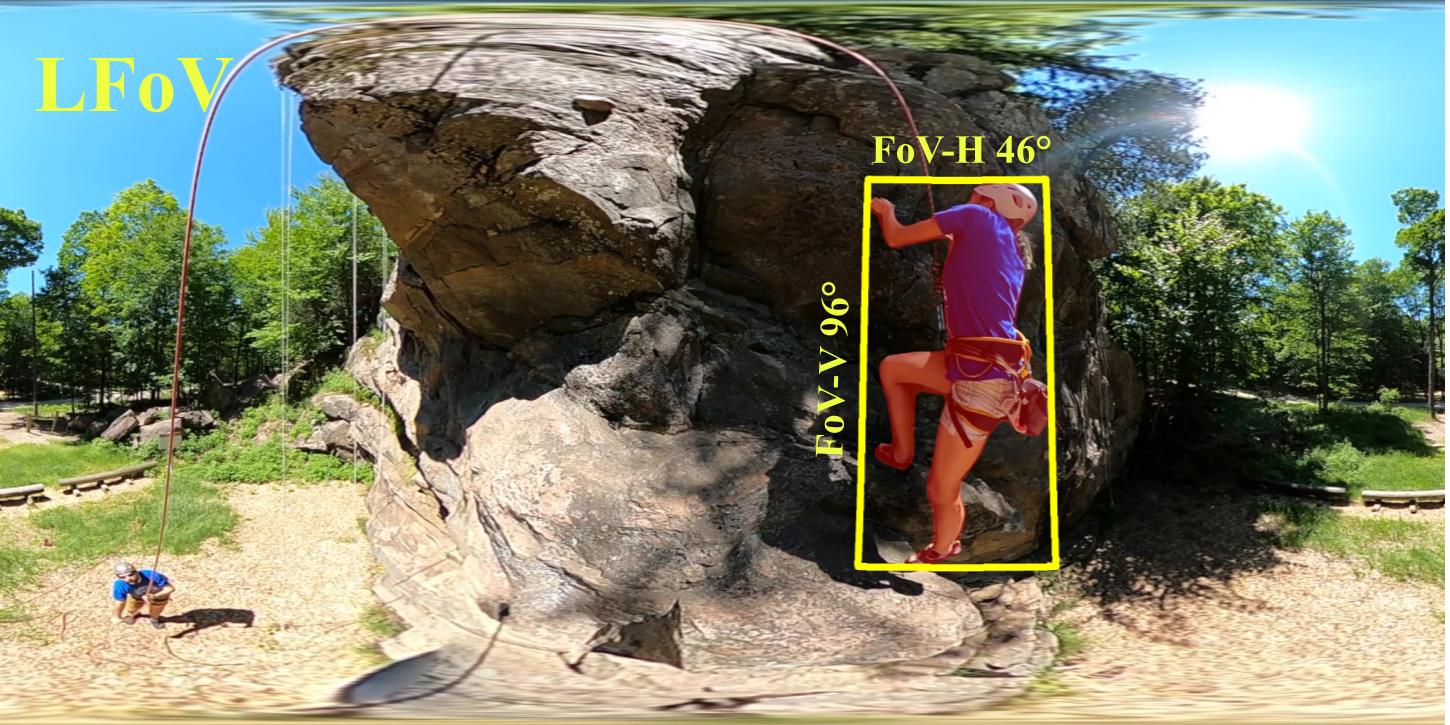}}
    \vspace{-0.9em}    \\
    \subfloat{\includegraphics[width=\imgw\linewidth, height=\imgh\linewidth]{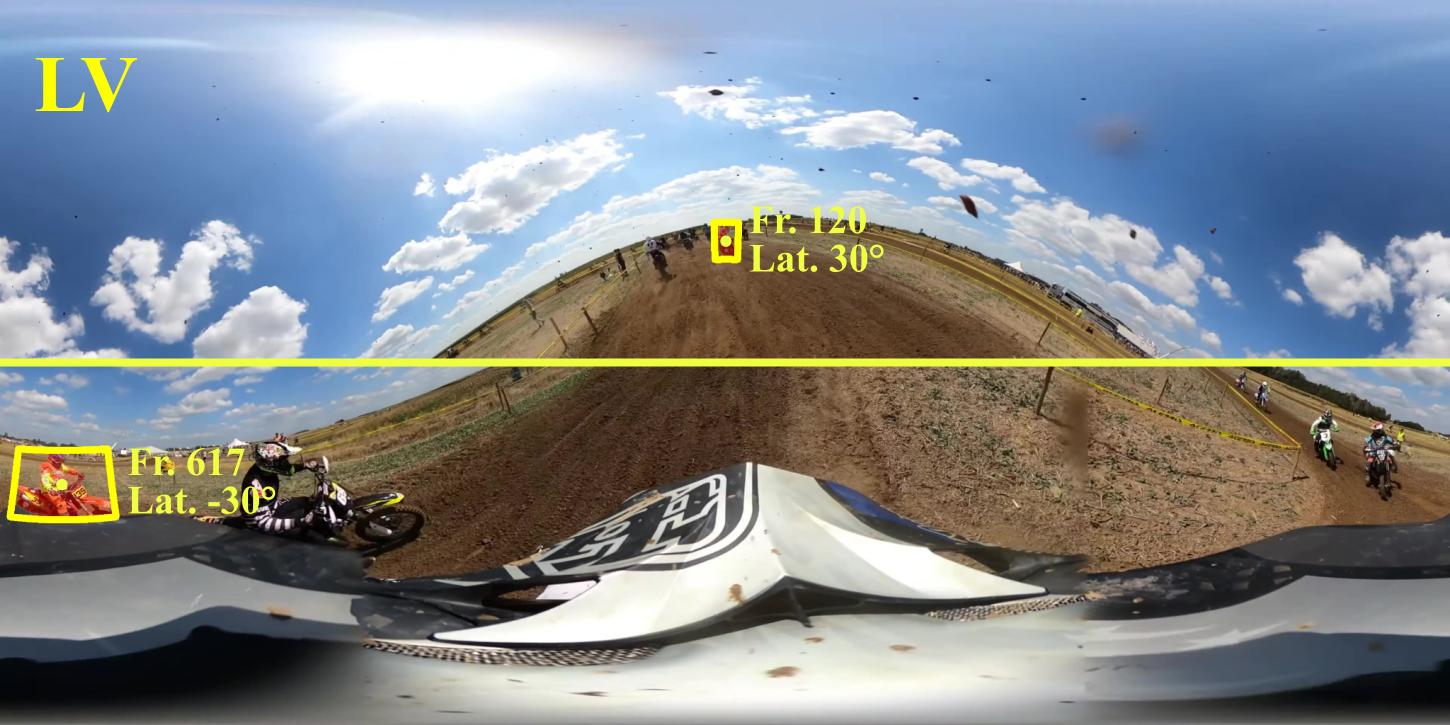}}\!
    \subfloat{\includegraphics[width=\imgw\linewidth, height=\imgh\linewidth]{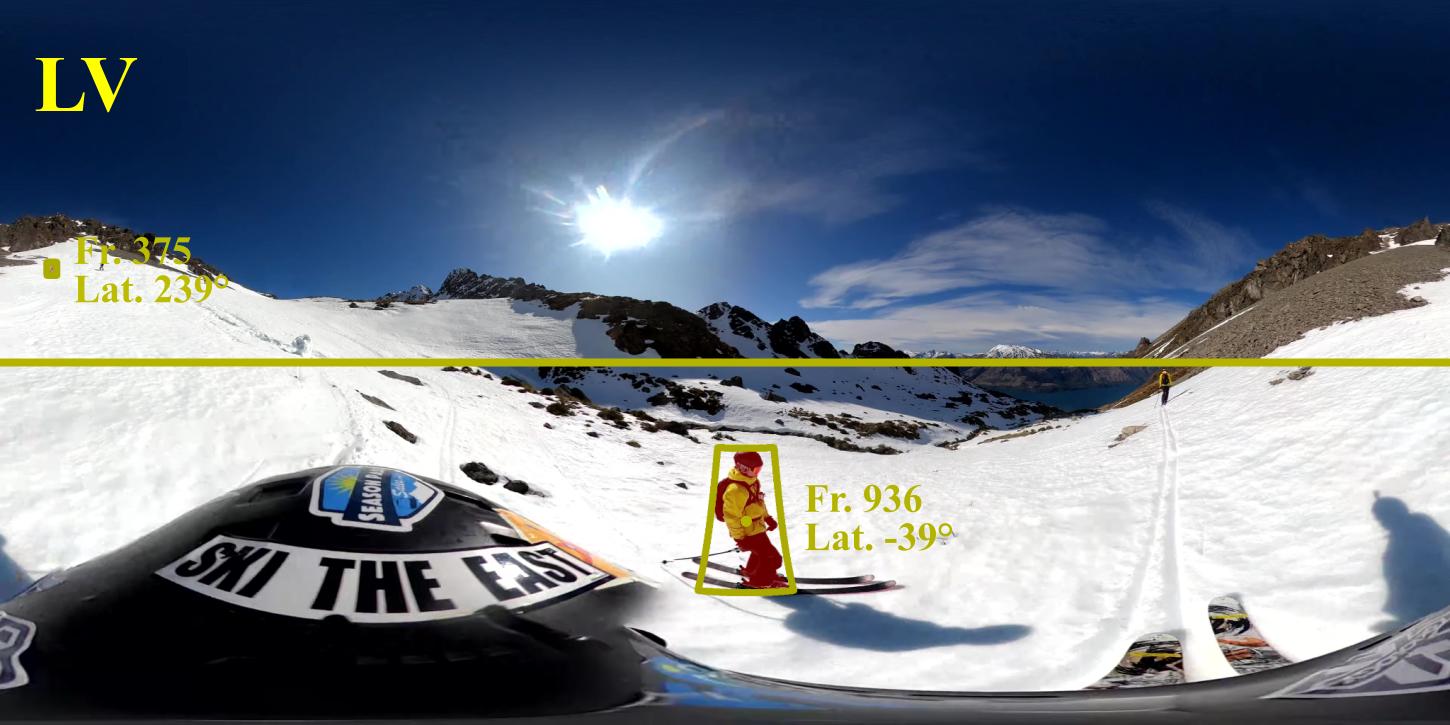}}\!
    \subfloat{\includegraphics[width=\imgw\linewidth, height=\imgh\linewidth]{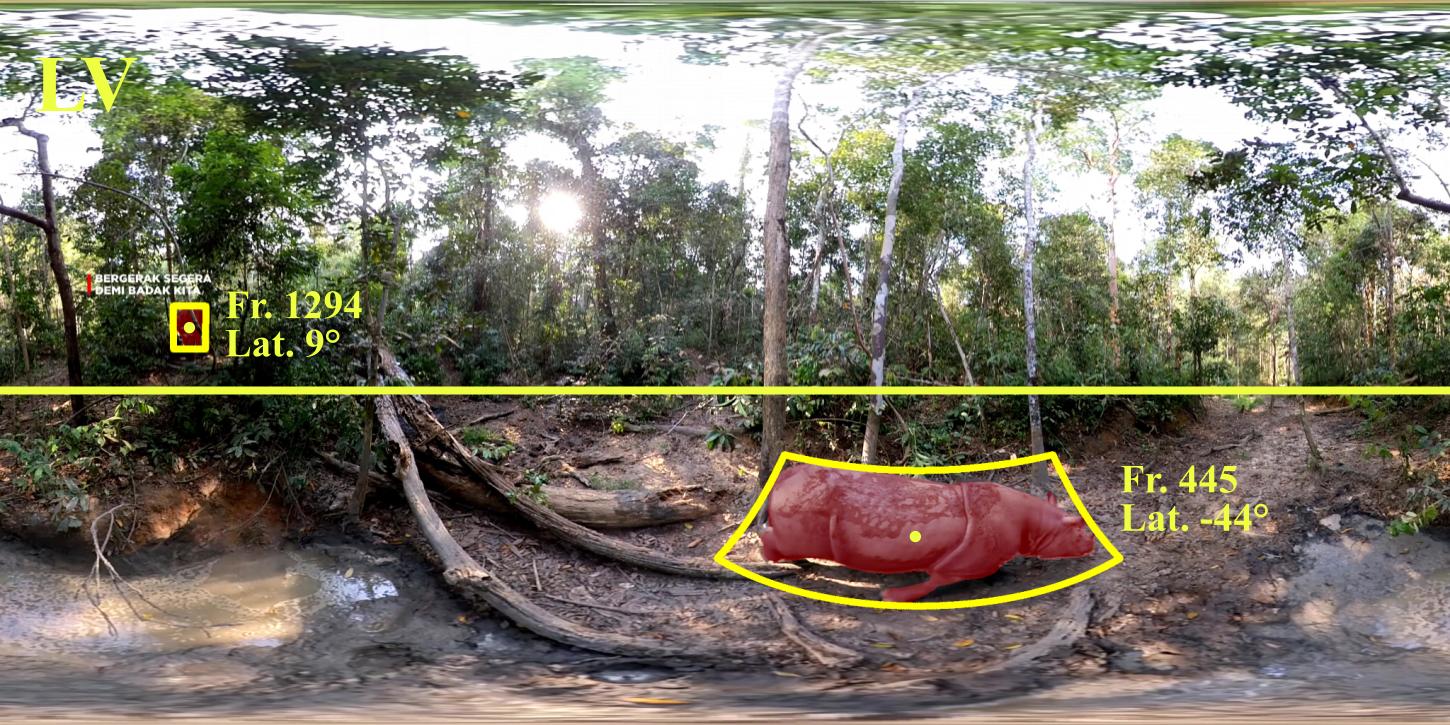}}
    \vspace{-0.9em}    \\
    \subfloat{\includegraphics[width=\imgw\linewidth, height=\imgh\linewidth]{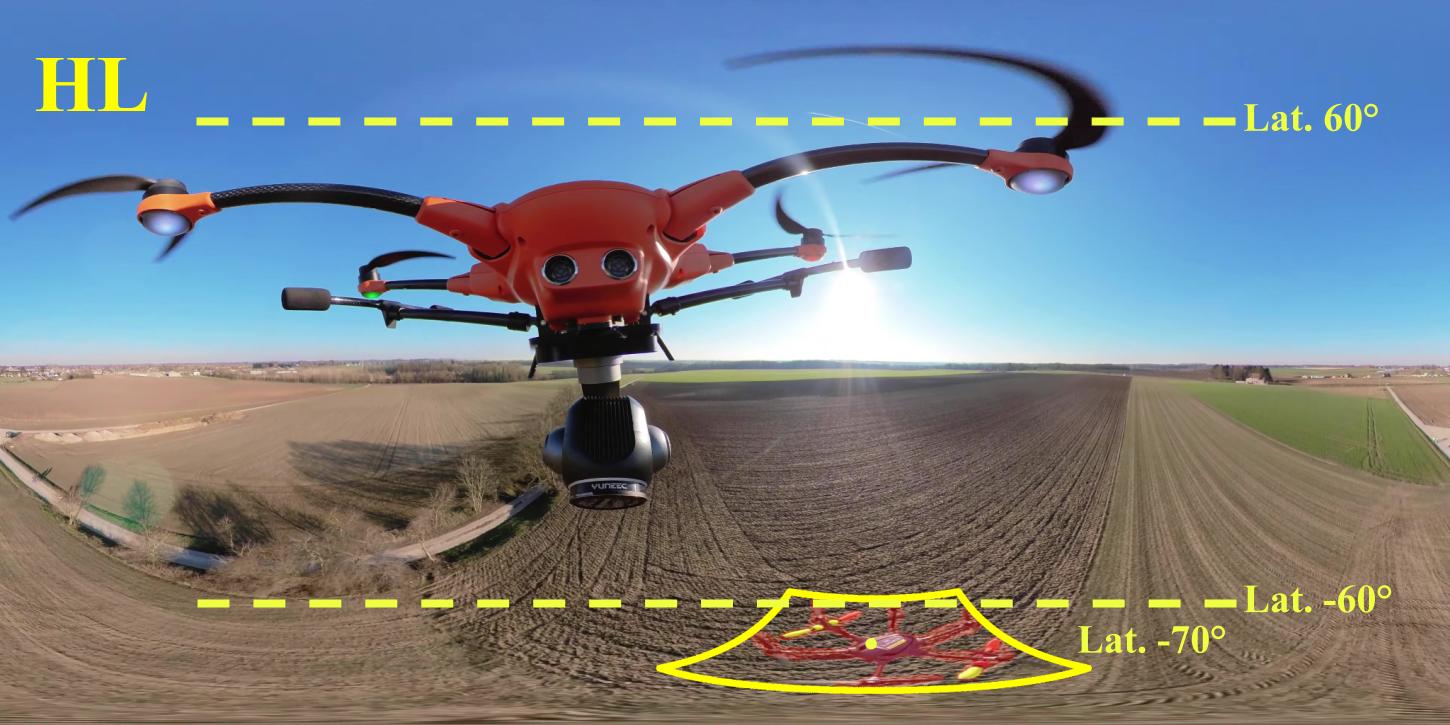}}\!
    \subfloat{\includegraphics[width=\imgw\linewidth, height=\imgh\linewidth]{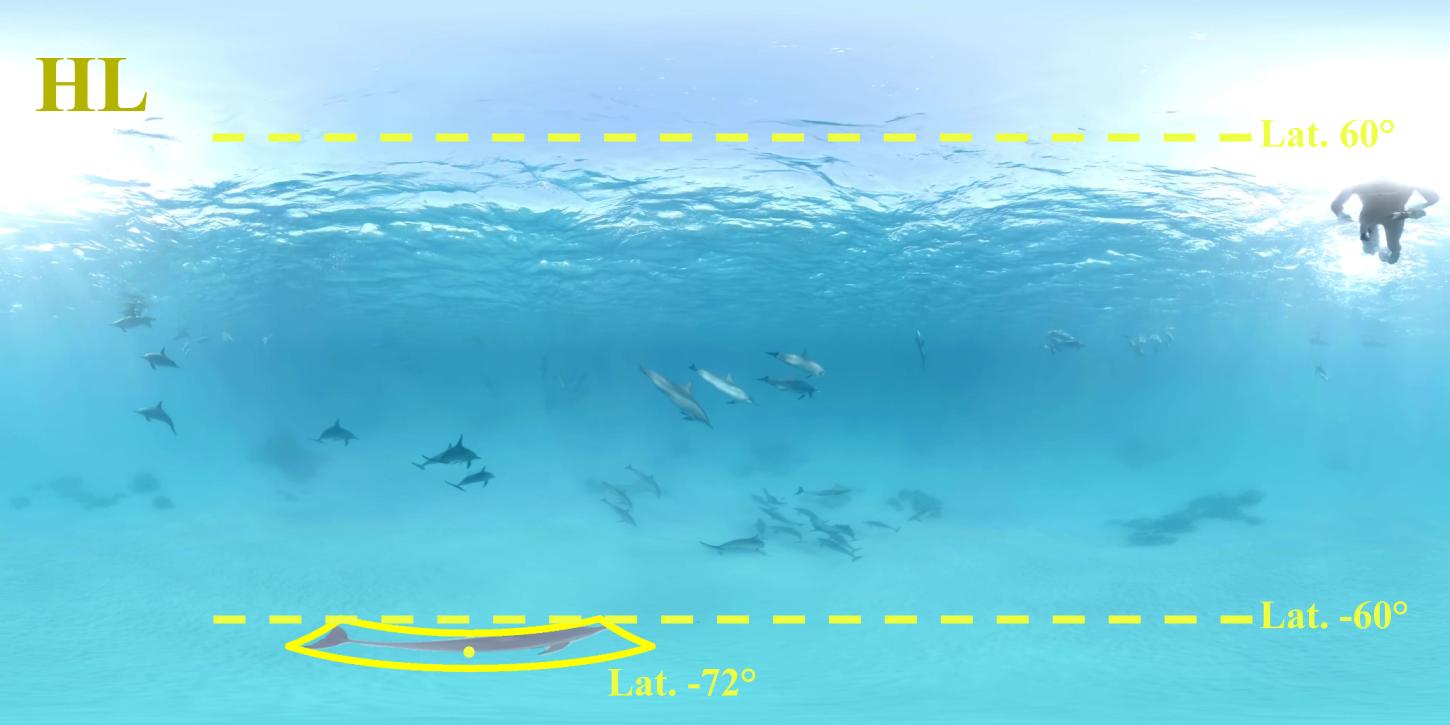}}\!
    \subfloat{\includegraphics[width=\imgw\linewidth, height=\imgh\linewidth]{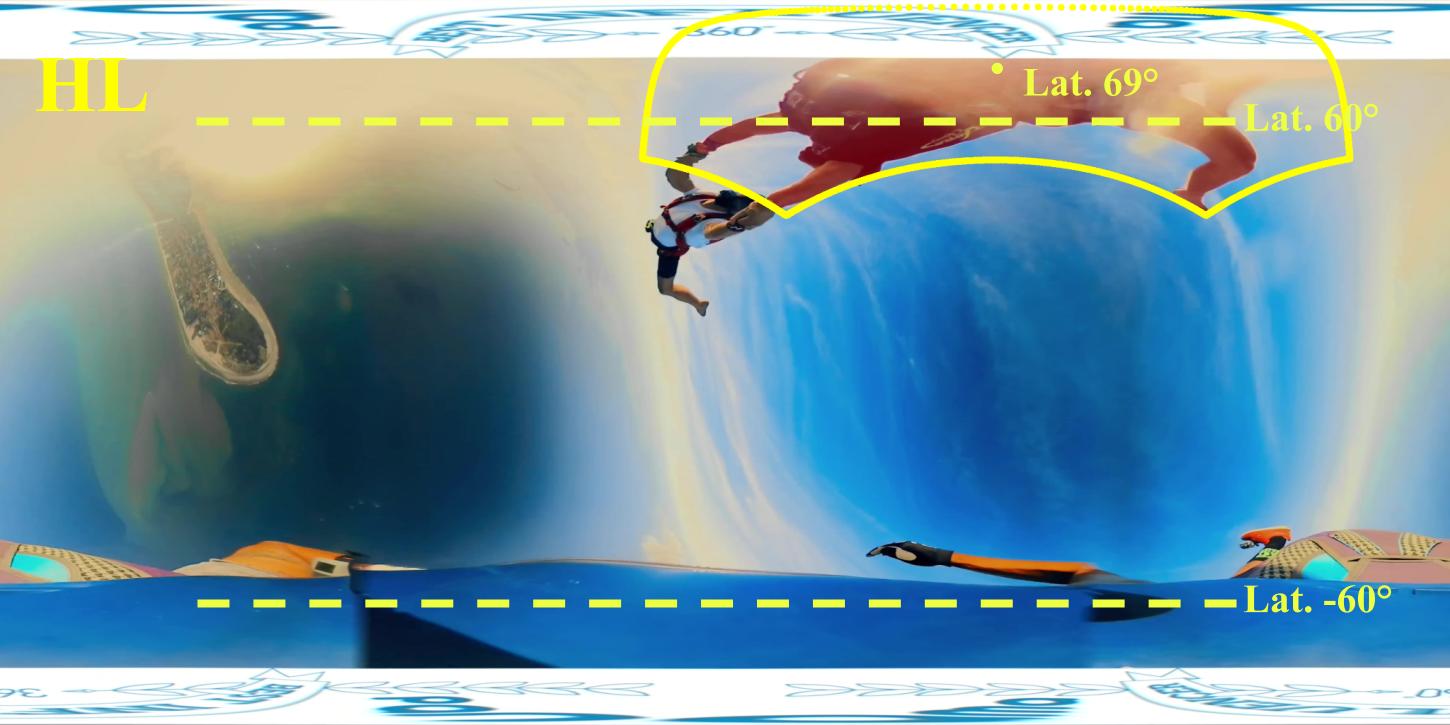}}
    \vspace{-0.9em}    \\
    \subfloat{\includegraphics[width=\imgw\linewidth, height=\imgh\linewidth]{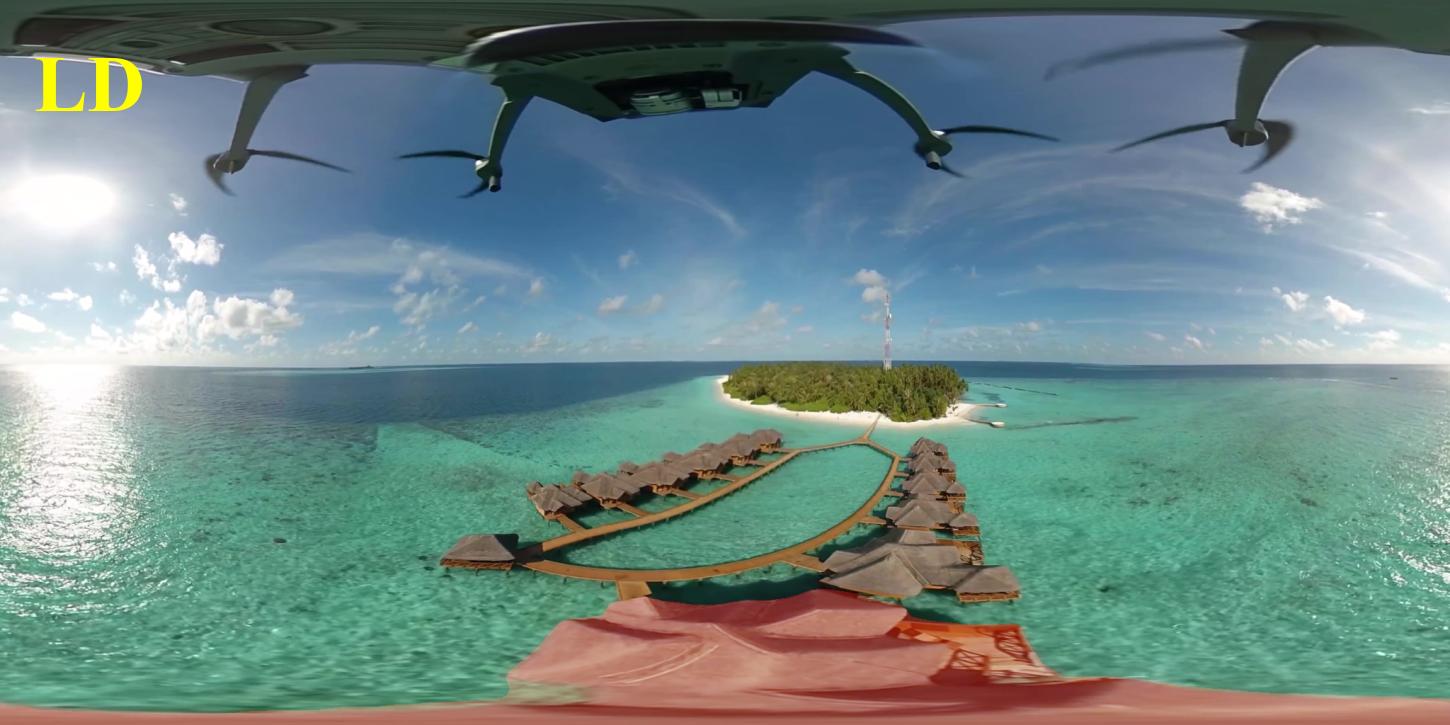}}\!
    \subfloat{\includegraphics[width=\imgw\linewidth, height=\imgh\linewidth]{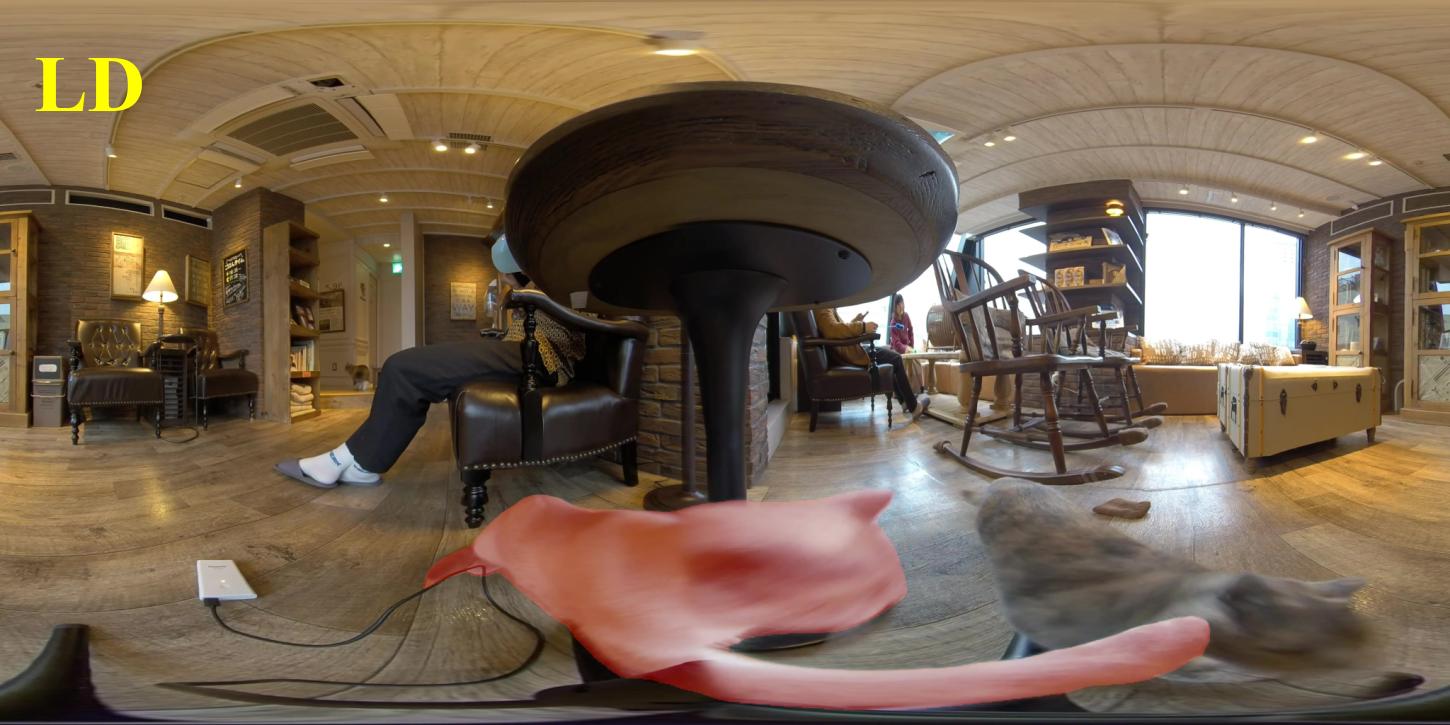}}\!
    \subfloat{\includegraphics[width=\imgw\linewidth, height=\imgh\linewidth]{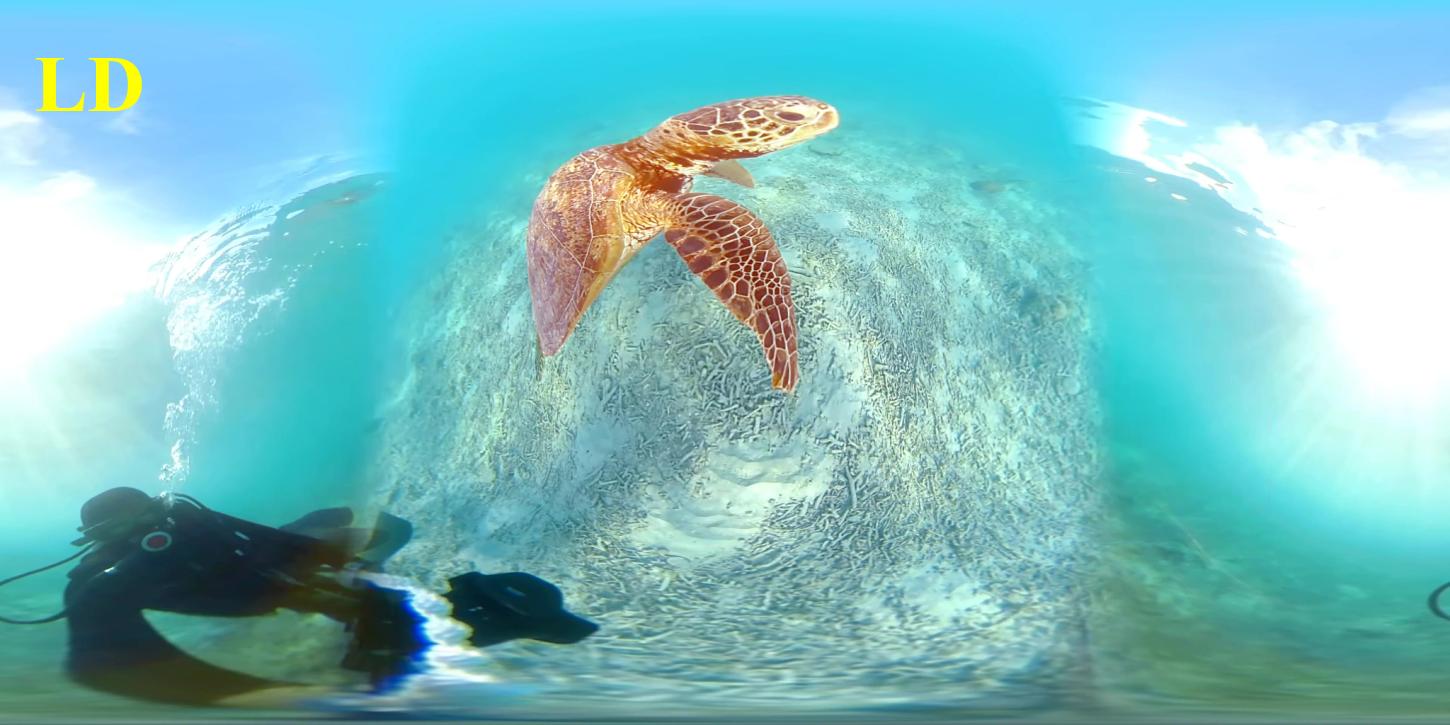}}
    \\
    
    \caption{Visual examples of the 7 distinct attributes of the 360$\degree$ images described in Section \ref{sec4b} and Table \ref{tab:attibute} of the main paper. The abbreviation of each attribute is labeled in the upper left corner of the images. For SA, the affected parts of the target are enlarged on the side. For CB, the left and right border parts are merged at the center. For FMS, the target's BFoV displacement between the current (Cur) and previous (Pre) frames is traced. For LFoV, the vertical (FoV-V) and horizontal (FoV-H) spans of the large BFoV are listed. For LV, the upper and lower patches from different frames (Fr.) are split by a solid line, with the target's BFoV latitude (Lat.) noted. For HL, pole regions are marked by dashed lines, showing their latitude (Lat.) and the target's BFoV. For LD, targets affected by this challenge are covered by masks. \\
    Included sequences: Test-074, Test-027, and Train-040 for SA; Test-029, Test-012, and Test-003 for CB; Test-021, Test-033, and Test-063 for FMS; Test-020, Test-004, and Test-016 for LFoV; Test-090, Test-109, and Test-103 for LV; Test-022, Train-044, and Test-113 for HL; Test-006, Test-049, and Train-166 for LD.}
    \label{fig:f12}
\end{figure}

\noindent \textbf{Visual Examples of Handling CB Attribute.}
As described in Section \ref{sec:framework} of the main paper, the proposed framework extracts local search regions around the target object in 360$\degree$ frames and applies an arbitrary tracker within these extracted regions. 
Instead of treating each input 360$\degree$ frame as just a flat image, our approach considers the spherical continuity of these frames. 
Specifically, when the target object crosses the left or right boundary of a 360$\degree$ frame, our framework ensures seamless tracking by extracting overlapping local search regions that span both sides of the border. These regions are then unwrapped and provided to the traditional tracker, allowing it to detect and track the target object without interruption.
This ensures the tracker maintains continuity and avoids losing the target during the CB challenge. 

We provide intermediate visual examples in Figure \ref{fig:360fw-iter} to further illustrate this capability. As shown in Figure \ref{fig:360fw}, the framework is divided into four steps for visualization purposes. In Step 1, the search region from the previous frame is projected onto the current frame. Step 2 shows the tracker estimating the target mask within the search region. In Step 3, the estimated mask is reprojected onto the original frame. Finally, in Step 4, the updated target mask and the next frame's search region are visualized. These steps demonstrate how the framework handles CB challenges seamlessly across iterations.

\subsection{Benchmark Dataset: 360VOTS}
\noindent \textbf{Visual Examples of Unique Attributes.}
As discussed in Section \ref{sec4b} of the main paper, the distinct features of the 360$\degree$ image are well represented in 360VOTS. The unique attributes in 360VOT including, stitching artifact (SA), crossing border (CB), fast motion on the sphere (FMS), large FoV (LFoV), latitude variation (LV), high latitude (HL), and large distortion (LD). To illustrate 7 distinct attributes in 360$\degree$ data completely, we further include visual examples in Figure \ref{fig:f12} here, each row demonstrates a representative attribute.
\\

\noindent \textbf{Dataset Categorization.}
Following the data collection and annotation schemes of \cite{vot360}, we provide 191 new sequences and refine 99 of the original 120 sequences in 360VOT\cite{vot360}. All the sequences are still grouped into 4 main classes and each main class is further subdivided into a total of 62 subclasses, covering \textit{humans} (pedestrian, skateboarder, skiing, skydiver, hiker, wing outfit, hockey player, dancer, kid, surfer, diver, rollerblader, ping-pong player, and boxing), \textit{objects} (car, kart, drone, jet, boat, building, RC car, F1 car, helicopter, basket, cloud, train, brand, bus, lego, pickup truck, fountain, shoes, cup, tire, parachute, wood, and helmet), \textit{animals} (kitty, rabbit, rhino, dog, elephant, bird, panda, tiger, dolphin, horse, squirrel, giraffe, kangaroo, turtle, monkey, fox, bear, wolf, shark, and penguin), and \textit{human \& carrier cases} (human \& boat, human \& bike, human \& moto, and human \& horse). 

\begin{figure*}[t]
    \centering
    \captionsetup[subfloat]{farskip=3pt,captionskip=-9pt,labelfont={rm,footnotesize},textfont=footnotesize}    
    \subfloat[The category tree of 360VOT.\label{fig:statistic-vot}]{\includegraphics[width=\linewidth]{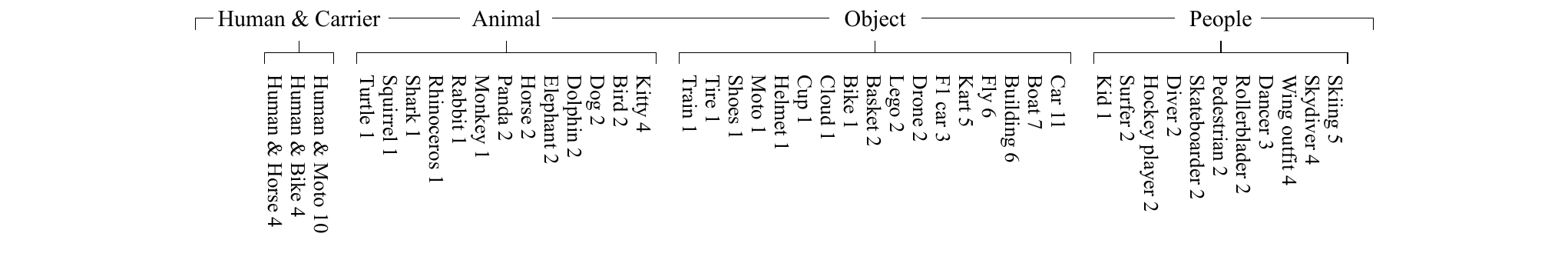}}\\
    \subfloat[The category tree of 360VOS.\label{fig:statistic-vots}]{\includegraphics[width=\linewidth]{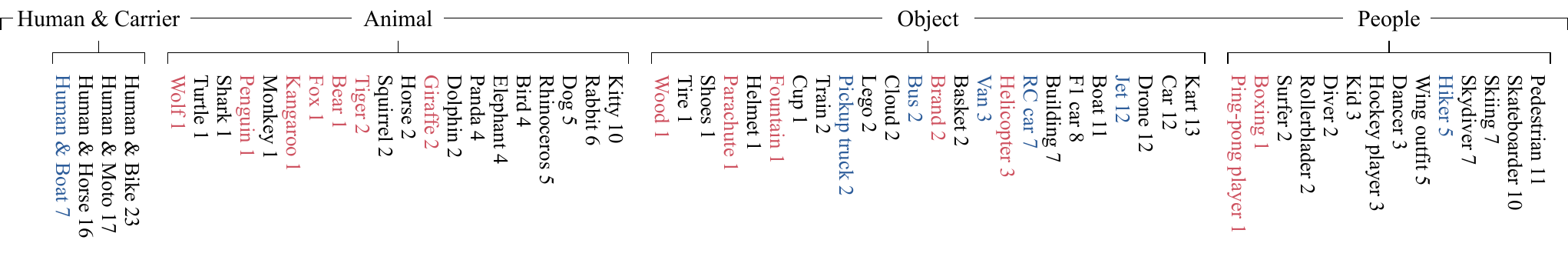}}
    
    \caption{The tree maps of the category structures for 360VOTS. Newly introduced classes in 360VOS are highlighted in \st{red}, while the classes that have been re-categorized from the original 360VOT are highlighted in \nd{blue}.} 
    \label{fig:statistic}
\end{figure*}

The proposed benchmark dataset, 360VOTS, is composed of two sub-datasets: 360VOT and 360VOS. The 360VOS sub-dataset introduces a broader and more comprehensive collection of sequences with additional subclasses. Moreover, the original categories in 360VOT \cite{vot360} have been refined through re-categorization and re-naming to better capture the specific features of the target objects. Figure~\ref{fig:statistic} illustrates the categorization hierarchy for both 360VOT and 360VOS through tree maps, where the branches represent the four main classes and the leaves correspond to the respective subclasses within each main class.

The primary objective of our dataset is to provide a diverse and challenging set of scenarios, including edge cases where certain categories have limited real-world examples. Such cases are particularly valuable for evaluating the performance of object tracking and segmentation methods under rare or underrepresented conditions. To this end, we intentionally include categories with small instance counts to test the generalization and robustness of these methods in handling rare or unique objects, a critical requirement for real-world applications. Additionally, as the dataset focuses on long-term tracking and segmentation, even categories with a single instance often span hundreds of frames, thereby enabling models to learn and adapt effectively to these scenarios. The distributions of sequences and frames across categories are depicted in Figure~\ref{fig:frseq}.

\begin{figure*}[t]
    \centering
    \captionsetup[subfloat]{farskip=5pt,captionskip=-5pt,labelfont={rm,footnotesize},textfont=footnotesize}   
    \subfloat{\includegraphics[width=\linewidth]{figure/s1/c13/frames1.png}}\\
    \subfloat{\includegraphics[width=\linewidth]{figure/s1/c13/seq1.png}}
    \caption{Histogram of the target categories distribution in frame-based and sequence-based.  
    }
    \label{fig:frseq}
\end{figure*}

\subsection{Experiments}
\begin{figure}[t]
    \centering
    \captionsetup[subfloat]{labelfont={rm,footnotesize},textfont=footnotesize} 
    \def\imgw{0.49}
    \def\gpw{0.99}
    \subfloat[Overestimated sample where mask near the pole. \label{fig:ws_ga}]{
        \begin{minipage}{\gpw\linewidth}
            \centering
            \includegraphics[width=\imgw\linewidth]{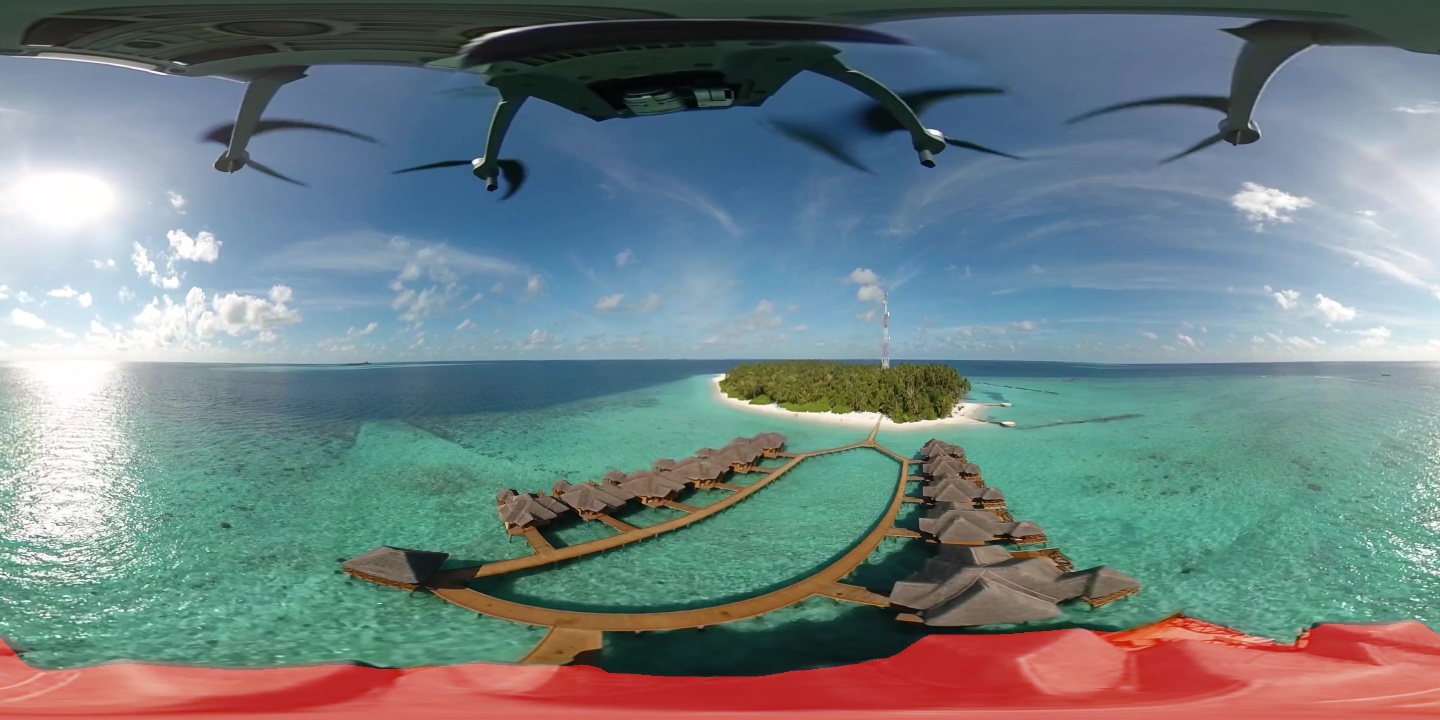}
            \includegraphics[width=\imgw\linewidth]{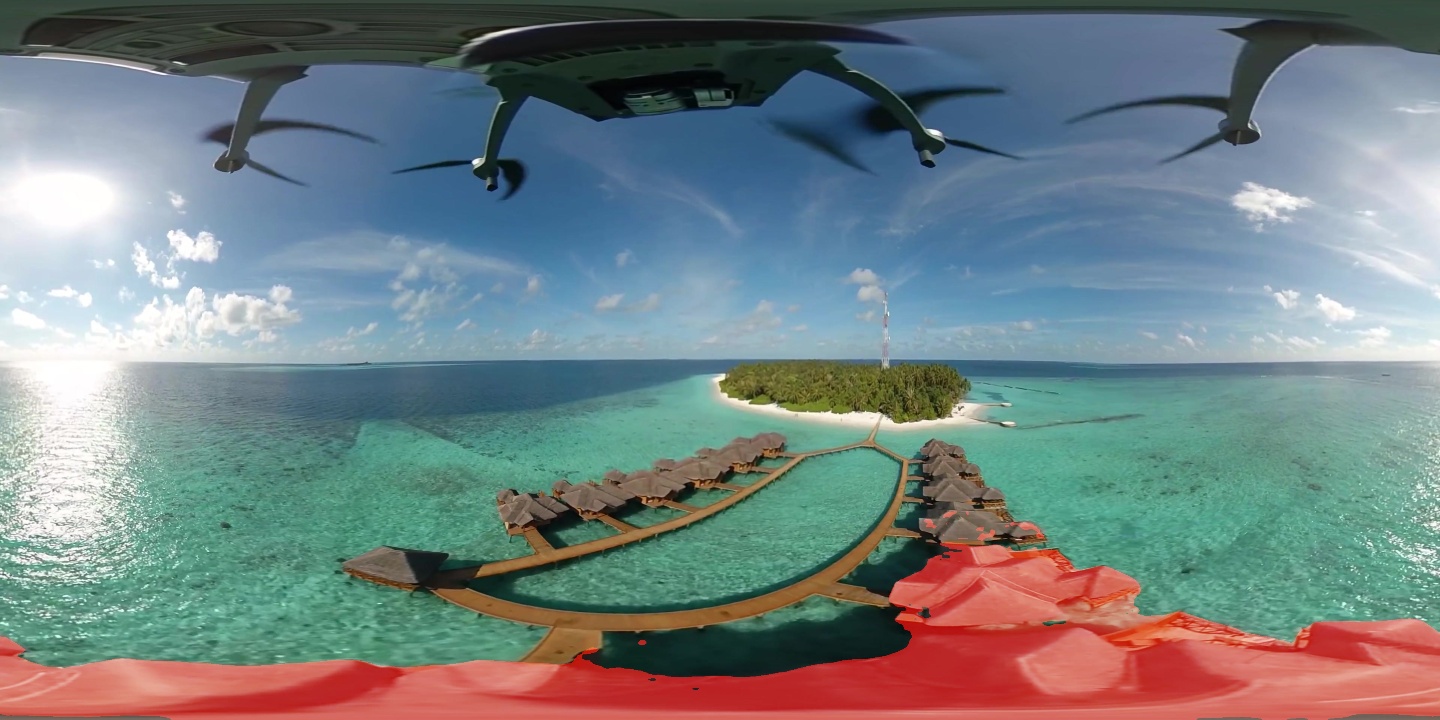}
        \end{minipage}
    }
    \\[0pt]
    \subfloat[Overestimated sample where mask near the equator. \label{fig:ws_gb}]{
        \begin{minipage}{\gpw\linewidth}
            \centering
            \includegraphics[width=\imgw\linewidth]{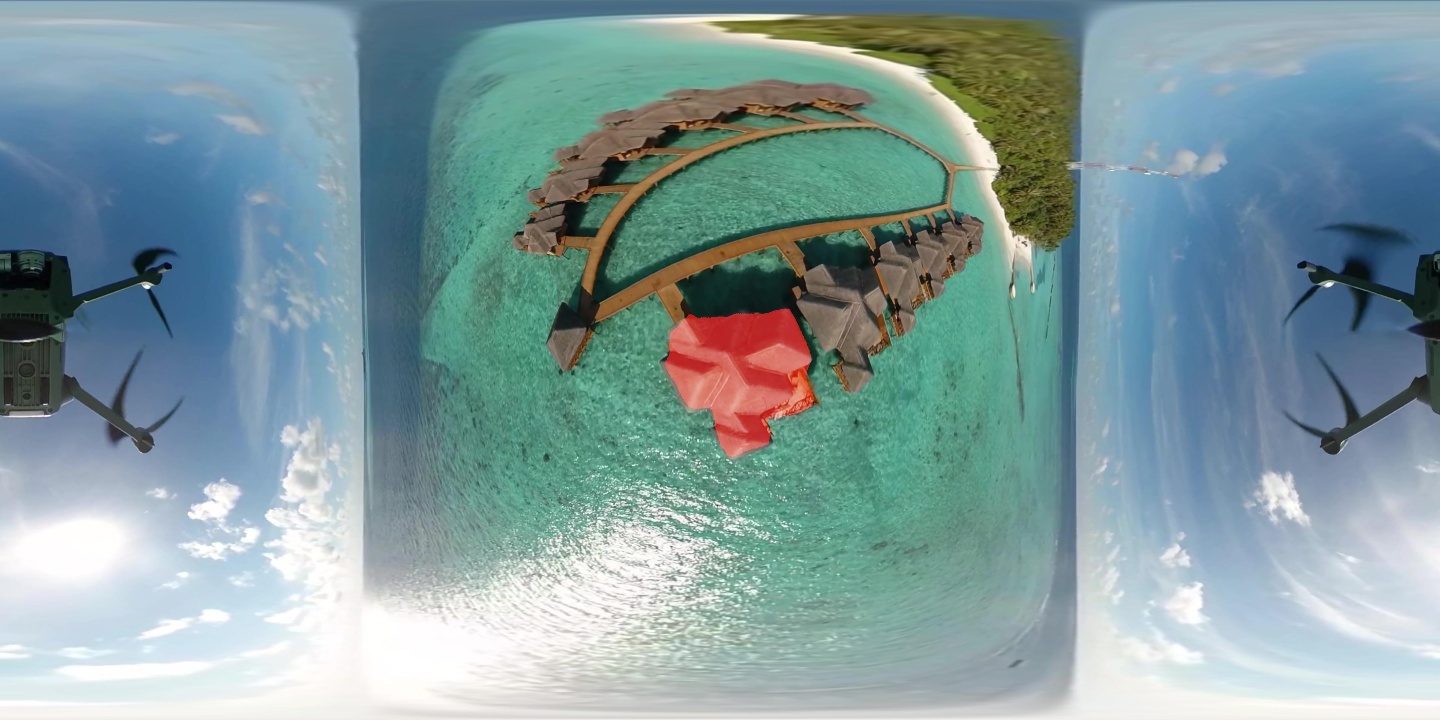}
            \includegraphics[width=\imgw\linewidth]{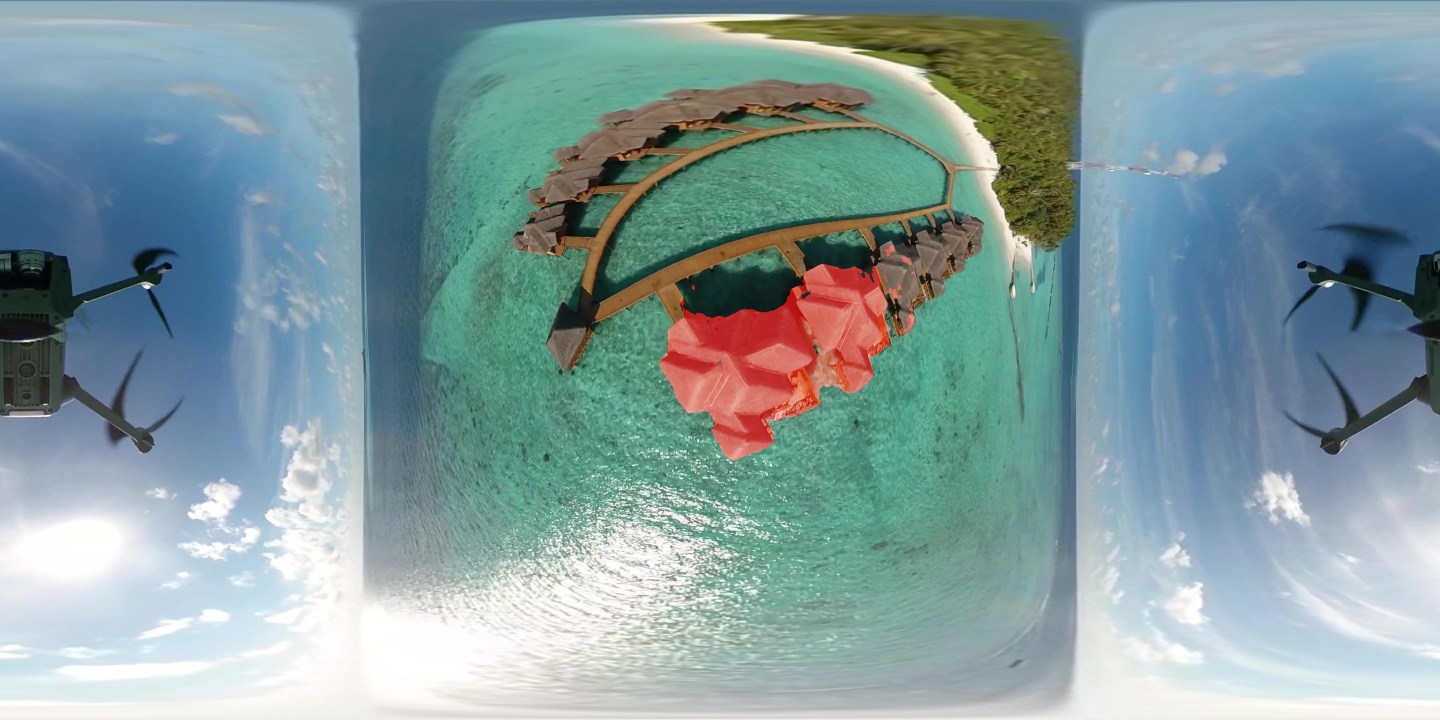}
        \end{minipage}
    }
    \\[0pt] 
    \subfloat[Underestimated sample where mask near the pole. \label{fig:ws_gc}]{
        \begin{minipage}{\gpw\linewidth}
            \centering
            \includegraphics[width=\imgw\linewidth]{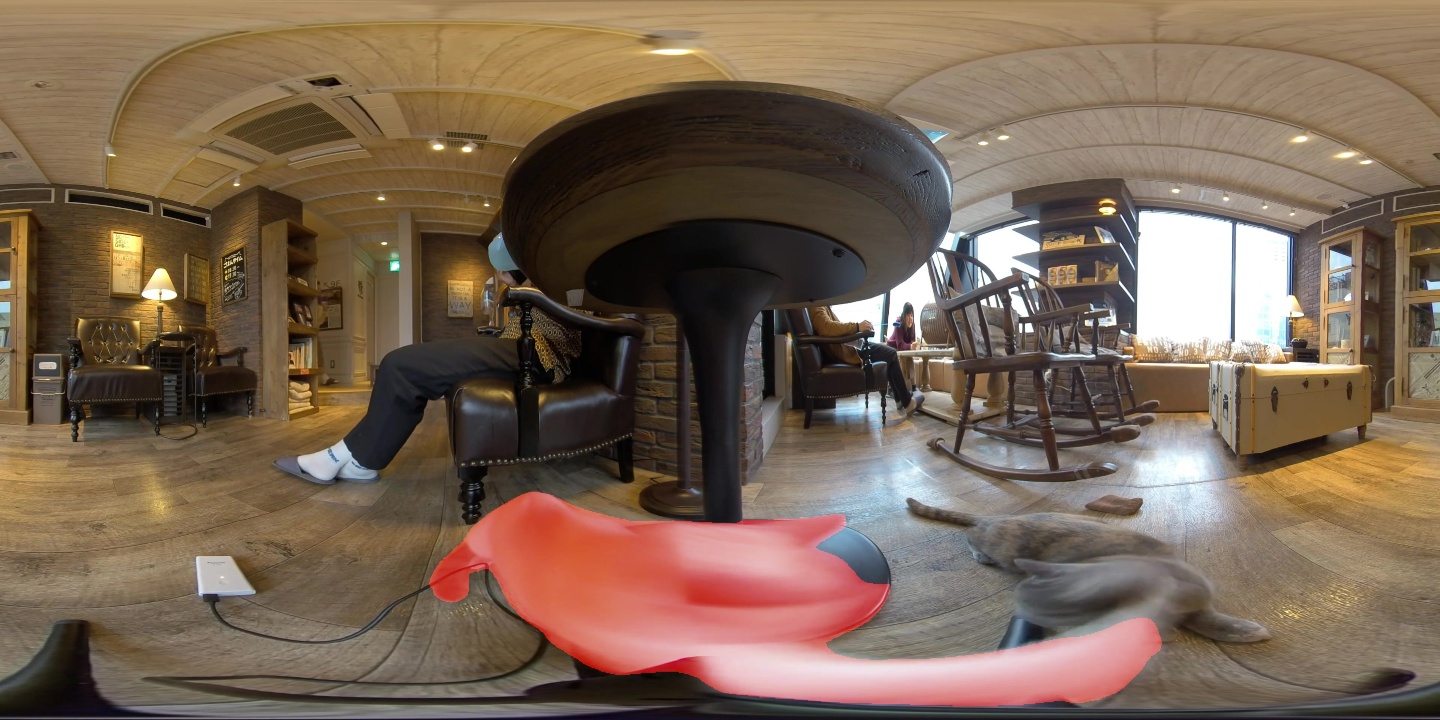}
            \includegraphics[width=\imgw\linewidth]{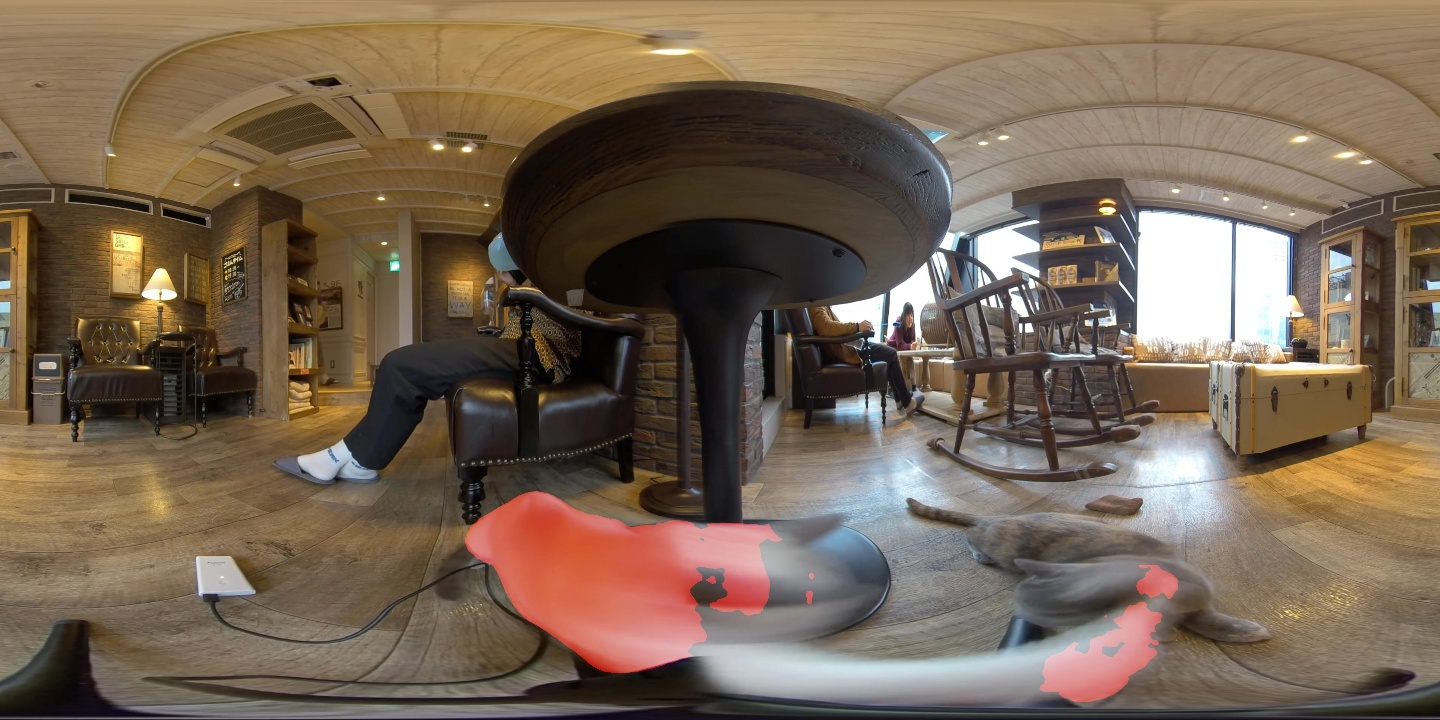}
        \end{minipage}
    }
    \\[0pt]
    \subfloat[Underestimated sample where mask near the equator. \label{fig:ws_gd}]{
        \begin{minipage}{\gpw\linewidth}
            \centering
            \includegraphics[width=\imgw\linewidth]{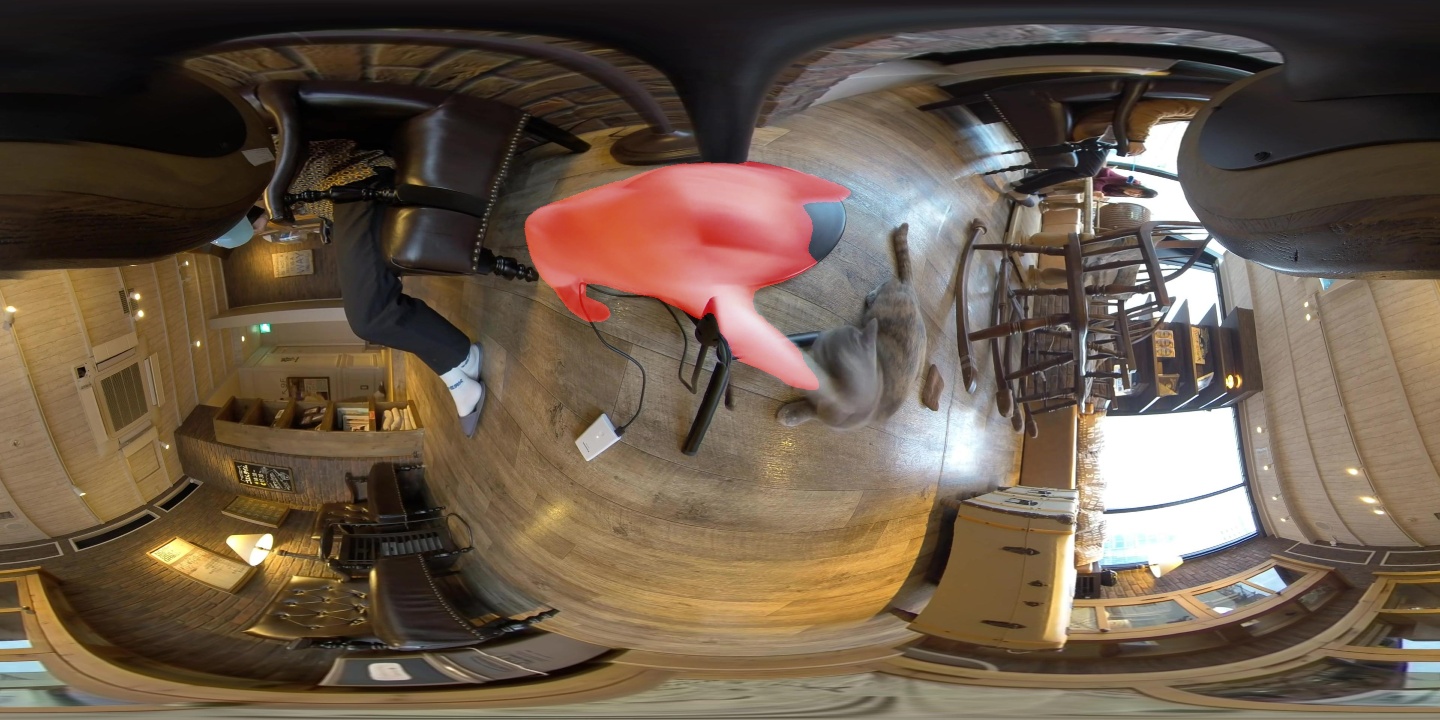}
            \includegraphics[width=\imgw\linewidth]{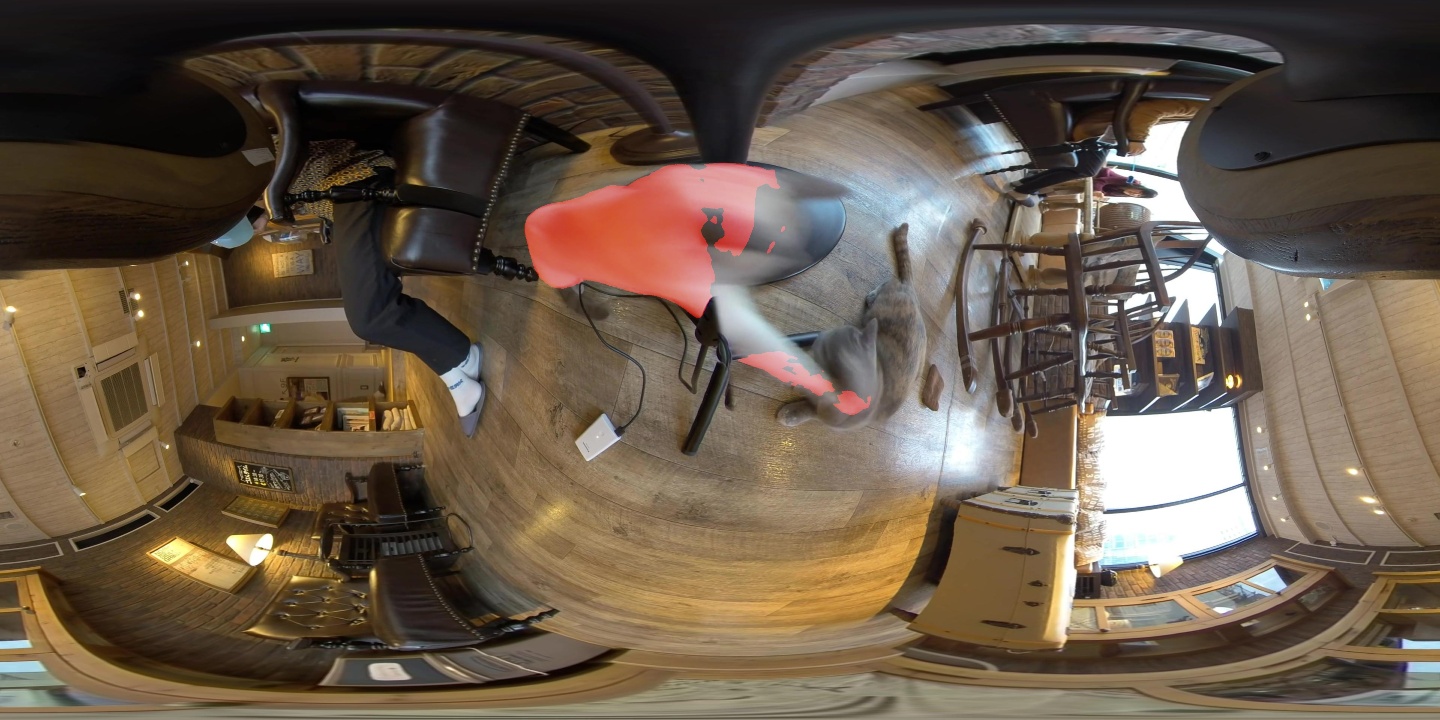}
        \end{minipage}
    }
    \caption{
    Sample 360$\degree$ images used to compute metrics in \textbf{Table} \ref{tab:met}, grouped into 4 pairs. Each image pair consists of a ground truth RGB frame (left) and its corresponding RGB frame (right) covered by the estimated semi-transparent mask. 
    Figure pairs \ref{fig:ws_ga} and \ref{fig:ws_gc} show original RGB frames and masks where the target objects are located near the pole. 
    Figure pairs \ref{fig:ws_gb} and \ref{fig:ws_gd} are obtained by rotating the original RGB frames and masks 90$\degree$ along the pitch axis (pulling the frame center upward), placing the target objects near the equator.
    This layout allows for a visual comparison between overestimated (Figures \ref{fig:ws_ga} and \ref{fig:ws_gb}) and underestimated (Figures \ref{fig:ws_gc} and \ref{fig:ws_gd}) samples. \\
    Included sequences: Test-006 for overestimated samples and Test-049 for underestimated samples.
    }
    \label{fig:ws_g}
\end{figure}

\noindent \textbf{Efficiency of the Spherical Metrics.}
The proposed spherical metrics $\mathcal{J}_{sphere}$ and $\mathcal{F}_{sphere}$ apply the pre-computed spherical weights $\mathfrak{W}$ on the standard metrics $\mathcal{J}$ and $\mathcal{F}$, alleviating the issues of overestimated segmentation performance in high-latitude areas (i.e., close to the top or bottom) on 360$\degree$ images. To further highlight the importance of this correction, we provide both a visual comparison (see Figure \ref{fig:ws_g}) and a quantitative comparison (see Table \ref{tab:met}) to demonstrate the differences between standard metrics and the proposed spherical metrics. For example, in the case of overestimated segmentation results near the poles (Figure \ref{fig:ws_ga}), $\mathcal{J}$ overstates the performance by a significant margin (+27.4\%) compared to $\mathcal{J}_{sphere}$. Conversely, for underestimated segmentation results near the poles (Figure \ref{fig:ws_gc}), $\mathcal{J}$ understates performance (-17.5\%). However, when the same segmentation results are evaluated near the equator, where distortions are minimal (Figures \ref{fig:ws_gb} and \ref{fig:ws_gd}), the differences between $\mathcal{J}$ and $\mathcal{J}_{sphere}$ are negligible ($\Delta\mathcal{J} \approx 0$). These results quantitatively confirm that the proposed spherical metrics more accurately reflect segmentation quality in 360$\degree$ videos by appropriately weighting regions based on their true spherical contribution.
\\

\begin{table}[t]
    \setlength{\tabcolsep}{2pt}
    \caption{Comparison of metrics for the sample 360$\degree$ images shown in \textbf{Figure} \ref{fig:ws_g}. The percentage differences $\Delta\mathcal{J}$ and $\Delta\mathcal{F}$ represent the relative change between the standard and spherical metrics.}
    \centering
    \footnotesize
    \begin{tabular}{lcccccc}
    \toprule
         Sample Frames & $\mathcal{J}$ & $\mathcal{F}$ & $\mathcal{J}_{sphere}$ & $\mathcal{F}_{sphere}$ & $\Delta\mathcal{J}$ & $\Delta\mathcal{F}$\\
    \midrule   
        Over-poles (Fig. \ref{fig:ws_ga}) & 0.792 & 0.655 & 0.575 & 0.550 & -27.4\% & -16.0\% \\
        Over-equator (Fig. \ref{fig:ws_gb}) & 0.571 & 0.600 & 0.574 & 0.603 & 0.5\% & 0.6\% \\
        Under-poles (Fig. \ref{fig:ws_gc}) & 0.508 & 0.518 & 0.597 & 0.558 & 17.5\% & 7.7\% \\
        Under-equator (Fig. \ref{fig:ws_gd}) & 0.609 & 0.595 & 0.597 & 0.591 & -1.9\% & -0.6\% \\
    \bottomrule
    \end{tabular}
    \label{tab:met}
\end{table}

\noindent \textbf{Details of Retraining VOS Tracker.}\label{supp:vos}
{We retrained XMem\cite{xmem} (denoted as XMem$^\star$) following the original training scheme and hyperparameters.} 
At the beginning of retraining, we used the initialized weights of XMem\cite{xmem}, which were pre-trained on static images and the BL30K dataset\cite{bl30k}. We then conducted the main training stage on a combined dataset consisting of our 360VOS training set, YouTubeVOS\cite{ytvos2018}, and DAVIS\cite{davis2017}. 
Unlike the original XMem\cite{xmem}, which was trained on 480p frames with a random cropping size of $384 \times 384$, we adjusted the training scheme to handle the high-resolution and unique characteristics of 360$\degree$ images. To maintain sufficient details and the large field-of-view advantage of 360$\degree$ images, we resized all frames in the dataset to 720p ($height = 720$) and applied random cropping to $512 \times 512$ patches accordingly. This adjustment ensured efficient training while preserving the resolution necessary for omnidirectional video object segmentation. For optimization, we employed a combination of bootstrapped cross-entropy loss and dice loss, using the AdamW\cite{adamw} optimizer with a learning rate of 1e-5 (reduced by a factor of 10 after the first 80K iterations) and a weight decay rate of 0.05. These settings closely follow the default parameters used in the original XMem\cite{xmem} training scheme. The model was trained for 100K iterations over approximately 3.5 days on 4 RTX3090 GPUs, with a batch size of 8.
\\

\begin{table}[t]
    \caption{Quantitative comparison of segmentation performance on sequences with large target objects or objects close to the camera in the 360VOS dataset. Metrics include standard metrics ($\mathcal{J}$ and $\mathcal{F}$), spherical metrics ($\mathcal{J}_{sphere}$ and $\mathcal{F}_{sphere}$), and the average spherical metric $(\mathcal{J} \& \mathcal{F})_{sphere}$. The outperformed scores of $(\mathcal{J} \& \mathcal{F})_{sphere}$ in each sequence are marked in bold. Test-031 is rendered, while Test-101 and Test-120 are from the real world.
    }
    \centering
    \footnotesize
    \begin{tabular}{llccc}
    \toprule
    \multirow{2}{*}{Seq.} & \multirow{2}{*}{Trackers} & \multicolumn{3}{c}{360VOS} \\ \cmidrule(lr){3-5}
      && $\mathcal{J}_{sphere}$ & $\mathcal{F}_{sphere}$ & $(\mathcal{J} \& \mathcal{F})_{sphere}$ \\
    \midrule
    \multirow{5}{*}{\shortstack[c]{Test-\\031}}
     & XMem\cite{xmem} & 0.798 & 0.905 & \textbf{0.851} \\ 
     & XMem++\cite{xmem2} & 0.797 & 0.902 & 0.850 \\
     & XMem$^\star$ & 0.796 & 0.885 & 0.840 \\
     & XMem-360 & 0.782 & 0.861 & 0.822 \\
     & XMem-360$^\star$ & 0.798 & 0.882 & 0.840 \\
    \midrule \midrule
    \multirow{5}{*}{\shortstack[c]{Test-\\101}} 
     & XMem\cite{xmem} & 0.837 & 0.947 & 0.892 \\
     & XMem++\cite{xmem2} & 0.835 & 0.946 & 0.890 \\
     & XMem$^\star$ & 0.407 & 0.607 & 0.507 \\
     & XMem-360 & 0.827 & 0.943 & 0.885 \\
     & XMem-360$^\star$ & 0.869 & 0.972 & \textbf{0.920} \\
    \midrule \midrule
    \multirow{5}{*}{\shortstack[c]{Test-\\120}} 
     & XMem\cite{xmem} & 0.853 & 0.984 & 0.918 \\
     & XMem++\cite{xmem2} & 0.063 & 0.169 & 0.116 \\
     & XMem$^\star$ & 0.880 & 0.993 & \textbf{0.937} \\
     & XMem-360 & 0.830 & 0.983 & 0.906 \\
     & XMem-360$^\star$ & 0.872 & 0.995 & 0.933 \\
    \bottomrule
    \end{tabular}
    \label{tab:lrgdist}
\end{table}

\noindent \textbf{Details of Finetuning VOT Tracker.} \label{supp:vot}
As we mentioned in the main paper, we finetuned LoRAT\cite{lorat} (denoted as LoRAT$^\star$) to further demonstrate the effectiveness of our new 360VTOS training set. We selected the most powerful pretrained variants of LoRAT\cite{lorat} with the backbone network of ViT-g\cite{vitg} and used the search region size of $[378\times 378]$.
In order to prepare the BBox ground truth to finetune LoRAT$^\star$, we computed the minimal axis-aligned rectangle that encloses all foreground pixels within each mask provided in the 360VOTS training sets.
There are 143,634 frames and 138,605 valid BBoxes (5,029 BBoxes are invalid due to occlusion or disappearance).
Finally, we finetuned the model for 10 epochs using the default training scheme and hyperparameters. The finetuning process took about 3 hours on 6 RTX4090D GPUs, with a batch size of 64.
\\


\noindent \textbf{Capability on Extreme Distortion.}
In some extreme scenarios during the tracking, the target objects may be huge or the objects close to the camera, presenting unique challenges due to significant distortion. In such cases, the increased size of the search region, or even treating the entire 360$\degree$ frame as the search region, may lead to unavoidable distortion.
While our 360 tracking framework may not always extract a search region with substantially reduced distortion in these extreme cases, it still provides an important advantage by rotating the 360$\degree$ frame and centering the target object. This ensures that the target remains centralized and facilitates trackers in localizing the object more effectively, despite the distortion.
In addition, we intentionally collected sequences in the 360VOS dataset that represent these edge cases, either captured from real-world videos or rendered. We selected sequences where the maximum BFoV of the target object across the entire sequence occupies more than 85\% FoV of the frame (e.g., vertically $\ge 153\degree$ or horizontally $\ge 306\degree$) and distributed them into the training set (e.g., Train-087, Train-089, Train-094, and Train-161) and the test set (e.g., Test-031, Test-101, and Test-120). 
Furthermore, we retrained XMem\cite{xmem} on the 360VOS training set to better adapt its features to handle these extreme scenarios.

\begin{figure}[t]
    \centering
    \includegraphics[width=0.99\linewidth]{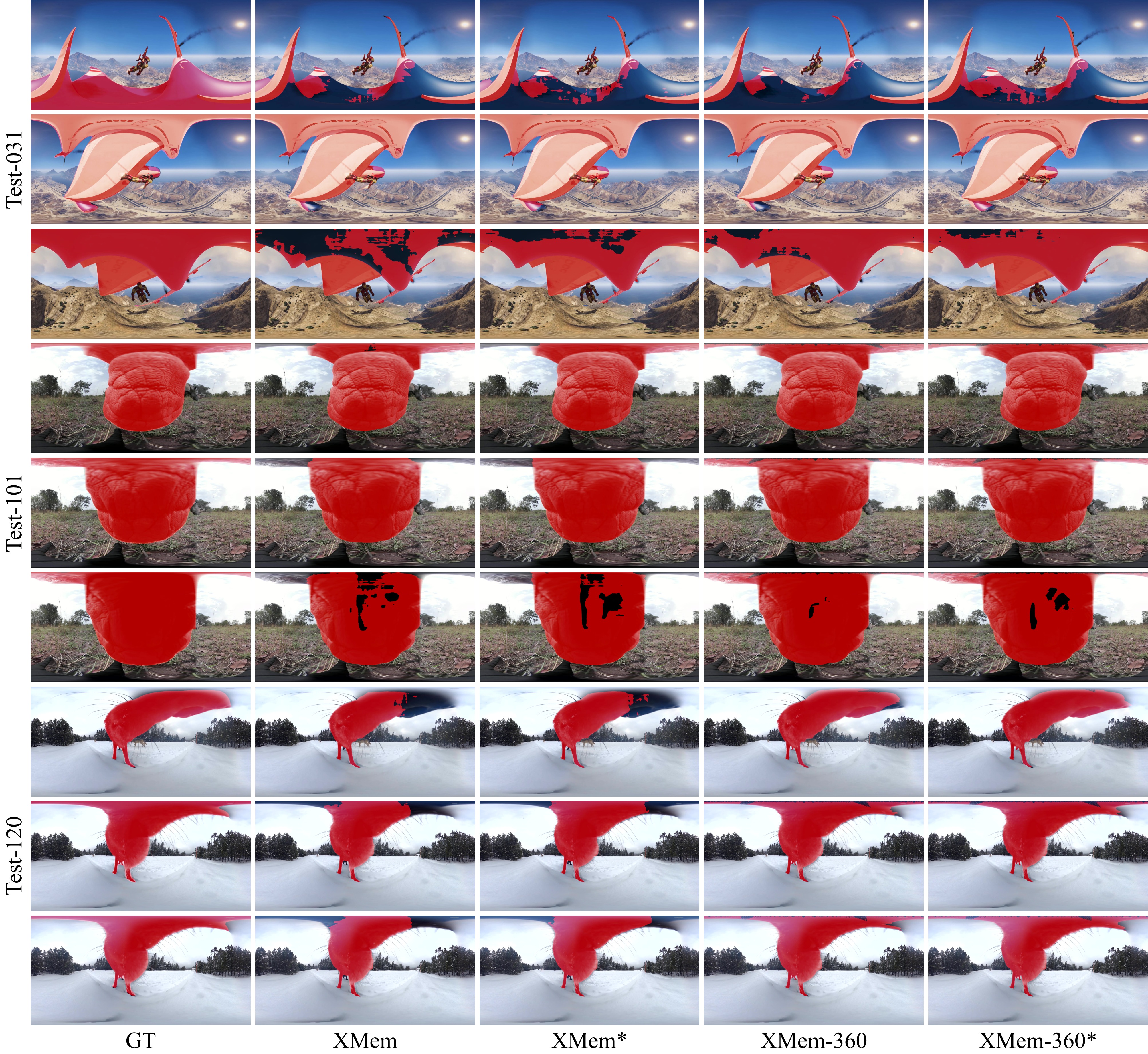}
    \caption{Visual comparisons of segmentation performance on sequences with large target objects or objects close to the camera in the 360VOS dataset. \\
    Included sequence: Test-031, Test-101, and Test-120.
    }
    \label{fig:lrgdist}
\end{figure}

In Table \ref{tab:lrgdist} and Figure \ref{fig:lrgdist}, we present a comprehensive comparison of both quantitative and visual performance on these high-distortion scenarios across several existing methods: XMem \cite{xmem}, XMem++ \cite{xmem2}, and our proposed baselines XMem$^\star$, XMem-360, and XMem-360$^\star$. 
While the baselines XMem-360 and XMem-360$^\star$ demonstrate competitive performance in some cases, such as Test-101 and Test-120, the results presented here are not fully representative due to the small number of sequences evaluated. For example, in Test-031, a rendered sequence, XMem \cite{xmem} outperforms our baselines, achieving a $(\mathcal{J} \& \mathcal{F})_{sphere}$ score of 0.851. Meanwhile, for Test-101 and Test-120, our framework demonstrates strong performance, particularly with XMem-360$^\star$ achieving a $(\mathcal{J} \& \mathcal{F})_{sphere}$ score of 0.920 and 0.933, respectively. Although these results suggest that the 360 tracking framework is capable of handling some challenging real-world scenarios, further evaluation of a more extensive and diverse set of edge-case sequences is required to draw more definitive conclusions.
\\

\begin{table}[h]
    \caption{Ablation study on the 360 tracking framework. The settings are divided into three groups and each group evaluates 7 different values while keeping the other 2 settings consistent with the baseline. The baseline settings are SR Ratio = 2.0, SR Min = 90$\degree$, and Max Loss = 4, represented by the rows with a dash $-$ in the $\Delta$ column. The table compares standard metrics ($\mathcal{J}$ and $\mathcal{F}$), spherical metrics ($\mathcal{J}_{sphere}$ and $\mathcal{F}_{sphere}$), and the average spherical metric $(\mathcal{J} \& \mathcal{F})_{sphere}$, along with the relative difference $\Delta$ compared to the baseline.}
    \centering
    \footnotesize
    \begin{tabular}{lccccr}
    \toprule
    \multicolumn{2}{l}{\multirow{2}{*}{Settings}} & \multicolumn{4}{c}{360VOS XMem-360$^\star$} \\ \cmidrule(lr){3-6}
      && $\mathcal{J}_{sphere}$ & $\mathcal{F}_{sphere}$ & $(\mathcal{J} \& \mathcal{F})_{sphere}$ & \multicolumn{1}{c}{$\Delta$} \\
    \midrule
    \multirow{7}{*}{\shortstack[c]{SR\\Ratio}} 
     & 1.2 & 0.665 & 0.785 & 0.725 & -1.9\% \\ 
     & 1.6 & 0.663 & 0.785 & 0.724 & -2.0\% \\
     & 2.0 & 0.677 & 0.801 & 0.739 & \multicolumn{1}{c}{-} \\
     & 2.4 & 0.664 & 0.784 & 0.724 & -2.0\% \\
     & 2.8 & 0.679 & 0.803 & 0.741 & 0.3\% \\
     & 3.2 & 0.680 & 0.804 & 0.742 & 0.4\% \\
     & 3.6 & 0.685 & 0.810 & 0.747 & 1.1\% \\
    \midrule \midrule
    \multirow{7}{*}{\shortstack[c]{SR\\Min}} 
     & 60$\degree$ & 0.628 & 0.750 & 0.689 & -6.8\% \\
     & 75$\degree$ & 0.657 & 0.779 & 0.718 & -2.8\% \\
     & 90$\degree$ & 0.677 & 0.801 & 0.739 & \multicolumn{1}{c}{-} \\
     & 105$\degree$ & 0.649 & 0.763 & 0.706 & -4.5\% \\
     & 120$\degree$ & 0.662 & 0.782 & 0.722 & -2.3\% \\
     & 135$\degree$ & 0.673 & 0.796 & 0.734 & -0.7\% \\
     & 150$\degree$ & 0.657 & 0.780 & 0.719 & -2.7\% \\
    \midrule \midrule
    \multirow{7}{*}{\shortstack[c]{Max\\Loss}} 
     & 0 & 0.644 & 0.758 & 0.701 & -5.1\% \\
     & 2 & 0.652 & 0.769 & 0.711 & -3.8\% \\
     & 4 & 0.677 & 0.801 & 0.739 & \multicolumn{1}{c}{-} \\
     & 6 & 0.667 & 0.787 & 0.727 & -1.6\% \\
     & 8 & 0.679 & 0.804 & 0.742 & 0.4\% \\
     & 10 & 0.670 & 0.793 & 0.731 & -1.1\% \\
     & 12 & 0.681 & 0.804 & 0.742 & 0.4\% \\
    \bottomrule
    \end{tabular}
    \label{tab:abltab}
\end{table}

\noindent \textbf{Ablation on 360 Tracking Framework.}
We have conducted ablation experiments to analyze the impact of different settings on our proposed 360 tracking framework in Section \ref{sec:vos-evaluation}. The completed quantitative results are shown in Table \ref{tab:abltab}. For the SR Ratio, which determines the scaling factor between the BFoV of the predicted target mask and the search region in the next frame, performance variations remain modest, with changes in $(\mathcal{J} \& \mathcal{F})_{sphere}$ between -2.0\% and +1.1\%. For the SR Min, which controls the minimum BFoV of the search region, the baseline setting of 90° provides the best trade-off between accuracy and robustness. The extreme settings, such as SR Min = 60°, result in a larger drop in performance (e.g., $-6.8\%$) due to overly restrictive search regions. Max Loss indicates how many frames the search region remains unchanged after the tracker loses the target. The framework performs consistently well after the baseline setting of 4. Disabling this mechanism entirely (e.g., Max Loss = 0) results in more significant performance degradation (e.g., $-5.1\%$), highlighting the importance of the search region updating strategy in our framework 

Importantly, across all tested settings, the majority of performance changes fall within a narrow range (e.g., $\pm 5\%$), indicating that the framework is robust to parameter variations. Significant performance drops are observed only in extreme cases, such as overly small search regions or the complete absence of the search region update mechanism. These results provide strong evidence that our framework is stable and reliable, maintaining consistent performance across a wide range of settings.

\end{document}